\documentclass[lettersize,journal]{IEEEtran}
\usepackage{amsmath,amsfonts}
\usepackage{algorithmic}
\usepackage{algorithm}
\usepackage{array}
\usepackage[caption=false,font=normalsize,labelfont=sf,textfont=sf]{subfig}
\usepackage{textcomp}
\usepackage{stfloats}
\usepackage{url}
\usepackage{verbatim}
\usepackage{graphicx}
\usepackage{cite}
\usepackage{longtable}
\usepackage{colortbl}
\usepackage{pdflscape} % for 'landscape' environment
\usepackage{tabularray}

\usepackage{caption}
\usepackage[table,xcdraw]{xcolor} % Load xcolor with all necessary options
\usepackage{graphicx}
\usepackage{amsmath}
\usepackage{amssymb}
\usepackage{booktabs}
\usepackage{tikz}
\usetikzlibrary{mindmap} %for mindmap
\usepackage{arydshln}
\usepackage[margin=1in]{geometry}
\usepackage[normalem]{ulem} % For underlining
\usepackage{rotating} % For rotating tables and figures
\usepackage{adjustbox}

\usepackage[accsupp]{axessibility} % Improves PDF readability for those with disabilities.

\definecolor{JINZAMOMI}{RGB}{225,122,119} %red1
\definecolor{AKEBONO}{RGB}{241,148,131} %red2
\definecolor{TOKI}{RGB}{238,169,169} %pink2
\definecolor{MOGEI}{RGB}{123,162,63} %green1
\definecolor{HIWA}{RGB}{190,194,63} %green2
\definecolor{NOSHIMEHANA}{RGB}{43,95,117}
\definecolor{KUCHINASHI}{RGB}{246,197,85} %Yellow1
\definecolor{USUKI}{RGB}{250,214,137} %Yellow2
\definecolor{HANAASAGI}{RGB}{30,136,168} %blue
\definecolor{SORA}{RGB}{88,178,220} %blue textcolor
\definecolor{NAE}{RGB}{134,193,102} %green textcolor
\definecolor{KOKE}{RGB}{131,138,45} %green!!!
\definecolor{WASURENAGUSA}{RGB}{125,185,222} %blue2
\definecolor{GUNJYO}{RGB}{81,168,221} %blue3
\definecolor{KAMENOZOKI}{RGB}{165,222,228} %blue4

\usepackage[pagebackref,breaklinks,colorlinks]{hyperref}

\usepackage[capitalize]{cleveref}
\crefname{section}{Sec.}{Secs.}
\Crefname{section}{Section}{Sections}
\Crefname{table}{Table}{Tables}
\crefname{table}{Tab.}{Tabs.}

\usepackage{multirow}
\usepackage{fontawesome}
\usepackage{colortbl}
\usepackage{soul}

\usepackage{tcolorbox}
\usepackage{xpatch}
\usepackage{diagbox}
\usepackage{utfsym}

\usepackage{amsfonts}
\usepackage{mathrsfs}

\usepackage[T1]{fontenc} 
\pdfmapline{+delphine < Delphine.ttf <T1-WGL4.enc}

\usepackage{MnSymbol}
\usepackage{makecell}

\usepackage{placeins} % To use the \FloatBarrier command
\usepackage{float} % To use the [H] option for placing floats
\usepackage{graphicx} % To use \resizebox
\usepackage{adjustbox} % For more flexibility with table size adjustments

\begin{document}

\title{Vision Foundation Models in Remote Sensing: A Survey}

\author{
Siqi Lu\textsuperscript{1}, 
Junlin Guo\textsuperscript{1}, 
James R Zimmer-Dauphinee\textsuperscript{2}, \\
Jordan M Nieusma\textsuperscript{4}, 
Xiao Wang\textsuperscript{3}, 
Parker VanValkenburgh\textsuperscript{5}, \\
Steven A Wernke\textsuperscript{2}, 
Yuankai Huo\textsuperscript{1,6} \\
\textsuperscript{1}Department of Electrical and Computer Engineering, Vanderbilt University,\\
\textsuperscript{2}Department of Anthropology, Vanderbilt University,\\
\textsuperscript{3}Oak Ridge National Laboratory,\\
\textsuperscript{4}Data Science Institute, Vanderbilt University,\\
\textsuperscript{5}Department of Anthropology, Brown University\\
\textsuperscript{6}Department of Computer Science, Vanderbilt University,\\
}
\maketitle

\begin{abstract}
Artificial Intelligence (AI) technologies have profoundly transformed the field of remote sensing, revolutionizing data collection, processing, and analysis. Traditionally reliant on manual interpretation and task-specific models, remote sensing research has been significantly enhanced by the advent of foundation models—large-scale, pre-trained AI models capable of performing a wide array of tasks with unprecedented accuracy and efficiency. This paper provides a comprehensive survey of foundation models in the remote sensing domain. We categorize these models based on their architectures, pre-training datasets, and methodologies. Through detailed performance comparisons, we highlight emerging trends and the significant advancements achieved by those foundation models. Additionally, we discuss technical challenges, practical implications, and future research directions, addressing the need for high-quality data, computational resources, and improved model generalization. Our research also finds that pre-training methods, particularly self-supervised learning techniques like contrastive learning and masked autoencoders, remarkably enhance the performance and robustness of foundation models. This survey aims to serve as a resource for researchers and practitioners by providing a panorama of advances and promising pathways for continued development and application of foundation models in remote sensing.

\end{abstract}


\section{Introduction}
\label{sec:introduction}

\IEEEPARstart{A}{rtificial} Intelligence (AI) technologies have profoundly transformed the field of remote sensing, revolutionizing how data is collected, processed, and analyzed. Traditionally, remote sensing projects relied heavily on manual interpretation and task-specific models that required extensive labeled datasets and significant computational resources. However, the advent of AI and deep learning (DL) has ushered in a new era in which large-scale, pre-trained models, known as foundation models, are capable of performing a wide array of tasks with unprecedented accuracy and efficiency. These advancements have not only enhanced the potential applications of remote sensing but have also opened new avenues for its usage across various domains. 

In recent years, numerous vision foundation models have emerged, demonstrating remarkable performance in handling diverse remote sensing tasks. These models have shown the potential to significantly improve performance on multiple downstream tasks such as scene classification, semantic segmentation, object detection, and more. By leveraging vast amounts of pre-training data and sophisticated architectures, these foundation models have set new benchmarks in the field, making them indispensable tools for researchers and engineers alike.

\begin{figure*}
    \centering
    \includegraphics[width=\textwidth]{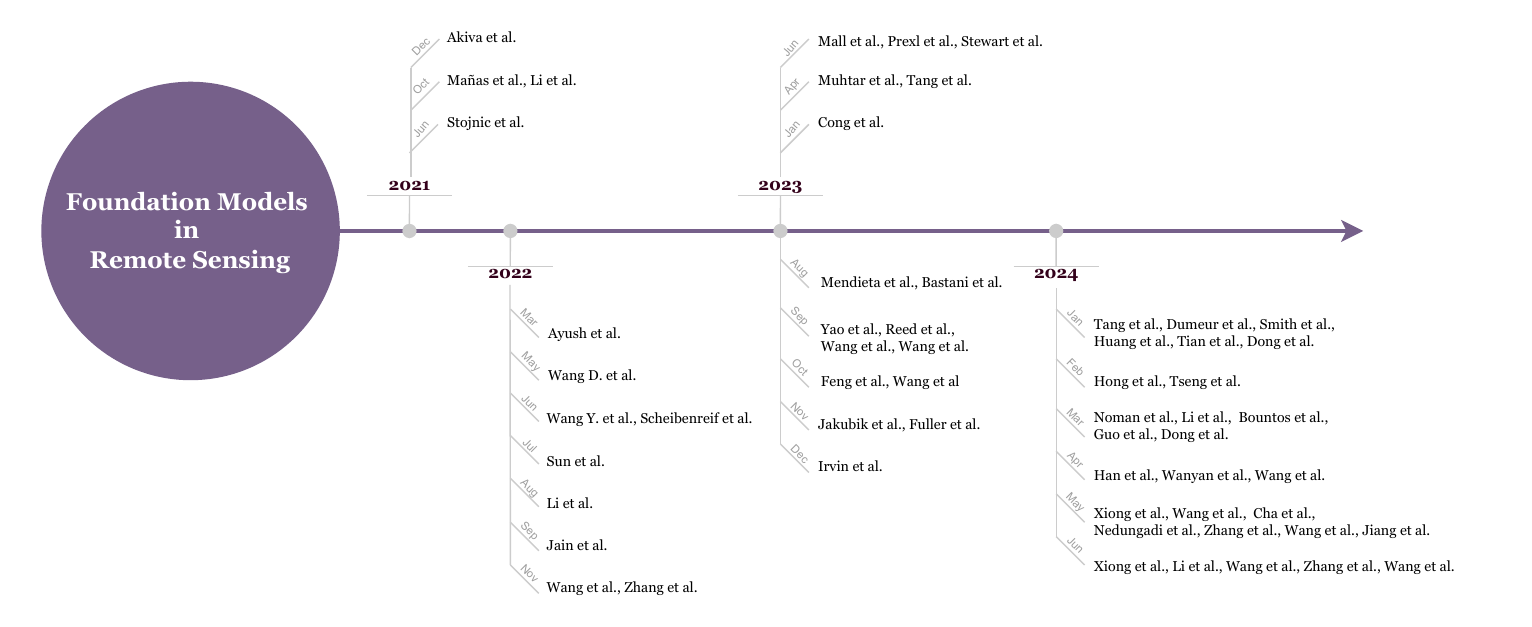}
    \caption{Overview of some well-known foundation models for remote sensing from 2021 June to 2024 June.}
    \label{fig:timeline}
\end{figure*}

This paper aims to provide a comprehensive survey of vision foundation models in the remote sensing domain ad rem, and is limited to foundation models released between June 2021 and June 2024. This timeframe marks a surge in the development of modern foundation models, including vision transformers and advanced self-supervised learning techniques. Although early models like Tile2Vec \cite{Tile2Vec} and others laid the groundwork for representation learning in remote sensing, they were typically limited in scale and generalization capabilities. Furthermore, numerous review papers have already provided comprehensive overviews of these pre-2021 models. Our review, therefore, focuses on recent developments to highlight the unique contributions and innovations that have emerged in the past few years.

In figure \ref{fig:timeline}, \textbf{58} vision foundation models are listed in chronological order. To facilitate navigation and enhance utility for researchers, we categorized existing models based on their perception levels (e.g., image-level, region-level, pixel-level). This organization helps clarify which models have been tested for general image-based challenges or specialized applications such as environmental monitoring, land cover mapping, archaeological exploration, disaster management, and more. It is essential to distinguish between applications that models have been explicitly tested on and those for which they could potentially be effective for. In this review, the fact that a model has not been tested on a particular application does not mean it won't perform well. Foundation models, especially convolutional neural network (CNN) backbones like residual networks (ResNet)\cite{resnet} and vision transformers (ViT) \cite{dosovitskiy2021imageworth16x16words}, may still be suitable for various downstream tasks, even if prior work has not yet demonstrated this.

Our contributions include:
\begin{enumerate}
    \item An exhaustive review of current state of vision foundation models proposed in the field of remote sensing, starting from the background and methodologies of these models to specific applications across different domains and tasks in a hierarchical and structured manner. 
    \item Categorization and analysis of the models based on their application in both image analysis (table \ref{tab:CV_task}) and practical applications (table \ref{tab:cv-domain}). We discuss the architecture, pre-training datasets, pre-training methods, and performance of each model.
    \item Discussion of challenges and unresolved aspects related to foundation models in remote sensing. We pinpoint new trends, raise important questions, and proposed future directions for further exploration. 
\end{enumerate}

\begin{figure*}[ht]
    \centering
    \includegraphics[width=\textwidth]{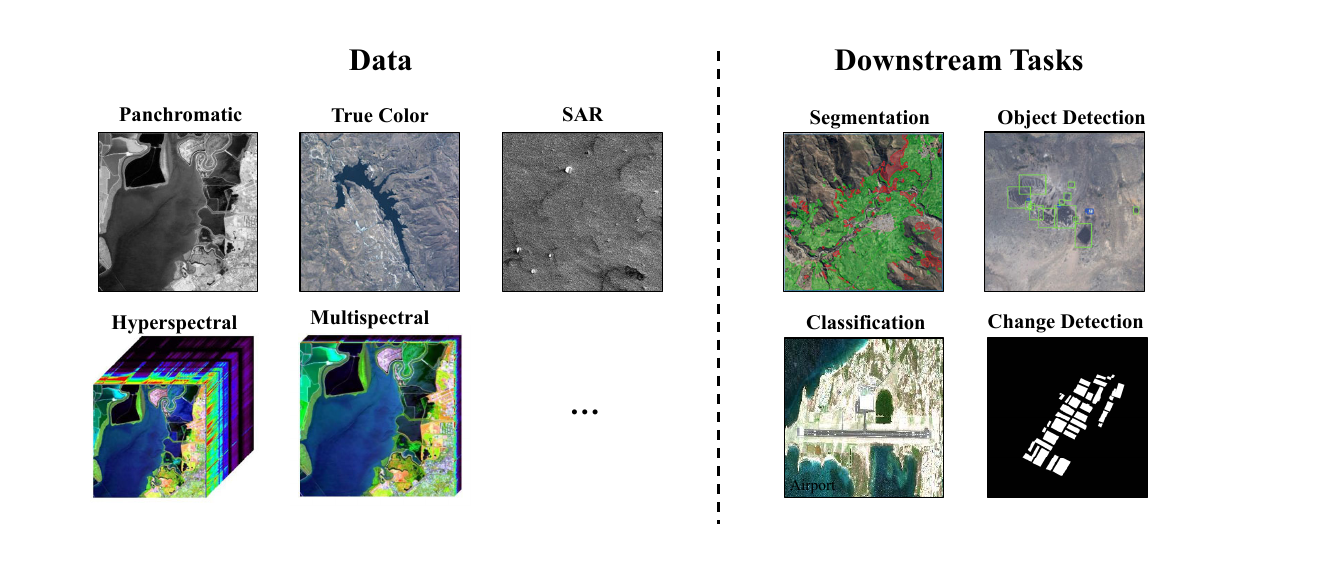}
    \caption{Examples of data types used in those foundation models and downstream tasks that can be done by foundation models. Data: (1) Panchromatic\cite{arbeck_english_2013}, (2) True Color, (3) SAR\cite{SAR}, (4) Hyperspectral\cite{arbeck_english_2013}, (5) Multispectral\cite{arbeck_english_2013}. Downstream tasks: (1) Segmentation, (2) Object Detection, (3) Classification\cite{fmow2018}, (4) Change Detection\cite{ChangeDetection}. \protect\footnotemark}
    \label{fig:data-task-example}
\end{figure*}

\section{Background} 
\subsection{Remote Sensing}

Remote sensing (RS) refers to the process of acquiring information about objects or areas from a distance, typically using satellite or airborne sensors. These technologies and techniques serve vital roles in diverse fields, enabling the collection of data over geographic areas without physical contact. Applications of remote sensing include earth observation, digital archaeology, urban planning and development, and disaster management. The field of remote sensing has developed rapidly since the mid 20th century. Initially, remote sensing predominately consisted of analog photographic techniques via aerial and satellite platforms, which provided limited spectral and spatial resolution. The launch of early Earth observation satellites, such as Landsat program commenced in 1967 \cite{Landset}, marked a significant advancement, enabling consistent and wide-ranging data collection for environmental monitoring. 

Modern remote sensing employs a variety of sensors suited for specific types of data collection, including optical, thermal, and radar.. Optical sensors capture a wide variety of spectral bands, including visible and near-infrared light, allowing for detailed imaging of land cover and vegetation health. Thermal sensors detect heat emitted or reflected from the Earth's surface, useful for monitoring volcanic activity, forest fires, and climate change monitoring. Radar sensors can penetrate clouds and vegetation, providing critical information in all-weather conditions and for applications such as soil moisture estimation and urban infrastructure mapping \cite{Navalgund2007, RSOverview}.

% \usepackage{graphicx}
\begin{table*}[]
\centering
\resizebox{\textwidth}{!}{%
\begin{tabular}{cc|
    >{\columncolor[HTML]{FFF2CC}}c |cccc|
    % >{\columncolor[HTML]{DCDCDC}}c c
    >{\columncolor[HTML]{DCDCDC}}c
    >{\columncolor[HTML]{DCDCDC}}c}

  \cellcolor[HTML]{D9E7FD}\textbf{Year-Month} &
  \cellcolor[HTML]{D9E7FD}\textbf{Architecture} &
  \cellcolor[HTML]{D9EAD3}\textbf{Model Name} &
  \cellcolor[HTML]{D9E7FD}\textbf{Image-Level} &
  \cellcolor[HTML]{D9E7FD}\textbf{Pixel-Level} &
  \cellcolor[HTML]{D9E7FD}\textbf{Region-Level} &
  \cellcolor[HTML]{D9E7FD}\textbf{Spatial-Temporal} &
  \cellcolor[HTML]{D9E7FD}\textbf{Contrastive Learning} &
  \cellcolor[HTML]{D9E7FD}\textbf{Predictive Coding} \\ \hline

\cellcolor{teal!0} 2021 Jun &
  ResNet-50 &
  CMC-RSSR \cite{CMC-RSSR} &
  $\checkmark$ &
   &
   &
   &
   $\checkmark$ &
   
   \\
\cellcolor{teal!1} 2021 Oct &
  ResNet-50 &
  SeCo \cite{SeCo} &
  $\checkmark$ &
   &
   &
  $\checkmark$ &
  & \\
\cellcolor{teal!2} 2021 Oct &
  ResNet-50 &
  GeoKR \cite{GeoKR} &
  $\checkmark$ &
  $\checkmark$ &
  $\checkmark$ &
  &
  &
   \\
\cellcolor{teal!3} 2021 Dec &
  ResNet-34 &
  MATTER \cite{MATTER}&
  $\checkmark$ &
  $\checkmark$ &
   &
  $\checkmark$ &
  &
  $\checkmark$ 
  \\
\cellcolor{teal!4} 2022 Mar &
  ResNet-50 &
  GASSL \cite{GASSL} &
  $\checkmark$ &
  $\checkmark$ &
  $\checkmark$ &
  &
  $\checkmark$ &
   \\
\cellcolor{teal!5} 2022 May &
  ViTAEv2-S &
  RSP \cite{RSP}& %Superviesed learning
  $\checkmark$ &
  $\checkmark$ &
  $\checkmark$ &
  $\checkmark$ &
  &\\
\cellcolor{teal!6} 2022 Jun &
  ViT-S/8 &
  DINO-MM \cite{DINO-MM}&
  $\checkmark$ &
   &
   &
   &
   $\checkmark$ &
   \\
\cellcolor{teal!7} 2022 Jun &
  Swin Transformer &
  Scheibenreif, et al. \cite{Scheibenreif_2022_CVPR}&
  $\checkmark$ &
  $\checkmark$ &
   &
   &
   $\checkmark$ &
   \\
\cellcolor{teal!8} 2022 Jul &
  ViT/Swin Transformer &
  RingMo \cite{RingMo}&
  $\checkmark$ &
  $\checkmark$ &
  $\checkmark$ &
  $\checkmark$ 
  &
  &
  $\checkmark$ \\
\cellcolor{teal!9} 2022 Aug &
  ResNet-50 &
  GeCO \cite{GeCo}&
  $\checkmark$ &
  $\checkmark$ &
  $\checkmark$ &
  &
  &
  $\checkmark$ \\   
\cellcolor{teal!10} 2022 Sep &
  BYOL &
  RS-BYOL \cite{RS-BYOL}&
  $\checkmark$ &
  $\checkmark$ &
   &
   &
  $\checkmark$ &
   \\
\cellcolor{teal!11} 2022 Nov &
  ViT-B &
  CSPT \cite{CSPT}&
  $\checkmark$ &
   &
  $\checkmark$ &
  &
  &
  $\checkmark$  \\
\cellcolor{teal!12} 2022 Nov &
  ViT &
  RVSA \cite{RVSA}&
  $\checkmark$ &
  $\checkmark$ &
  $\checkmark$ &
  &
  &
  $\checkmark$  \\
\cellcolor{teal!13} 2023 Jan &
  MAE-based Framework &
  SatMAE \cite{SatMAE}&
  $\checkmark$ &
  $\checkmark$ &
   &
   &
   &
  $\checkmark$  \\
\cellcolor{teal!14} 2023 Apr &
  TOV &
  TOV \cite{TOV}&
  $\checkmark$ &
  $\checkmark$ &
  $\checkmark$ &
  &
  &
  $\checkmark$ \\
\cellcolor{teal!15} 2023 Apr &
  \begin{tabular}[c]{@{}c@{}}Teacher-student\\ Self-distillation\end{tabular} &
  CMID \cite{CMID}&
  $\checkmark$ &
  $\checkmark$ &
  $\checkmark$ &
  $\checkmark$ 
  &
  & \\
\cellcolor{teal!16} 2023 Jun &
  CACo &
  CACo \cite{CACo}&
  $\checkmark$ &
  $\checkmark$ &
   &
  $\checkmark$ 
  &
  $\checkmark$ &\\ 
\cellcolor{teal!17} 2023 Jun &
  ResNet-18 &
  IaI-SimCLR \cite{IaI-SimCLR}&
  $\checkmark$ &
   &
   &
   &
   $\checkmark$ &
   \\
\cellcolor{teal!18} 2023 Jun &
  ResNet &
  SSL4EO-L \cite{SSL4EO-L} &
  &
  $\checkmark$ &
  &
  &
  $\checkmark$ &
  \\
\cellcolor{teal!19} 2023 Aug &
  Teacher-Student &
  GFM \cite{GFM}&
  $\checkmark$ &
  $\checkmark$ &
   &
  $\checkmark$ &
  &
  $\checkmark$ \\
\cellcolor{teal!20} 2023 Aug &
  Swim Transformer &
  SatLasPretrain \cite{SatlasPretrain}&
  $\checkmark$ &
  $\checkmark$ &
   &
   &
   &
   \\
\cellcolor{teal!21} 2023 Sep &
  Multi-Branch &
  RingMo-Sense \cite{RingMo-Sense}&
   &
  $\checkmark$ &
   &
   &
   & 
   $\checkmark$ \\
\cellcolor{teal!22} 2023 Sep &
  ViT &
  Scale-MAE \cite{Scale-MAE}&
  $\checkmark$ &
  $\checkmark$ &
   &
   &
   &
  $\checkmark$ \\
\cellcolor{teal!23} 2023 Sep &
  CNN-Transformer &
  RingMo-lite \cite{RingMo-lite}&
  $\checkmark$ &
  $\checkmark$ &
  $\checkmark$ &
  $\checkmark$ &
  &
  $\checkmark$ \\
\cellcolor{teal!24} 2023 Sep &
  Multimodel SSL &
  DeCUR \cite{DeCUR}&
  $\checkmark$ &
  $\checkmark$ &
   &
   &
   &
  $\checkmark$ \\
\cellcolor{teal!25} 2023 Oct &
  MSFE+MMFH &
  Feng et al. \cite{Feng2023}&
  $\checkmark$ &
  $\checkmark$ &
  $\checkmark$ &
  $\checkmark$ &
  &
  $\checkmark$ \\
\cellcolor{teal!26} 2023 Oct &
  ViT &
  FG-MAE \cite{FG-MAE}&
  $\checkmark$ &
  $\checkmark$ &
   &
   &
   &
  $\checkmark$  \\
\cellcolor{teal!27} 2023 Nov &
  ViT &
  Prithvi \cite{Prithvi}&
   &
  $\checkmark$ &
   &
   &
   &
  $\checkmark$  \\
\cellcolor{teal!28} 2023 Nov &
  Multimodal Encoder &
  CROMA \cite{CROMA}&
  $\checkmark$ &
  $\checkmark$ &
   &
   &
   $\checkmark$ &
   $\checkmark$ \\
\cellcolor{teal!29} 2023 Dec &
  ViT &
  USat \cite{USat}&
  $\checkmark$ &
   &
   &
   &
   &
  $\checkmark$  \\
\cellcolor{teal!30} 2024 Jan &
  ViT-B &
  Cross-Scale MAE \cite{Cross-Scale-MAE}&
  $\checkmark$ &
  $\checkmark$ &
   &
   &
   &
  $\checkmark$  \\
\cellcolor{teal!31} 2024 Jan &
  Unet+Transformer &
  U-BARN \cite{U-BARN}&
  $\checkmark$ &
  $\checkmark$ &
   &
   &
   &
   \\
\cellcolor{teal!32} 2024 Jan &
  \begin{tabular}[c]{@{}c@{}}Autoregressive\\ Transformer\end{tabular} &
  EarthPT \cite{EarthPT}&
  $\checkmark$ &
   &
   &
   &
   &
  $\checkmark$  \\
\cellcolor{teal!33} 2024 Jan &
  Teacher-Student Network &
  GeRSP \cite{GeRSP}&
  $\checkmark$ &
  $\checkmark$ &
  $\checkmark$ &
  &
  $\checkmark$ &
  $\checkmark$ \\
\cellcolor{teal!34} 2024 Jan &
  Dual-Branch &
  SwiMDiff \cite{SwiMDiff}&
  $\checkmark$ &
   &
   &
   &
  &
  \\
\cellcolor{teal!35} 2024 Jan &
  Generative ConvNet &
  SMLFR \cite{SMLFR}&
   &
  $\checkmark$ &
  $\checkmark$ &
  &
  &
  $\checkmark$ \\
\cellcolor{teal!36} 2024 Feb &
  3D GPT &
  SpectralGPT \cite{SpectralGPT}&
  $\checkmark$ &
  $\checkmark$ &
   &
  $\checkmark$ &
  &
  $\checkmark$ \\
\cellcolor{teal!37} 2024 Feb &
  MAE-based Framework &
  Presto \cite{Presto}&
   &
  $\checkmark$ &
   &
   &
  $\checkmark$  &
  $\checkmark$  \\
\cellcolor{teal!38} 2024 Mar &
  SatMAE &
  SatMAE++ \cite{SatMAE++}&
  $\checkmark$ &
   &
   &
   &
   &
  $\checkmark$  \\
\cellcolor{teal!39} 2024 Mar &
  \begin{tabular}[c]{@{}c@{}}Joint-Embedding \\ Predictive Architecture\end{tabular} &
  SAR-JEPA \cite{SAR-JEPA}&
  $\checkmark$ &
   &
   &
   &
   &
  $\checkmark$  \\
\cellcolor{teal!40} 2024 Mar &
  ViT &
  FoMo-Bench \cite{FoMo-Bench}&
  $\checkmark$ &
  $\checkmark$ &
  $\checkmark$ &
  &
  &
  $\checkmark$ \\
\cellcolor{teal!41} 2024 Mar &
  \begin{tabular}[c]{@{}c@{}}Factorized Multi-Modal\\ Spatiotemporal Encoder\end{tabular} &
  SkySense \cite{SkySense}&
  $\checkmark$ &
  $\checkmark$ &
  $\checkmark$ &
  $\checkmark$ &
  &
  $\checkmark$ \\
\cellcolor{teal!42} 2024 Mar &
  Multi-Modules &
  UPetu \cite{UPetu}&
  $\checkmark$ &
  $\checkmark$ &
   &
  $\checkmark$ &
  &
  $\checkmark$ \\
\cellcolor{teal!43} 2024 Apr &
  Swim Transformer &
  msGFM \cite{msGFM}&
  $\checkmark$ &
  $\checkmark$ &
   &
   &
   &
  $\checkmark$  \\
% \cellcolor{teal!44} 2024 Apr &
%   ViT &
%   ORBIT \cite{ORBIT}&
%    &
%    &
%    &
%    \\

\cellcolor{teal!45} 2024 Apr &
  DINO &
  DINO-MC \cite{DINO-MC}&
  $\checkmark$ &
   &
   &
  $\checkmark$ &
  $\checkmark$ &
  \\
\cellcolor{teal!46} 2024 May &
  OFA-Net &
  OFA-Net \cite{OFA-Net}&
  $\checkmark$ &
  $\checkmark$ &
   &
   &
   &
  $\checkmark$  \\
\cellcolor{teal!47} 2024 May &
  \begin{tabular}[c]{@{}c@{}}Shared Encoder,\\ Task-Specific Decoders\end{tabular} &
  MTP \cite{MTP}&
  $\checkmark$ &
  $\checkmark$ &
  $\checkmark$ &
  $\checkmark$ &
  &\\
\cellcolor{teal!48} 2024 May &
  ViT &
  BFM \cite{BFM}&
   &
  $\checkmark$ &
  $\checkmark$ &
   &
   &
  $\checkmark$  \\
\cellcolor{teal!49} 2024 May &
  MP-MAE &
  MMEarth \cite{MMEarth}&
  $\checkmark$ &
  $\checkmark$ &
   &
   &
   &
  $\checkmark$  \\
\cellcolor{teal!50} 2024 May &
  ViT &
  CtxMIM \cite{CtxMIM}&
  $\checkmark$ &
  $\checkmark$ &
  $\checkmark$ &
  &
  &
  $\checkmark$  \\
\cellcolor{teal!51} 2024 May &
  HiViT &
  SARATR-X \cite{SARATR-X}&
  $\checkmark$ &
   &
  $\checkmark$ &
  &
  &
  $\checkmark$ \\
\cellcolor{teal!52} 2024 May &
  Transformer &
  SoftCon \cite{SoftCon} &
  $\checkmark$ &
  $\checkmark$ &
  &
  $\checkmark$ &
  $\checkmark$ &
  \\
\cellcolor{teal!53} 2024 May &
  ViT &
  LeMeViT \cite{LeMeViT} &
  &
  $\checkmark$ &
  $\checkmark$ &
  $\checkmark$ &
  &
  \\
\cellcolor{teal!54} 2024 Jun &
  Masked Autoencoder &
  S2MAE \cite{S2MAE} &
  $\checkmark$ &
  &
  &
  $\checkmark$ &
  &
  $\checkmark$ \\
\cellcolor{teal!55} 2024 Jun &
  CNN - Transformer &
  RS-DFM \cite{RS-DFM} &
  &
  $\checkmark$ &
  $\checkmark$ &
  &
  &
  \\
\cellcolor{teal!56} 2024 Jun &
  MAE-based &
  A2-MAE \cite{A2-MAE} &
  $\checkmark$ &
  $\checkmark$ &
  &
  $\checkmark$ &
  &
  \\
\cellcolor{teal!57} 2024 Jun &
  ViT &
  HyperSIGMA \cite{HyperSIGMA} &
  $\checkmark$ &
  $\checkmark$ &
  $\checkmark$ &
  $\checkmark$ &
  &
  $\checkmark$ \\
\cellcolor{teal!58} 2024 Jun &
  Dynamic OFA &
  DOFA \cite{DOFA}&
  $\checkmark$ &
  $\checkmark$ &
   &
   &
   &
  $\checkmark$  \\

\end{tabular}%
}
\caption{Summary of the pretraining methods utilized and image analysis tasks evaluated across different models. Image-level, pixel-level, region-level, and spatial-temporal classify the tasks in image analysis, while contrastive learning and predictive coding indicates the different self-supervised pretraining strategies that each study used.}
\label{tab:CV_task}
\end{table*}

In recent years, remote sensing has found applications in many fields. With regard to environmental monitoring, it is used to track deforestation, to monitor air and water quality, and to assess the impacts of climate change \cite{Himeur2022, RSObjectDetectInDL}. In agriculture, remote sensing helps in crop health monitoring, yield estimation, and efficient resource management \cite{Navalgund2007}. Urban planning and development benefit from remote sensing through the monitoring of urban sprawl, infrastructure development, and land-use planning \cite{Jha2021, RSOverview}. Furthermore, in disaster management, remote sensing is crucial for assessing the damage caused by natural disasters, aiding in the planning and execution of relief operations \cite{Abid2021, RSObjectDetectInDL}.

The integration of remote sensing data with Geographic Information Systems (GIS) has further enhanced its utility. GIS provides a framework for capturing, storing, analyzing, and visualizing spatial and geographic data. When combined with remote sensing data, GIS can be used to create detailed and dynamic maps and models for various applications. This synergy is particularly valuable in resource management, urban planning, and disaster response, where accurate and timely information is critical \cite{Navalgund2007, RSOverview, RSObjectDetectInDL}.

\footnotetext{True Color, Segmentation, and Object detection images © MAXAR 2024, provided through the NextView License Agreement.}

\subsection{Foundation Models for Remote Sensing}
Foundation models (FMs) refer to large-scale, pre-trained models that provide a robust starting point for various downstream tasks across different domains \cite{Jiao2023}. These models leverage extensive datasets and advanced architectures, enabling them to capture complex patterns and features that can be fine-tuned for specific applications with minimal additional training. In remote sensing, FMs are particularly valuable due to the diverse and complex nature of the data, including multi-spectral and multi-temporal imagery. Techniques such as self-supervised learning (SSL) \cite{jing2019selfsupervised} and transformers \cite{vaswani2023attention} have significantly enhanced the performance and efficiency of tasks such as image classification, object detection, and change detection, addressing the unique challenges posed by remote sensing data \cite{Dias2023}.

A major strength of these models lies in their ability to utilize SSL to learn effective representations from largely unlabeled data, which is often abundant in remote sensing scenarios \cite{zhou2023comprehensive}. By integrating advanced architectures like transformers \cite{vaswani2023attention}, FMs in remote sensing can handle the unique characteristics of geospatial data, such as varying spatial resolutions and temporal dynamics, without requiring separate task-specific models.

The evolution of FMs has been driven by advancements in deep learning and the availability of large datasets. Initially, convolutional neural networks (CNNs) like ResNet \cite{he2015deep} paved the way for improved image recognition and classification tasks \cite{Ma2024}. The introduction of transformers, which use self-attention mechanisms to model long-range dependencies, has further advanced the capabilities of FMs in handling large-scale image data \cite{SatMAE}. Vision transformers (ViTs) \cite{dosovitskiy2021imageworth16x16words} extend the transformer architecture to process image data by treating image patches as sequences of tokens, enabling models to learn both local and global relationships. This capability makes transformers particularly effective for semantic segmentation and change detection tasks, where capturing long-range dependencies is crucial, especially in high-resolution satellite imagery.

Notable foundation models in remote sensing include SatMAE \cite{SatMAE}, which pre-trains transformers for temporal and multi-spectral satellite imagery; Scale-MAE\cite{Scale-MAE}, a scale-aware masked autoencoder for multiscale geospatial representation learning; and DINO-MC \cite{DINO-MC}, which extends global-local view alignment for SSL with remote sensing imagery. These models have shown remarkable performance in various remote sensing tasks such as scene classification, object detection, and change detection.

Despite their success, FMs face several challenges, including the need for high-quality and diverse training data, significant computational resources, and effective domain adaptation to specific remote sensing tasks \cite{SatMAE++}. Addressing these challenges will be crucial for the continued advancement of FMs in remote sensing.
\section{Related Review Papers} \label{sec: related work}

Artificial intelligence in remote sensing has been a growing area of research, with numerous review papers providing insights into AI advancements and their applications. In this section, we summarize the most influential reviews on foundation models in remote sensing.

Zhang et al. (2016), in their foundational review "\textit{Deep Learning for Remote Sensing Data: A Technical Tutorial on the State of the Art}" \cite{Zhang2016SOA}, introduced deep learning techniques to RS, focusing on convolutional neural networks (CNNs) for tasks such as image classification and object detection. This work highlighted both the promise and challenges of early AI integration in RS, setting the stage for subsequent advancements.

In 2017, Zhu et al.’s "\textit{Deep Learning in Remote Sensing: A Comprehensive Review and List of Resources}" \cite{Zhu2017DLReview} delved into diverse AI applications, including hyperspectral analysis and synthetic aperture radar (SAR) interpretation. It also provided an extensive resource list, capturing the rapid adoption of deep learning in addressing complex RS challenges, paving the way for more advanced AI models in the following years.

More recent reviews have focused on advanced AI models and methods. Wang et al.'s 2022 review, "\textit{Self-Supervised Learning in Remote Sensing}" \cite{SSL}, highlighted the ability of self-supervised learning (SSL) methods to utilize large volumes of unlabeled data, significantly reducing dependence on labeled datasets while maintaining high performance in RS tasks. The review also identified key challenges and future directions, emphasizing SSL’s potential to handle large-scale RS data complexities.

Zhang et al. (2022), in "\textit{Artificial Intelligence for Remote Sensing Data Analysis: A Review of Challenges and Opportunities}" \cite{Zhang2022AIReview}, offered a comprehensive overview of AI algorithms, synthesizing findings from over 270 studies. It emphasized ongoing challenges such as explainability, security, and integrating AI with other computational techniques, serving as a roadmap for future innovation in AI-driven RS.

Aleissaee et al.'s 2023 survey, "\textit{Transformers in Remote Sensing}" \cite{aleissaee2022transformersremotesensingsurvey}, explored the impact of transformer-based models across various RS tasks, comparing them with CNNs. It identified both strengths and limitations, along with unresolved challenges, providing a detailed roadmap for future research on transformers' role in RS.

Li et al.'s 2024 review, "\textit{Vision-Language Models in Remote Sensing}" \cite{Li2024vlreview}, examined the increasing significance of vision-language models (VLMs), which combine visual and textual data. It highlighted VLMs' potential in applications like image captioning and visual question answering, emphasizing a shift toward richer semantic understanding in RS tasks.

Additionally, the recent work, "\textit{On the Foundations of Earth and Climate Foundation Models}" \cite{zhu2024foundationsearthclimatefoundation}, provided a comprehensive review of existing foundation models, proposing features like geolocation embedding and multisensory capability. It outlined key traits for future Earth and climate models, contributing to a broader discussion on foundational advancements in geospatial AI.

Building on these reviews, our study provides a comprehensive analysis of foundation models developed from June 2021 to June 2024, focusing on advances in self-supervised learning and transformer-based architectures. Unlike previous reviews, which focused mainly on individual techniques, we explore their combined potential in remote sensing tasks like semantic segmentation, multi-spectral analysis, and change detection. For instance, SatMAE \cite{SatMAE} demonstrates effective use of SSL for pre-training transformers, enabling improved segmentation in complex multi-spectral imagery, while Scale-MAE employs scale-aware masked autoencoders for better handling of varied spatial resolutions in remote sensing data.

Our study also highlights new models like DINO-MC \cite{DINO-MC}, which integrates global-local view alignment for SSL, making it particularly effective for identifying changes in high-resolution satellite imagery. By systematically examining these innovations, we illustrate how recent models address persistent challenges like domain adaptation and computational efficiency. For example, efficient self-attention mechanisms in Scale-MAE\cite{Scale-MAE} help reduce computation costs, while enhanced geolocation embeddings in models like SatMAE improve performance in geospatial feature extraction.

In contrast to earlier reviews, which often remained theoretical, we emphasize both the theoretical advancements and practical applications of recent models. For example, DINO-MC \cite{DINO-MC} and ORBIT’s \cite{ORBIT} real-world applications in environmental monitoring and disaster response highlight its practical impact, demonstrating how new FMs can be effectively leveraged to address pressing challenges in geospatial analysis.

\section{Pretraining Methods}
\label{pretrain_methods}

\begin{figure*}[h]
    \centering
    \includegraphics[width=\textwidth]{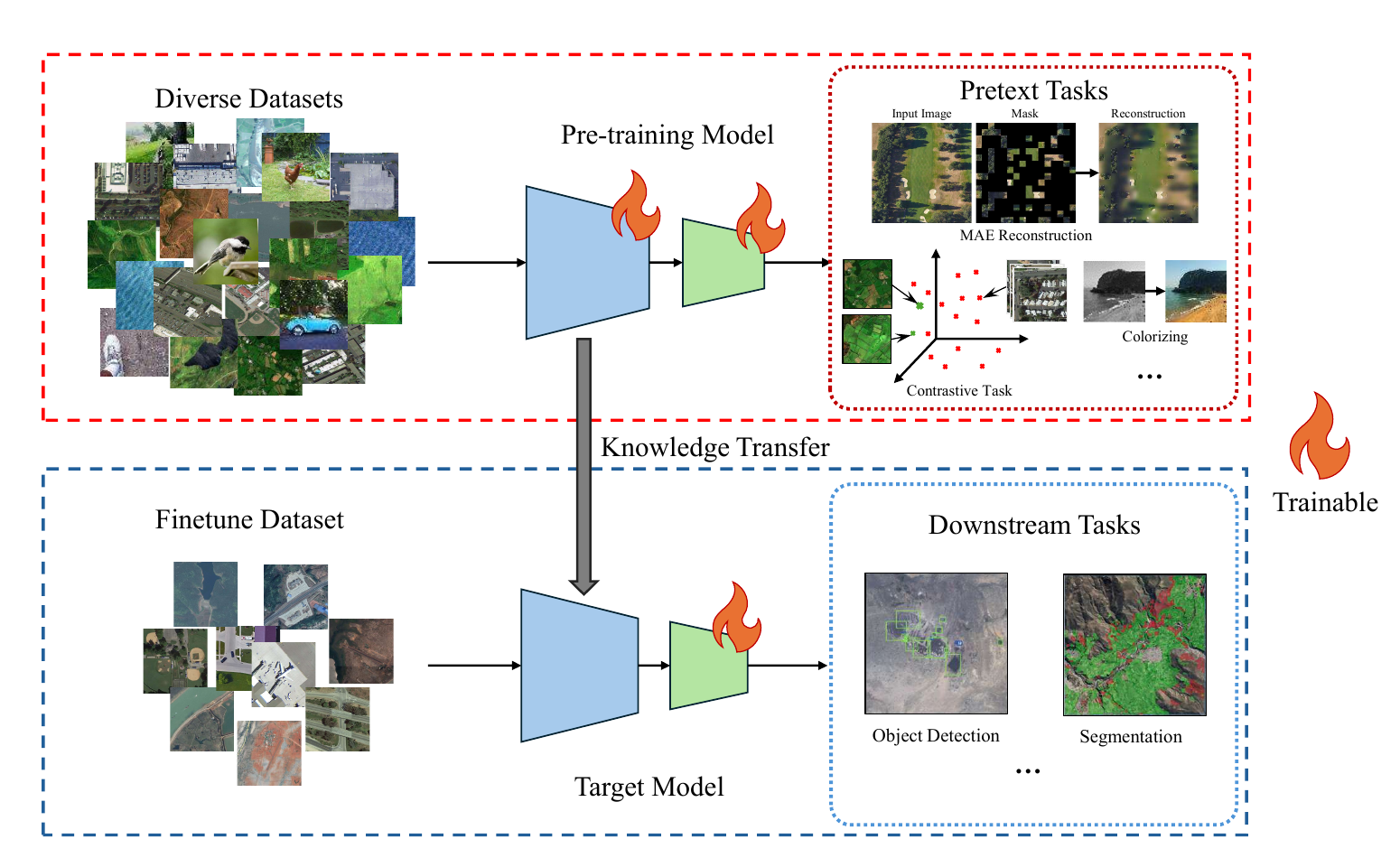}
    \caption{General pipeline of SSL\cite{jing2019selfsupervised}. Diverse datasets images and pretext task images are acquired from ImageNet\cite{imageNet}, BigEarthNet\cite{Sumbul_2019_Bigearthnet}, and MillionAID\cite{Long2022ASP_millionaid}. Finetune dataset includes images from DIOR\cite{Li_2020_DIOR}. \protect\footnotemark}
    \label{fig:ssl}
\end{figure*}

Pretraining serves as a critical step in developing foundation models (FM), enabling them to learn transferable and generalized representations from large-scale datasets. This process leverage self-supervised or supervised learning methods to extract domain-agnostic features that can be adapted to various downstream tasks. In this section, we explore the key pretraining methods utilized commonly in foundation models for remote sensing, explaining the mechanism of these methods and their roles in enhancing model performance and addressing challenges in this field.

\subsection{Self-Supervised Learning}
Self-supervised learning has emerged as a cornerstone of pre-training foundation models, offering a paradigm where models learn representations by predicting parts of the input data from other parts. This approach reduces reliance on expensive and time-consuming labeled datasets, making it particularly advantageous in fields like remote sensing, where labeled data is often scarce or challenging to obtain.

SSL allows models to exploit vast amounts of unlabeled data, learning rich, generalizable representations that transfer well to downstream tasks such as scene classification, semantic segmentation, object detection, and change detection. By uncovering underlying data structures and patterns, SSL not only enhances model robustness but also improves adaptability across diverse domains and resolutions of remote sensing imagery \cite{SSL_remotesensing}. Figure \ref{fig:ssl} illustrates the general pipeline of self-supervised learning. 

Two SSL methods commonly used in vision foundation models for remote sensing are contrastive learning and predictive coding, each offering unique mechanisms to harness information from unlabeled data.

    \subsubsection{Predictive Coding}

   Predictive coding leverages a generative approach, where the model learns to predict missing or occluded parts of an image based on visible portions. This strategy helps capture spatial and contextual relationships in remote sensing imagery, which often contains diverse textures, complex scenes, and varying resolutions.

   \footnotetext{Object detection and Segmentation © MAXAR 2024, provided through the NextView License Agreement.}
    
    In remote sensing, predictive coding can be applied to tasks such as gap filling in satellite imagery, where the model learns to infer missing data caused by sensor limitations or occlusions like cloud cover. Popular implementations of predictive coding frameworks include autoencoder-based architectures, masked image modeling techniques like those used in MAE (Masked Autoencoders) \cite{MAE}, and autoregressive models. These methods are particularly effective in learning fine-grained details critical for high-resolution imagery and specialized tasks.

    \subsubsection{Contrastive Learning}

   Contrastive learning is another powerful SSL technique that focuses on distinguishing between similar and dissimilar samples in the data. The key idea is to bring representations of similar (positive) samples closer together while pushing apart those of dissimilar (negative) samples. This encourages the model to learn discriminative and invariant features that are crucial for remote sensing tasks.
    
    Contrastive learning frameworks such as SimCLR \cite{SimCLR}, MoCo \cite{MoCo}, DINO \cite{DINO}, and BYOL \cite{BYOL} have shown promise in remote sensing applications. They use augmentations like random cropping, rotations, or spectral band dropping to generate positive pairs, enabling the model to learn robust representations invariant to these transformations. For instance, in multispectral or hyperspectral imagery, contrastive learning can help models capture spectral signatures across varying conditions, improving performance in tasks like crop classification or land cover mapping \cite{SSL_remotesensing}.
    
    Contrastive learning is especially relevant in remote sensing when labeled datasets are highly imbalanced, as it enables models to learn from underrepresented classes or regions without explicit labels.

By combining approaches like predictive coding and contrastive learning, self-supervised learning has significantly advanced the development of vision foundation models in remote sensing. These methods allow models to leverage vast, unlabeled datasets while maintaining adaptability across diverse spatial resolutions, spectral bands, and application scenarios. On the other hand, it is important to note that there are many other SSL methods that can be employed to such tasks. Other innovative methods, such as teacher-student self-distillation frameworks, have also demonstrated potential in remote sensing applications. For example, CMID \cite{CMID} achieves promising performance by combining contrastive learning and masked image modeling in a teacher-student self-distillation framework. This structure enables it to capture both global and local features, making it effective for diverse remote sensing tasks.. The diversity of SSL techniques highlights the versatility and evolving nature of self-supervised learning, underscoring its critical role in unlocking the full potential of remote sensing imagery.

\subsection{Supervised Pretraining}
Supervised pretraining is a fundamental approach in deep learning, where models are trained using labeled datasets to minimize prediction errors for specific tasks, such as image classification. This method allows models to learn direct mappings between input features and target labels, fostering the development of detailed, task-specific representations. For instance, models like ResNet \cite{resnet} and VGGNet (Visual Geometry Group Network) \cite{VGGNet} trained on large-scale datasets such as ImageNet \cite{imageNet} have demonstrated how supervised pretraining can capture robust feature hierarchies that are highly transferable to related tasks, including semantic segmentation or object detection.

In remote sensing, supervised pretraining has shown promise for tasks such as land cover classification and object detection using high-resolution satellite imagery \cite{SL_wang2023}. However, the dependency on large-scale labeled datasets presents a major limitation. Creating labeled datasets for remote sensing tasks, particularly when involving multispectral or hyperspectral data, is resource-intensive and often requires domain expertise for annotation. For example, labeling pixel-level data for land cover classification or delineating objects in complex urban environments can be prohibitively time-consuming. Furthermore, labeled data in remote sensing is often domain-specific, limiting the generalizability of models trained on one dataset to other applications or regions \cite{Zhu2017DLReview}.

These challenges highlight the need for innovative strategies to address the reliance on labeled data. Such limitations have motivated the development of alternative approaches, including self-supervised pretraining methods, which leverage the abundance of unlabeled data to learn general-purpose representations without manual annotation.
\section{Image Analysis Methods}
\label{image_analysis_methods}

\subsection{Image Perception at Different Levels}
Foundation models in remote sensing enable image analysis at three primary levels: image-level, region-level, and pixel-level. These levels address varying spatial, contextual, and application-specific needs, providing the foundation for a wide range of tasks such as environmental monitoring, urban planning, disaster response, and more. The following subsections outline the distinct objectives and applications at each level. A detailed summary of the models evaluated for these tasks is provided in table \ref{tab:cv-task}. The following subsections outline the distinct objectives and applications at each level.

    \subsubsection{Image-Level}
Image-level analysis focuses on classification tasks, categorizing entire images or large image segments into predefined classes, such as urban, forest, water bodies, or agricultural areas. This approach provides broad, high-level insights into geographic regions and is instrumental in large-scale applications like land use mapping, land cover classification, and resource management. By classifying entire scenes, this level of analysis enables efficient monitoring of extensive areas, supporting decision-making in environmental management and policy planning.

    \subsubsection{Region-Level}
Region-level analysis identifies and localizes specific objects within an image, such as buildings, vehicles, ships, or other structures. Unlike image-level analysis, which provides holistic classifications, region-level tasks focus on object detection which is to detect individual entities and their spatial locations. This analysis is critical for targeted applications like urban planning, where the detection of infrastructure is essential, as well as disaster response and security, where identifying damaged buildings or vulnerable areas can significantly aid in timely interventions.
    
    \onecolumn
\begin{landscape}

\setlength\LTleft{-3cm}  % Change from 1pt to move table to center
\setlength\LTright{-3cm} % Change from 1pt to move table to center
\addtolength{\LTleft}{0pt plus 1fill}    % Add these two lines
\addtolength{\LTright}{0pt plus 1fill}   % to center the table

\fontsize{8}{10}\selectfont  % Smaller font size

\begin{longtable}{|cll>{\columncolor[HTML]{EFEFEF}}cl>{\columncolor[HTML]{EFEFEF}}lc>{\columncolor[HTML]{F3F3F3}}c|}
% Caption
\caption{Overview of recent foundation models in remote sensing, categorized by architecture, model name, pre-training dataset, resolution, geographic coverage, image analysis lebels, visual encoder, pre-training methods, and the number of parameters. \textbf{Abbreviations for pre-training methods as specified in the original work:} SSL refers to Self-Supervised Learning, CL stands for Contrastive Learning, MIM is Masked Image Modeling, MAE is Masked Autoencoders, and FD-MIM is Feature-Distilled Masked Image Modeling.} \label{tab:cv-task} \\

% Header first page
\hline
\multicolumn{1}{|c}{\cellcolor[HTML]{D9EAD3}\centering\textbf{Model Name}} &
\multicolumn{1}{c}{\cellcolor[HTML]{D9E7FD}\centering\textbf{Architecture}} &
\multicolumn{1}{c}{\cellcolor[HTML]{D9E7FD}\centering\textbf{Pre-training Dataset}} &
\multicolumn{1}{c}{\cellcolor[HTML]{D9E7FD}\centering\textbf{Resolution (m)}} &
\multicolumn{1}{c}{\cellcolor[HTML]{D9E7FD}\centering\textbf{Geographic Coverage}} &
\multicolumn{1}{c}{\cellcolor[HTML]{D9E7FD}\centering\textbf{\begin{tabular}[c]{@{}c@{}} Image Analysis \\ Levels \end{tabular}}} &
\multicolumn{1}{c}{\cellcolor[HTML]{D9E7FD}\centering\textbf{Pretrain methods}} &
\multicolumn{1}{c|}{\cellcolor[HTML]{D9E7FD}\centering\textbf{\# of Params}} \\ \hline
\endfirsthead

% Header for subsequent pages
\multicolumn{8}{c}{\tablename\ \thetable{} -- continued from previous page} \\
\hline
\multicolumn{1}{|c}{\cellcolor[HTML]{D9EAD3}\centering\textbf{Model Name}} &
\multicolumn{1}{c}{\cellcolor[HTML]{D9E7FD}\centering\textbf{Architecture}} &
\multicolumn{1}{c}{\cellcolor[HTML]{D9E7FD}\centering\textbf{Pre-training Dataset}} &
\multicolumn{1}{c}{\cellcolor[HTML]{D9E7FD}\centering\textbf{Resolution (m)}} &
\multicolumn{1}{c}{\cellcolor[HTML]{D9E7FD}\centering\textbf{Geographic Coverage}} &
\multicolumn{1}{c}{\cellcolor[HTML]{D9E7FD}\centering\textbf{Image Analysis Levels}} &
\multicolumn{1}{c}{\cellcolor[HTML]{D9E7FD}\centering\textbf{Pretrain methods}} &
\multicolumn{1}{c|}{\cellcolor[HTML]{D9E7FD}\centering\textbf{\# of Params}} \\ \hline
\endhead

% Footer for all pages except the last
\hline
\multicolumn{8}{r}{\textit{Continued on next page}} \\
\endfoot

% Footer for last page
\hline
\endlastfoot

% CMC-RSSR
\multicolumn{1}{|c}{\cellcolor[HTML]{FFF2CC}CMC-RSSR \cite{CMC-RSSR}} &
\multicolumn{1}{c}{ResNet-50} &
\multicolumn{1}{c}{\cellcolor[HTML]{EFEFEF}\begin{tabular}[c]{@{}c@{}}NWPU-DOTA\cite{xia2019dotalargescaledatasetobject},\\ BigEarthNet \cite{Sumbul_2019_Bigearthnet}, \\ImageNet \cite{imageNet} \end{tabular}} &
\multicolumn{1}{c}{0.2 to 60} &
\multicolumn{1}{c}{\cellcolor[HTML]{EFEFEF}Global} &
\multicolumn{1}{c}{\begin{tabular}[c]{@{}c@{}}Image-level\end{tabular}} &
\multicolumn{1}{c}{\cellcolor[HTML]{EFEFEF}\begin{tabular}[c]{@{}c@{}}Contrastive \\ Multiview Coding\end{tabular}} &
\multicolumn{1}{c|}{23M} \\ 

% SeCo
\multicolumn{1}{|c}{\cellcolor[HTML]{FFF2CC}SeCo \cite{SeCo}} &
  \multicolumn{1}{c}{ResNet-50} &
  \multicolumn{1}{c}{\cellcolor[HTML]{EFEFEF}Sentinel-2 Imagery} &
  \multicolumn{1}{c}{10, 20, 60} &
  \multicolumn{1}{c}{\cellcolor[HTML]{EFEFEF}\begin{tabular}[c]{@{}c@{}}200k Locations \\ Worldwide\end{tabular}} &
  \multicolumn{1}{c}{\begin{tabular}[c]{@{}c@{}}Image-level, \\ Spacial-temporal\end{tabular}} &
  \multicolumn{1}{c}{\cellcolor[HTML]{EFEFEF}CL} &
  \multicolumn{1}{c|}{23.5M} \\

% GeoKR
\multicolumn{1}{|c}{\cellcolor[HTML]{FFF2CC}GeoKR \cite{GeoKR}} &
  \multicolumn{1}{c}{ResNet-50} &
  \multicolumn{1}{c}{\cellcolor[HTML]{EFEFEF}Levir-KR \cite{GeoKR}} &
  \multicolumn{1}{c}{0.8 to 16} &
  \multicolumn{1}{c}{\cellcolor[HTML]{EFEFEF}Global} &
  \multicolumn{1}{c}{\begin{tabular}[c]{@{}c@{}}Image-level, Pixel-level, \\ Region-level\end{tabular}} &
  \multicolumn{1}{c}{\cellcolor[HTML]{EFEFEF}\begin{tabular}[c]{@{}c@{}}Geographical Knowledge\\ Supervision\end{tabular}} &
  \multicolumn{1}{c|}{23.5M/138M} \\

% MATTER
\multicolumn{1}{|c}{\cellcolor[HTML]{FFF2CC}MATTER \cite{MATTER}} &
  \multicolumn{1}{c}{ResNet-34} &
  \multicolumn{1}{c}{\cellcolor[HTML]{EFEFEF}Sentinel-2 Imagery} &
  \multicolumn{1}{c}{-} &
  \multicolumn{1}{c}{\cellcolor[HTML]{EFEFEF}\begin{tabular}[c]{@{}c@{}}Rural and Remote Regions\\ with Little Changes\end{tabular}} &
  \multicolumn{1}{c}{\begin{tabular}[c]{@{}c@{}}Image-level, \\Pixel-level\end{tabular}} &
  \multicolumn{1}{c}{\cellcolor[HTML]{EFEFEF}SSL} &
  \multicolumn{1}{c|}{21.3M} \\

% GASSL
\multicolumn{1}{|c}{\cellcolor[HTML]{FFF2CC}GASSL \cite{GASSL}} &
  \multicolumn{1}{c}{ResNet-50} &
  \multicolumn{1}{c}{\cellcolor[HTML]{EFEFEF}fMoW \cite{fmow2018}, GeoImageNet \cite{imageNet}} &
  \multicolumn{1}{c}{-} &
  \multicolumn{1}{c}{\cellcolor[HTML]{EFEFEF}7 Continents} &
  \multicolumn{1}{c}{\begin{tabular}[c]{@{}c@{}}Image-level, Pixel-level \\ Region-level\end{tabular}} &
  \multicolumn{1}{c}{\cellcolor[HTML]{EFEFEF}CL} &
  \multicolumn{1}{c|}{23.5M} \\

% RSP
\multicolumn{1}{|c}{\cellcolor[HTML]{FFF2CC}RSP \cite{RSP}} &
  \multicolumn{1}{c}{ViTAEv2-S} &
  \multicolumn{1}{c}{\cellcolor[HTML]{EFEFEF}MillionAID \cite{Long2022ASP_millionaid, Long2021DiRS_millionaid}} &
  \multicolumn{1}{c}{0.5 to 153} &
  \multicolumn{1}{c}{\cellcolor[HTML]{EFEFEF}Global} &
  \multicolumn{1}{c}{\begin{tabular}[c]{@{}c@{}}Image-level, Pixel-level, \\ Region-level, Spacial-temporal\end{tabular}} &
  \multicolumn{1}{c}{\cellcolor[HTML]{EFEFEF}Supervised Learning} &
  \multicolumn{1}{c|}{24.8M/23.5M/29M} \\

% DINO-MM
\multicolumn{1}{|c}{\cellcolor[HTML]{FFF2CC}DINO-MM \cite{DINO-MM}} &
  \multicolumn{1}{c}{ViT-S/8} &
  \multicolumn{1}{c}{\cellcolor[HTML]{EFEFEF}BigEarthNet-MM \cite{BigEarthNet-MM}} &
  \multicolumn{1}{c}{10} &
  \multicolumn{1}{c}{\cellcolor[HTML]{EFEFEF}Global} &
  \multicolumn{1}{c}{Image-level} &
  \multicolumn{1}{c}{\cellcolor[HTML]{EFEFEF}SSL} &
  \multicolumn{1}{c|}{22M} \\

% Scheibenreif, et al.
\multicolumn{1}{|c}{\cellcolor[HTML]{FFF2CC}Scheibenreif, et al. \cite{Scheibenreif_2022_CVPR}} &
  \multicolumn{1}{c}{Swin Transformer} &
  \multicolumn{1}{c}{\cellcolor[HTML]{EFEFEF}SEN12MS \cite{schmitt2019sen12mscurateddataset}} &
  \multicolumn{1}{c}{10} &
  \multicolumn{1}{c}{\cellcolor[HTML]{EFEFEF}Global} &
  \multicolumn{1}{c}{\begin{tabular}[c]{@{}c@{}}Image-level, \\ Pixel-level\end{tabular}} &
  \multicolumn{1}{c}{\cellcolor[HTML]{EFEFEF}CL} &
  \multicolumn{1}{c|}{-} \\

% RingMo
\multicolumn{1}{|c}{\cellcolor[HTML]{FFF2CC}RingMo \cite{RingMo}} &
  \multicolumn{1}{c}{ViT/Swin Transformer} &
  \multicolumn{1}{c}{\cellcolor[HTML]{EFEFEF}2 million RS images} &
  \multicolumn{1}{c}{0.3 to 30} &
  \multicolumn{1}{c}{\cellcolor[HTML]{EFEFEF}6 Continents} &
  \multicolumn{1}{c}{\begin{tabular}[c]{@{}c@{}}Image-level, Pixel-level, \\ Region-level, Spacial-temporal\end{tabular}} &
  \multicolumn{1}{c}{\cellcolor[HTML]{EFEFEF}MIM} &
  \multicolumn{1}{c|}{-} \\

% GeCO
\multicolumn{1}{|c}{\cellcolor[HTML]{FFF2CC}GeCO \cite{GeCo}} &
  \multicolumn{1}{c}{ResNet-50} &
  \multicolumn{1}{c}{\cellcolor[HTML]{EFEFEF}Levir-KR \cite{GeoKR}} &
  \multicolumn{1}{c}{0.8 to 16} &
  \multicolumn{1}{c}{\cellcolor[HTML]{EFEFEF}Global} &
  \multicolumn{1}{c}{\begin{tabular}[c]{@{}c@{}}Image-level, Pixel-level, \\ Region-level\end{tabular}} &
  \multicolumn{1}{c}{\cellcolor[HTML]{EFEFEF}SSL} &
  \multicolumn{1}{c|}{23.5M} \\

% RS-BYOL
\multicolumn{1}{|c}{\cellcolor[HTML]{FFF2CC}RS-BYOL \cite{RS-BYOL}} &
  \multicolumn{1}{c}{BYOL} &
  \multicolumn{1}{c}{\cellcolor[HTML]{EFEFEF}Sen12MS \cite{schmitt2019sen12mscurateddataset}} &
  \multicolumn{1}{c}{10 to 20} &
  \multicolumn{1}{c}{\cellcolor[HTML]{EFEFEF}Global} &
  \multicolumn{1}{c}{\begin{tabular}[c]{@{}c@{}}Image-level, \\ Pixel-level\end{tabular}} &
  \multicolumn{1}{c}{\cellcolor[HTML]{EFEFEF}SSL} &
  \multicolumn{1}{c|}{23.5M} \\

% CSPT
\multicolumn{1}{|c}{\cellcolor[HTML]{FFF2CC}CSPT \cite{CSPT}} &
  \multicolumn{1}{c}{ViT-B} &
  \multicolumn{1}{c}{\cellcolor[HTML]{EFEFEF}ImageNet-1K \cite{imageNet}} &
  \multicolumn{1}{c}{-} &
  \multicolumn{1}{c}{\cellcolor[HTML]{EFEFEF}Global} &
  \multicolumn{1}{c}{\begin{tabular}[c]{@{}c@{}}Image-level, \\ Region-level\end{tabular}} &
  \multicolumn{1}{c}{\cellcolor[HTML]{EFEFEF}SSL} &
  \multicolumn{1}{c|}{86M} \\

% RVSA
\multicolumn{1}{|c}{\cellcolor[HTML]{FFF2CC}RVSA \cite{RVSA}} &
  \multicolumn{1}{c}{ViT} &
  \multicolumn{1}{c}{\cellcolor[HTML]{EFEFEF}MillionAID \cite{Long2022ASP_millionaid, Long2021DiRS_millionaid}} &
  \multicolumn{1}{c}{0.5 to 153} &
  \multicolumn{1}{c}{\cellcolor[HTML]{EFEFEF}Global} &
  \multicolumn{1}{c}{\begin{tabular}[c]{@{}c@{}}Image-level, Pixel-level, \\ Region-level\end{tabular}} &
  \multicolumn{1}{c}{\cellcolor[HTML]{EFEFEF}MAE} &
  \multicolumn{1}{c|}{100M} \\

% SatMAE
\multicolumn{1}{|c}{\cellcolor[HTML]{FFF2CC}SatMAE \cite{SatMAE}} &
  \multicolumn{1}{c}{\begin{tabular}[c]{@{}c@{}}MAE-based \\ Framework\end{tabular}} &
  \multicolumn{1}{c}{\cellcolor[HTML]{EFEFEF}fMoW Sentinel-2 \cite{fmow2018}} &
  \multicolumn{1}{c}{10, 20, 60} &
  \multicolumn{1}{c}{\cellcolor[HTML]{EFEFEF}Global} &
  \multicolumn{1}{c}{\begin{tabular}[c]{@{}c@{}}Image-level, \\ Pixel-level\end{tabular}} &
  \multicolumn{1}{c}{\cellcolor[HTML]{EFEFEF}MAE} &
  \multicolumn{1}{c|}{307M} \\

% TOV
\multicolumn{1}{|c}{\cellcolor[HTML]{FFF2CC}TOV \cite{TOV}} &
  \multicolumn{1}{c}{TOV} &
  \multicolumn{1}{c}{\cellcolor[HTML]{EFEFEF}\begin{tabular}[c]{@{}c@{}}TOV-NI, \\ TOV-RS\end{tabular}} &
  \multicolumn{1}{c}{-} &
  \multicolumn{1}{c}{\cellcolor[HTML]{EFEFEF}Global} &
  \multicolumn{1}{c}{\begin{tabular}[c]{@{}c@{}}Image-level, Pixel-level, \\ Region-level\end{tabular}} &
  \multicolumn{1}{c}{\cellcolor[HTML]{EFEFEF}SSL} &
  \multicolumn{1}{c|}{-} \\

% CMID
\multicolumn{1}{|c}{\cellcolor[HTML]{FFF2CC}CMID \cite{CMID}} &
  \multicolumn{1}{c}{\begin{tabular}[c]{@{}c@{}}Teacher-student \\ Self-distillation\end{tabular}} &
  \multicolumn{1}{c}{\cellcolor[HTML]{EFEFEF}MillionAID \cite{Long2022ASP_millionaid, Long2021DiRS_millionaid}} &
  \multicolumn{1}{c}{Varied} &
  \multicolumn{1}{c}{\cellcolor[HTML]{EFEFEF}Global} &
  \multicolumn{1}{c}{\begin{tabular}[c]{@{}c@{}}Image-level, Pixel-level, \\ Region-level, Spacial-temporal\end{tabular}} &
  \multicolumn{1}{c}{\cellcolor[HTML]{EFEFEF}SSL} &
  \multicolumn{1}{c|}{25.6M/87.8M} \\

% CACo
\multicolumn{1}{|c}{\cellcolor[HTML]{FFF2CC}CACo \cite{CACo}} &
  \multicolumn{1}{c}{ResNet-18/50} &
  \multicolumn{1}{c}{\cellcolor[HTML]{EFEFEF}Sentinel-2 Imagery} &
  \multicolumn{1}{c}{10} &
  \multicolumn{1}{c}{\cellcolor[HTML]{EFEFEF}Global} &
  \multicolumn{1}{c}{\begin{tabular}[c]{@{}c@{}}Image-level, Pixel-level, \\ Spacial-temporal\end{tabular}} &
  \multicolumn{1}{c}{\cellcolor[HTML]{EFEFEF}SSL} &
  \multicolumn{1}{c|}{11.7M/23.5M} \\

% IaI-SimCLR
\multicolumn{1}{|c}{\cellcolor[HTML]{FFF2CC}IaI-SimCLR \cite{IaI-SimCLR}} &
  \multicolumn{1}{c}{ResNet-18} &
  \multicolumn{1}{c}{\cellcolor[HTML]{EFEFEF}SEN12MS} &
  \multicolumn{1}{c}{-} &
  \multicolumn{1}{c}{\cellcolor[HTML]{EFEFEF}Global} &
  \multicolumn{1}{c}{Image-level} &
  \multicolumn{1}{c}{\cellcolor[HTML]{EFEFEF}CL} &
  \multicolumn{1}{c|}{11.7M} \\

% SSL4EO-L
\multicolumn{1}{|c}{\cellcolor[HTML]{FFF2CC}SSL4EO-L \cite{SSL4EO-L}} &
  \multicolumn{1}{c}{ResNet/ViT} &
  \multicolumn{1}{c}{\cellcolor[HTML]{EFEFEF}\begin{tabular}[c]{@{}c@{}}ImageNet \cite{imageNet}, \\ MoCo \cite{MoCo}, SimCLR \cite{SimCLR}\end{tabular}} &
  \multicolumn{1}{c}{30} &
  \multicolumn{1}{c}{\cellcolor[HTML]{EFEFEF}Global} &
  \multicolumn{1}{c}{Pixel-level} &
  \multicolumn{1}{c}{\cellcolor[HTML]{EFEFEF}SSL} &
  \multicolumn{1}{c|}{11.7M/23.5M/86M} \\

% GFM
\multicolumn{1}{|c}{\cellcolor[HTML]{FFF2CC}GFM \cite{GFM}} &
  \multicolumn{1}{c}{Teacher-Student} &
  \multicolumn{1}{c}{\cellcolor[HTML]{EFEFEF}GeoPile \cite{GFM}} &
  \multicolumn{1}{c}{-} &
  \multicolumn{1}{c}{\cellcolor[HTML]{EFEFEF}Global} &
  \multicolumn{1}{c}{\begin{tabular}[c]{@{}c@{}}Image-level, \\ Pixel-level\end{tabular}} &
  \multicolumn{1}{c}{\cellcolor[HTML]{EFEFEF}Continual Pretraining} &
  \multicolumn{1}{c|}{-} \\

% SatLasPretrain
\multicolumn{1}{|c}{\cellcolor[HTML]{FFF2CC}SatlasPretrain \cite{SatlasPretrain}} &
  \multicolumn{1}{c}{SatlasNet} &
  \multicolumn{1}{c}{\cellcolor[HTML]{EFEFEF}GeoPile \cite{GFM}} &
  \multicolumn{1}{c}{1, 10} &
  \multicolumn{1}{c}{\cellcolor[HTML]{EFEFEF}Global} &
  \multicolumn{1}{c}{\begin{tabular}[c]{@{}c@{}}Image-level, \\ Pixel-level\end{tabular}} &
  \multicolumn{1}{c}{\cellcolor[HTML]{EFEFEF}Multi-task Learning} &
  \multicolumn{1}{c|}{88M} \\

% RingMo-Sense
\multicolumn{1}{|c}{\cellcolor[HTML]{FFF2CC}RingMo-Sense \cite{RingMo-Sense}} &
  \multicolumn{1}{c}{Multi-Branch} &
  \multicolumn{1}{c}{\cellcolor[HTML]{EFEFEF}\begin{tabular}[c]{@{}c@{}}RS Spatiotemporal \\ Dataset\end{tabular}} &
  \multicolumn{1}{c}{-} &
  \multicolumn{1}{c}{\cellcolor[HTML]{EFEFEF}Global} &
  \multicolumn{1}{c}{Pixel-level} &
  \multicolumn{1}{c}{\cellcolor[HTML]{EFEFEF}SSL} &
  \multicolumn{1}{c|}{-} \\ 

% Scale-MAE
\multicolumn{1}{|c}{\cellcolor[HTML]{FFF2CC}Scale-MAE \cite{Scale-MAE}} &
  \multicolumn{1}{c}{ViT-Large} &
  \multicolumn{1}{c}{\cellcolor[HTML]{EFEFEF}FMoW \cite{fmow2018}} &
  \multicolumn{1}{c}{-} &
  \multicolumn{1}{c}{\cellcolor[HTML]{EFEFEF}Global} &
  \multicolumn{1}{c}{\begin{tabular}[c]{@{}c@{}}Image-level, \\ Pixel-level\end{tabular}} &
  \multicolumn{1}{c}{\cellcolor[HTML]{EFEFEF}MAE} &
  \multicolumn{1}{c|}{322.9M} \\

% RingMo-lite
\multicolumn{1}{|c}{\cellcolor[HTML]{FFF2CC}RingMo-lite \cite{RingMo-lite}} &
  \multicolumn{1}{c}{CNN-Transformer} &
  \multicolumn{1}{c}{\cellcolor[HTML]{EFEFEF}AID \cite{Xia_2017}} &
  \multicolumn{1}{c}{0.3 to 30} &
  \multicolumn{1}{c}{\cellcolor[HTML]{EFEFEF}Global} &
  \multicolumn{1}{c}{\begin{tabular}[c]{@{}c@{}}Image-level, \\ Pixel-level, \\ Region-level, \\ Spacial-temporal\end{tabular}} &
  \multicolumn{1}{c}{\cellcolor[HTML]{EFEFEF}FD-MIM} &
  \multicolumn{1}{c|}{\begin{tabular}[c]{@{}c@{}}60\% \\ less than RingMo\end{tabular}} \\

% DeCUR
\multicolumn{1}{|c}{\cellcolor[HTML]{FFF2CC}DeCUR \cite{DeCUR}} &
  \multicolumn{1}{c}{Multimodel SSL} &
  \multicolumn{1}{c}{\cellcolor[HTML]{EFEFEF}\begin{tabular}[c]{@{}c@{}}SSL4EO-S12 \cite{wang2023ssl4eos12largescalemultimodalmultitemporal}, \\ RGB-DEM/depth\end{tabular}} &
  \multicolumn{1}{c}{Varied} &
  \multicolumn{1}{c}{\cellcolor[HTML]{EFEFEF}Global} &
  \multicolumn{1}{c}{\begin{tabular}[c]{@{}c@{}}Image-level, \\ Pixel-level\end{tabular}} &
  \multicolumn{1}{c}{\cellcolor[HTML]{EFEFEF}SSL} &
  \multicolumn{1}{c|}{23.5M} \\

% Feng et al.
\multicolumn{1}{|c}{\cellcolor[HTML]{FFF2CC}Feng et al. \cite{Feng2023}} &
  \multicolumn{1}{c}{MSFE+MMFH} &
  \multicolumn{1}{c}{\cellcolor[HTML]{EFEFEF}Multi-modal Dataset} &
  \multicolumn{1}{c}{Varied} &
  \multicolumn{1}{c}{\cellcolor[HTML]{EFEFEF}Global} &
  \multicolumn{1}{c}{\begin{tabular}[c]{@{}c@{}}Image-level, \\ Pixel-level, \\ Region-level, \\ Spacial-temporal\end{tabular}} &
  \multicolumn{1}{c}{\cellcolor[HTML]{EFEFEF}SSL} &
  \multicolumn{1}{c|}{-} \\ 

% FG-MAE
\multicolumn{1}{|c}{\cellcolor[HTML]{FFF2CC}FG-MAE \cite{FG-MAE}} &
  \multicolumn{1}{c}{ViT} &
  \multicolumn{1}{c}{\cellcolor[HTML]{EFEFEF}SSL4EO-S12 \cite{wang2023ssl4eos12largescalemultimodalmultitemporal}} &
  \multicolumn{1}{c}{10} &
  \multicolumn{1}{c}{\cellcolor[HTML]{EFEFEF}Global} &
  \multicolumn{1}{c}{\begin{tabular}[c]{@{}c@{}}Image-level, \\ Pixel-level\end{tabular}} &
  \multicolumn{1}{c}{\cellcolor[HTML]{EFEFEF}MAE} &
  \multicolumn{1}{c|}{-} \\

% Prithvi
\multicolumn{1}{|c}{\cellcolor[HTML]{FFF2CC}Prithvi \cite{Prithvi}} &
  \multicolumn{1}{c}{ViT} &
  \multicolumn{1}{c}{\cellcolor[HTML]{EFEFEF}Harmonized Landsat Sentinel 2} &
  \multicolumn{1}{c}{30} &
  \multicolumn{1}{c}{\cellcolor[HTML]{EFEFEF}Contiguous U.S.} &
  \multicolumn{1}{c}{Pixel-level} &
  \multicolumn{1}{c}{\cellcolor[HTML]{EFEFEF}MAE} &
  \multicolumn{1}{c|}{100M} \\

% CROMA
\multicolumn{1}{|c}{\cellcolor[HTML]{FFF2CC}CROMA \cite{CROMA}} &
  \multicolumn{1}{c}{Multimodal Encoder} &
  \multicolumn{1}{c}{\cellcolor[HTML]{EFEFEF}SSL4EO \cite{wang2023ssl4eos12largescalemultimodalmultitemporal}} &
  \multicolumn{1}{c}{10} &
  \multicolumn{1}{c}{\cellcolor[HTML]{EFEFEF}\begin{tabular}[c]{@{}c@{}}Areas Surrounding \\ Human Settlements\end{tabular}} &
  \multicolumn{1}{c}{\begin{tabular}[c]{@{}c@{}}Image-level, \\ Pixel-level\end{tabular}} &
  \multicolumn{1}{c}{\cellcolor[HTML]{EFEFEF}CL, MAE} &
  \multicolumn{1}{c|}{86M} \\

% USat
\multicolumn{1}{|c}{\cellcolor[HTML]{FFF2CC}USat \cite{USat}} &
  \multicolumn{1}{c}{ViT} &
  \multicolumn{1}{c}{\cellcolor[HTML]{EFEFEF}Satlas \cite{SatlasPretrain}} &
  \multicolumn{1}{c}{Varied} &
  \multicolumn{1}{c}{\cellcolor[HTML]{EFEFEF}Global} &
  \multicolumn{1}{c}{Pixel-level} &
  \multicolumn{1}{c}{\cellcolor[HTML]{EFEFEF}MAE} &
  \multicolumn{1}{c|}{-} \\

% Cross-Scale MAE
\multicolumn{1}{|c}{\cellcolor[HTML]{FFF2CC}Cross-Scale MAE \cite{Cross-Scale-MAE}} &
  \multicolumn{1}{c}{ViT-B} &
  \multicolumn{1}{c}{\cellcolor[HTML]{EFEFEF}fMoW \cite{fmow2018}} &
  \multicolumn{1}{c}{-} &
  \multicolumn{1}{c}{\cellcolor[HTML]{EFEFEF}Global} &
  \multicolumn{1}{c}{\begin{tabular}[c]{@{}c@{}}Image-level, \\ Pixel-level\end{tabular}} &
  \multicolumn{1}{c}{\cellcolor[HTML]{EFEFEF}MAE} &
  \multicolumn{1}{c|}{86M} \\

% U-BARN
\multicolumn{1}{|c}{\cellcolor[HTML]{FFF2CC}U-BARN \cite{U-BARN}} &
  \multicolumn{1}{c}{Unet+Transformer} &
  \multicolumn{1}{c}{\cellcolor[HTML]{EFEFEF}Sentinel-2 Imagery} &
  \multicolumn{1}{c}{Varied} &
  \multicolumn{1}{c}{\cellcolor[HTML]{EFEFEF}France} &
  \multicolumn{1}{c}{\begin{tabular}[c]{@{}c@{}}Image-level, \\ Pixel-level\end{tabular}} &
  \multicolumn{1}{c}{\cellcolor[HTML]{EFEFEF}SSL} &
  \multicolumn{1}{c|}{-} \\

% EarthPT
\multicolumn{1}{|c}{\cellcolor[HTML]{FFF2CC}EarthPT \cite{EarthPT}} &
  \multicolumn{1}{c}{Transformer} &
  \multicolumn{1}{c}{\cellcolor[HTML]{EFEFEF}Sentinel-2 Imagery} &
  \multicolumn{1}{c}{10} &
  \multicolumn{1}{c}{\cellcolor[HTML]{EFEFEF}UK} &
  \multicolumn{1}{c}{Image-level} &
  \multicolumn{1}{c}{\cellcolor[HTML]{EFEFEF}Autoregressive SSL} &
  \multicolumn{1}{c|}{700M} \\

% GeRSP
\multicolumn{1}{|c}{\cellcolor[HTML]{FFF2CC}GeRSP \cite{GeRSP}} &
  \multicolumn{1}{c}{\begin{tabular}[c]{@{}c@{}}Teacher-Student \\ Network\end{tabular}} &
  \multicolumn{1}{c}{\cellcolor[HTML]{EFEFEF}\begin{tabular}[c]{@{}c@{}}ImageNet \cite{imageNet}, \\ MillionAID \cite{Long2022ASP_millionaid, Long2021DiRS_millionaid}\end{tabular}} &
  \multicolumn{1}{c}{0.5 to 153} &
  \multicolumn{1}{c}{\cellcolor[HTML]{EFEFEF}Global} &
  \multicolumn{1}{c}{\begin{tabular}[c]{@{}c@{}}Image-level, \\ Pixel-level, \\ Region-level\end{tabular}} &
  \multicolumn{1}{c}{\cellcolor[HTML]{EFEFEF}SSL, SL} &
  \multicolumn{1}{c|}{-} \\

% SwiMDiff
\multicolumn{1}{|c}{\cellcolor[HTML]{FFF2CC}SwiMDiff \cite{SwiMDiff}} &
  \multicolumn{1}{c}{Dual-Branch} &
  \multicolumn{1}{c}{\cellcolor[HTML]{EFEFEF}Sen12MS \cite{schmitt2019sen12mscurateddataset}} &
  \multicolumn{1}{c}{Varied} &
  \multicolumn{1}{c}{\cellcolor[HTML]{EFEFEF}Global} &
  \multicolumn{1}{c}{\begin{tabular}[c]{@{}c@{}}Image-level, \\ Spacial-temporal\end{tabular}} &
  \multicolumn{1}{c}{\cellcolor[HTML]{EFEFEF}SSL} &
  \multicolumn{1}{c|}{11.7M} \\

% SMLFR
\multicolumn{1}{|c}{\cellcolor[HTML]{FFF2CC}SMLFR \cite{SMLFR}} &
  \multicolumn{1}{c}{Generative ConvNet} &
  \multicolumn{1}{c}{\cellcolor[HTML]{EFEFEF}GeoSense \cite{GeoSense}} &
  \multicolumn{1}{c}{0.05 to 150} &
  \multicolumn{1}{c}{\cellcolor[HTML]{EFEFEF}Multiple Continents} &
  \multicolumn{1}{c}{\begin{tabular}[c]{@{}c@{}}Pixel-level, \\ Region-level\end{tabular}} &
  \multicolumn{1}{c}{\cellcolor[HTML]{EFEFEF}SSL} &
  \multicolumn{1}{c|}{88M/197M} \\

% SpectralGPT
\multicolumn{1}{|c}{\cellcolor[HTML]{FFF2CC}SpectralGPT \cite{SpectralGPT}} &
  \multicolumn{1}{c}{3D GPT} &
  \multicolumn{1}{c}{\cellcolor[HTML]{EFEFEF}Sentinel-2 Imagery} &
  \multicolumn{1}{c}{Varied} &
  \multicolumn{1}{c}{\cellcolor[HTML]{EFEFEF}Global} &
  \multicolumn{1}{c}{\begin{tabular}[c]{@{}c@{}}Image-level, \\ Pixel-level, \\ Spacial-temporal\end{tabular}} &
  \multicolumn{1}{c}{\cellcolor[HTML]{EFEFEF}MAE} &
  \multicolumn{1}{c|}{100M/300M/600M} \\

% Presto
\multicolumn{1}{|c}{\cellcolor[HTML]{FFF2CC}Presto \cite{Presto}} &
  \multicolumn{1}{c}{\begin{tabular}[c]{@{}c@{}}MAE-based \\ Framework\end{tabular}} &
  \multicolumn{1}{c}{\cellcolor[HTML]{EFEFEF}Presto-21.5M \cite{Presto}} &
  \multicolumn{1}{c}{10} &
  \multicolumn{1}{c}{\cellcolor[HTML]{EFEFEF}Global} &
  \multicolumn{1}{c}{Crop Type Segmentation} &
  \multicolumn{1}{c}{\cellcolor[HTML]{EFEFEF}MAE} &
  \multicolumn{1}{c|}{402K} \\

% SatMAE++
\multicolumn{1}{|c}{\cellcolor[HTML]{FFF2CC}SatMAE++ \cite{SatMAE++}} &
  \multicolumn{1}{c}{SatMAE} &
  \multicolumn{1}{c}{\cellcolor[HTML]{EFEFEF}fMoW \cite{fmow2018}} &
  \multicolumn{1}{c}{Varied} &
  \multicolumn{1}{c}{\cellcolor[HTML]{EFEFEF}Global} &
  \multicolumn{1}{c}{Image-level} &
  \multicolumn{1}{c}{\cellcolor[HTML]{EFEFEF}\begin{tabular}[c]{@{}c@{}}Multi-Scale Pre-training\end{tabular}} &
  \multicolumn{1}{c|}{-} \\ 

% SAR-JEPA
\multicolumn{1}{|c}{\cellcolor[HTML]{FFF2CC}SAR-JEPA \cite{SAR-JEPA}} &
  \multicolumn{1}{c}{\begin{tabular}[c]{@{}c@{}}Joint-Embedding \\ Predictive Architecture\end{tabular}} &
  \multicolumn{1}{c}{\cellcolor[HTML]{EFEFEF}100K SAR Images} &
  \multicolumn{1}{c}{Varied} &
  \multicolumn{1}{c}{\cellcolor[HTML]{EFEFEF}Global} &
  \multicolumn{1}{c}{Image-level} &
  \multicolumn{1}{c}{\cellcolor[HTML]{EFEFEF}SSL} &
  \multicolumn{1}{c|}{-} \\

% FoMo-Bench
\multicolumn{1}{|c}{\cellcolor[HTML]{FFF2CC}FoMo-Bench \cite{FoMo-Bench}} &
  \multicolumn{1}{c}{ViT} &
  \multicolumn{1}{c}{\cellcolor[HTML]{EFEFEF}Multiple} &
  \multicolumn{1}{c}{Varied} &
  \multicolumn{1}{c}{\cellcolor[HTML]{EFEFEF}Global} &
  \multicolumn{1}{c}{\begin{tabular}[c]{@{}c@{}}Image-level, \\ Pixel-level, Region-level\end{tabular}} &
  \multicolumn{1}{c}{\cellcolor[HTML]{EFEFEF}MAE} &
  \multicolumn{1}{c|}{101M/110M} \\

% SkySense
\multicolumn{1}{|c}{\cellcolor[HTML]{FFF2CC}SkySense \cite{SkySense}} &
  \multicolumn{1}{c}{\begin{tabular}[c]{@{}c@{}}Factorized Multi-Modal \\ Spatiotemporal Encoder\end{tabular}} &
  \multicolumn{1}{c}{\cellcolor[HTML]{EFEFEF}Multiple} &
  \multicolumn{1}{c}{Varied} &
  \multicolumn{1}{c}{\cellcolor[HTML]{EFEFEF}Global} &
  \multicolumn{1}{c}{\begin{tabular}[c]{@{}c@{}}Image-level, \\ Pixel-level, \\ Region-level, \\ Spacial-temporal\end{tabular}} &
  \multicolumn{1}{c}{\cellcolor[HTML]{EFEFEF}CL} &
  \multicolumn{1}{c|}{2.06B} \\

% UPetu
\multicolumn{1}{|c}{\cellcolor[HTML]{FFF2CC}UPetu \cite{UPetu}} &
  \multicolumn{1}{c}{Multi-Modules} &
  \multicolumn{1}{c}{\cellcolor[HTML]{EFEFEF}GeoSense \cite{GeoSense}} &
  \multicolumn{1}{c}{-} &
  \multicolumn{1}{c}{\cellcolor[HTML]{EFEFEF}Global} &
  \multicolumn{1}{c}{\begin{tabular}[c]{@{}c@{}}Image-level, \\ Pixel-level, Spacial-temporal\end{tabular}} &
  \multicolumn{1}{c}{\cellcolor[HTML]{EFEFEF}SSL} &
  \multicolumn{1}{c|}{0.65M} \\

% msGFM
\multicolumn{1}{|c}{\cellcolor[HTML]{FFF2CC}msGFM \cite{msGFM}} &
  \multicolumn{1}{c}{Swin Transformer} &
  \multicolumn{1}{c}{\cellcolor[HTML]{EFEFEF}GeoPile-2 \cite{GFM}} &
  \multicolumn{1}{c}{0.1 to 153} &
  \multicolumn{1}{c}{\cellcolor[HTML]{EFEFEF}Global} &
  \multicolumn{1}{c}{\begin{tabular}[c]{@{}c@{}}Image-level, Pixel-level\end{tabular}} &
  \multicolumn{1}{c}{\cellcolor[HTML]{EFEFEF}MIM} &
  \multicolumn{1}{c|}{89M} \\

% % ORBIT
% \multicolumn{1}{|c}{\cellcolor[HTML]{FFF2CC}ORBIT \cite{ORBIT}} &
%   \multicolumn{1}{c}{ViT} &
%   \multicolumn{1}{c}{\cellcolor[HTML]{EFEFEF}CMIP6 \cite{CMIP6}} &
%   \multicolumn{1}{c}{1.40625 degrees*} &
%   \multicolumn{1}{c}{\cellcolor[HTML]{EFEFEF}Global} &
%   \multicolumn{1}{c}{\begin{tabular}[c]{@{}c@{}}Climate Prediction\end{tabular}} &
%   \multicolumn{1}{c}{-} &
%   \multicolumn{1}{c|}{113B} \\ 

% DINO-MC
\multicolumn{1}{|c}{\cellcolor[HTML]{FFF2CC}DINO-MC \cite{DINO-MC}} &
  \multicolumn{1}{c}{DINO} &
  \multicolumn{1}{c}{\cellcolor[HTML]{EFEFEF}SeCo-100K \cite{SeCo}} &
  \multicolumn{1}{c}{10 to 60} &
  \multicolumn{1}{c}{\cellcolor[HTML]{EFEFEF}Global} &
  \multicolumn{1}{c}{\begin{tabular}[c]{@{}c@{}}Image-level, Spacial-temporal\end{tabular}} &
  \multicolumn{1}{c}{\cellcolor[HTML]{EFEFEF}SSL} &
  \multicolumn{1}{c|}{-} \\

% OFA-Net
\multicolumn{1}{|c}{\cellcolor[HTML]{FFF2CC}OFA-Net \cite{OFA-Net}} &
  \multicolumn{1}{c}{OFA-Net} &
  \multicolumn{1}{c}{\cellcolor[HTML]{EFEFEF}Multi-modal Dataset} &
  \multicolumn{1}{c}{Varied} &
  \multicolumn{1}{c}{\cellcolor[HTML]{EFEFEF}Global} &
  \multicolumn{1}{c}{\begin{tabular}[c]{@{}c@{}}Image-level, Pixel-level\end{tabular}} &
  \multicolumn{1}{c}{\cellcolor[HTML]{EFEFEF}MIM} &
  \multicolumn{1}{c|}{-} \\

% MTP
\multicolumn{1}{|c}{\cellcolor[HTML]{FFF2CC}MTP \cite{MTP}} &
  \multicolumn{1}{c}{\begin{tabular}[c]{@{}c@{}}Shared Encoder \\ Task-Specific Decoders\end{tabular}} &
  \multicolumn{1}{c}{\cellcolor[HTML]{EFEFEF}SAMRS \cite{SAMRS}} &
  \multicolumn{1}{c}{Varied} &
  \multicolumn{1}{c}{\cellcolor[HTML]{EFEFEF}Global} &
  \multicolumn{1}{c}{\begin{tabular}[c]{@{}c@{}}Image-level, Pixel-level, \\ Region-level, Spacial-temporal\end{tabular}} &
  \multicolumn{1}{c}{\cellcolor[HTML]{EFEFEF}\begin{tabular}[c]{@{}c@{}}Multi-Task \\ Pretraining\end{tabular}} &
  \multicolumn{1}{c|}{over 300M} \\ 

% BFM
\multicolumn{1}{|c}{\cellcolor[HTML]{FFF2CC}BFM \cite{BFM}} &
  \multicolumn{1}{c}{ViT} &
  \multicolumn{1}{c}{\cellcolor[HTML]{EFEFEF}MillionAID \cite{Long2022ASP_millionaid, Long2021DiRS_millionaid}} &
  \multicolumn{1}{c}{0.5 to 153} &
  \multicolumn{1}{c}{\cellcolor[HTML]{EFEFEF}Global} &
  \multicolumn{1}{c}{\begin{tabular}[c]{@{}c@{}}Pixel-level,\\ Region-level\end{tabular}} &
  \multicolumn{1}{c}{\cellcolor[HTML]{EFEFEF}MAE} &
  \multicolumn{1}{c|}{\begin{tabular}[c]{@{}c@{}}86M/605.26M/ \\1.36B/2.42B\end{tabular}} \\

% MMEarth
\multicolumn{1}{|c}{\cellcolor[HTML]{FFF2CC}MMEarth \cite{MMEarth}} &
  \multicolumn{1}{c}{MP-MAE} &
  \multicolumn{1}{c}{\cellcolor[HTML]{EFEFEF}\begin{tabular}[c]{@{}c@{}}Multi-modal, \\ Geospatial Data\end{tabular}} &
  \multicolumn{1}{c}{-} &
  \multicolumn{1}{c}{\cellcolor[HTML]{EFEFEF}Global} &
  \multicolumn{1}{c}{\begin{tabular}[c]{@{}c@{}}Image-level, Pixel-level\end{tabular}} &
  \multicolumn{1}{c}{\cellcolor[HTML]{EFEFEF}MP-MAE} &
  \multicolumn{1}{c|}{3.7M to 650M} \\
  
% CtxMIM
\multicolumn{1}{|c}{\cellcolor[HTML]{FFF2CC}CtxMIM \cite{CtxMIM}} &
  \multicolumn{1}{c}{ViT} &
  \multicolumn{1}{c}{\cellcolor[HTML]{EFEFEF}WorldView-3 Imagery} &
  \multicolumn{1}{c}{Varied} &
  \multicolumn{1}{c}{\cellcolor[HTML]{EFEFEF}Asia} &
  \multicolumn{1}{c}{\begin{tabular}[c]{@{}c@{}}Image-level, \\ Pixel-level, \\ Region-level\end{tabular}} &
  \multicolumn{1}{c}{\cellcolor[HTML]{EFEFEF}MIM} &
  \multicolumn{1}{c|}{88M} \\ 

% SARATR-X
\multicolumn{1}{|c}{\cellcolor[HTML]{FFF2CC}SARATR-X \cite{SARATR-X}} &
  \multicolumn{1}{c}{HiViT} &
  \multicolumn{1}{c}{\cellcolor[HTML]{EFEFEF}SAR Datasets} &
  \multicolumn{1}{c}{0.1 to 3} &
  \multicolumn{1}{c}{\cellcolor[HTML]{EFEFEF}Global} &
  \multicolumn{1}{c}{\begin{tabular}[c]{@{}c@{}}Image-level, Region-level\end{tabular}} &
  \multicolumn{1}{c}{\cellcolor[HTML]{EFEFEF}MIM} &
  \multicolumn{1}{c|}{66M} \\

% SoftCon
\multicolumn{1}{|c}{\cellcolor[HTML]{FFF2CC}SoftCon \cite{SoftCon}} &
  \multicolumn{1}{c}{\begin{tabular}[c]{@{}c@{}}Siamese Network \\ with ResNet \\ and ViT Backbones\end{tabular}} &
  \multicolumn{1}{c}{\cellcolor[HTML]{EFEFEF}SSL4EO-S12-ML \cite{wang2023ssl4eos12largescalemultimodalmultitemporal}} &
  \multicolumn{1}{c}{-} &
  \multicolumn{1}{c}{\cellcolor[HTML]{EFEFEF}Global} &
  \multicolumn{1}{c}{\begin{tabular}[c]{@{}c@{}}Image-level, \\ Pixel-level, \\ Spacial-temporal\end{tabular}} &
  \multicolumn{1}{c}{\cellcolor[HTML]{EFEFEF}\begin{tabular}[c]{@{}c@{}}Multi-label Soft \\ Contrastive Learning\end{tabular}} &
  \multicolumn{1}{c|}{23M, 23M, 86M} \\ 

% LeMeViT
\multicolumn{1}{|c}{\cellcolor[HTML]{FFF2CC}LeMeViT \cite{LeMeViT}} &
  \multicolumn{1}{c}{Hierarchical ViT} &
  \multicolumn{1}{c}{\cellcolor[HTML]{EFEFEF}MillionAID \cite{Long2022ASP_millionaid, Long2021DiRS_millionaid}} &
  \multicolumn{1}{c}{-} &
  \multicolumn{1}{c}{\cellcolor[HTML]{EFEFEF}-} &
  \multicolumn{1}{c}{\begin{tabular}[c]{@{}c@{}}Image-level, Pixel-level, \\Region-level , Spacial-temporal\end{tabular}} &
  \multicolumn{1}{c}{\cellcolor[HTML]{EFEFEF}\begin{tabular}[c]{@{}c@{}}Dual Cross-Attention \\ with Learnable Meta Token \\ Adaptation\end{tabular}} &
  \multicolumn{1}{c|}{8.33M to 52.61M} \\

% S2MAE
\multicolumn{1}{|c}{\cellcolor[HTML]{FFF2CC}S2MAE \cite{S2MAE}} &
  \multicolumn{1}{c}{\begin{tabular}[c]{@{}c@{}}3D Transformer-based\\ MAE\end{tabular}} &
  \multicolumn{1}{c}{\cellcolor[HTML]{EFEFEF}\begin{tabular}[c]{@{}c@{}}fMoW-Sentinel \cite{fmow2018}, \\ BigEarthNet \cite{Sumbul_2019_Bigearthnet}\end{tabular}} &
  \multicolumn{1}{c}{-} &
  \multicolumn{1}{c}{\cellcolor[HTML]{EFEFEF}Global} &
  \multicolumn{1}{c}{\begin{tabular}[c]{@{}c@{}}Image-level, Spacial-temporal\end{tabular}} &
  \multicolumn{1}{c}{\cellcolor[HTML]{EFEFEF}3D MAE} &
  \multicolumn{1}{c|}{-} \\ 
  
% RS-DFM
\multicolumn{1}{|c}{\cellcolor[HTML]{FFF2CC}RS-DFM \cite{RS-DFM}} &
  \multicolumn{1}{c}{\begin{tabular}[c]{@{}c@{}}Multi-platform \\ Inference Framework\end{tabular}} &
  \multicolumn{1}{c}{\cellcolor[HTML]{EFEFEF}AirCo-MultiTasks \cite{RS-DFM}} &
  \multicolumn{1}{c}{-} &
  \multicolumn{1}{c}{\cellcolor[HTML]{EFEFEF}-} &
  \multicolumn{1}{c}{\begin{tabular}[c]{@{}c@{}}3D Region-level, \\ Pixel-level\end{tabular}} &
  \multicolumn{1}{c}{\cellcolor[HTML]{EFEFEF}\begin{tabular}[c]{@{}c@{}}Generalized Feature \\ Mapping with Relative \\Depth Estimation\end{tabular}} &
  \multicolumn{1}{c|}{-} \\ 

% A2-MAE
\multicolumn{1}{|c}{\cellcolor[HTML]{FFF2CC}A2-MAE \cite{A2-MAE}} &
  \multicolumn{1}{c}{ViT-Large} &
  \multicolumn{1}{c}{\cellcolor[HTML]{EFEFEF}\begin{tabular}[c]{@{}c@{}}STSSD \\ (Spatial-Temporal-Spectral\\Structured Dataset) \end{tabular}} &
  \multicolumn{1}{c}{0.8 - 30m} &
  \multicolumn{1}{c}{\cellcolor[HTML]{EFEFEF}Global} &
  \multicolumn{1}{c}{\begin{tabular}[c]{@{}c@{}}Image-level, \\ Pixel-level, \\ Spacial-temporal\end{tabular}} &
  \multicolumn{1}{c}{\cellcolor[HTML]{EFEFEF}\begin{tabular}[c]{@{}c@{}}Anchor-aware\\Masking Strategy\\and Geographic \\Encoding Module\end{tabular}} &
  \multicolumn{1}{c|}{304M} \\

% HyperSIGMA
\multicolumn{1}{|c}{\cellcolor[HTML]{FFF2CC}HyperSIGMA \cite{HyperSIGMA}} &
  \multicolumn{1}{c}{ViT-based} &
  \multicolumn{1}{c}{\cellcolor[HTML]{EFEFEF}HyperGlobal-450K \cite{HyperSIGMA}} &
  \multicolumn{1}{c}{30m} &
  \multicolumn{1}{c}{\cellcolor[HTML]{EFEFEF}Global} &
  \multicolumn{1}{c}{\begin{tabular}[c]{@{}c@{}}Image-level, Region-level, \\ Anomaly Detection, \\ Spacial-temporal\end{tabular}} &
  \multicolumn{1}{c}{\cellcolor[HTML]{EFEFEF}MAE} &
  \multicolumn{1}{c|}{over 1B} \\ 

% DOFA
\multicolumn{1}{|c}{\cellcolor[HTML]{FFF2CC}DOFA \cite{DOFA}} &
  \multicolumn{1}{c}{Dynamic OFA} &
  \multicolumn{1}{c}{\cellcolor[HTML]{EFEFEF}Multiple} &
  \multicolumn{1}{c}{1 to 30} &
  \multicolumn{1}{c}{\cellcolor[HTML]{EFEFEF}Global} &
  \multicolumn{1}{c}{\begin{tabular}[c]{@{}c@{}}Image-level, \\ Pixel-level\end{tabular}} &
  \multicolumn{1}{c}{\cellcolor[HTML]{EFEFEF}MIM} &
  \multicolumn{1}{c|}{111M/337M} \\

\hline

\end{longtable}
\end{landscape}
\twocolumn

    \subsubsection{Pixel-Level}
Pixel-level analysis offers the most granular form of image perception, assigning a label to every pixel within an image. This includes tasks such as semantic segmentation, where each pixel is classified into categories like vegetation, water, or buildings; it also include change detection, which identifies temporal differences between images captured at different times. Pixel-level analysis is indispensable for creating highly detailed maps used in applications like precision agriculture, deforestation tracking, and disaster management. The ability to analyze fine-grained details enables more accurate assessments and actionable insights for these critical areas.

    % \input{tables/cv_model1}
    % \input{tables/cv_model2}

% (figure for ViT/transformer)
\begin{figure*}[ht]
  \centering
  \includegraphics[scale=0.6]{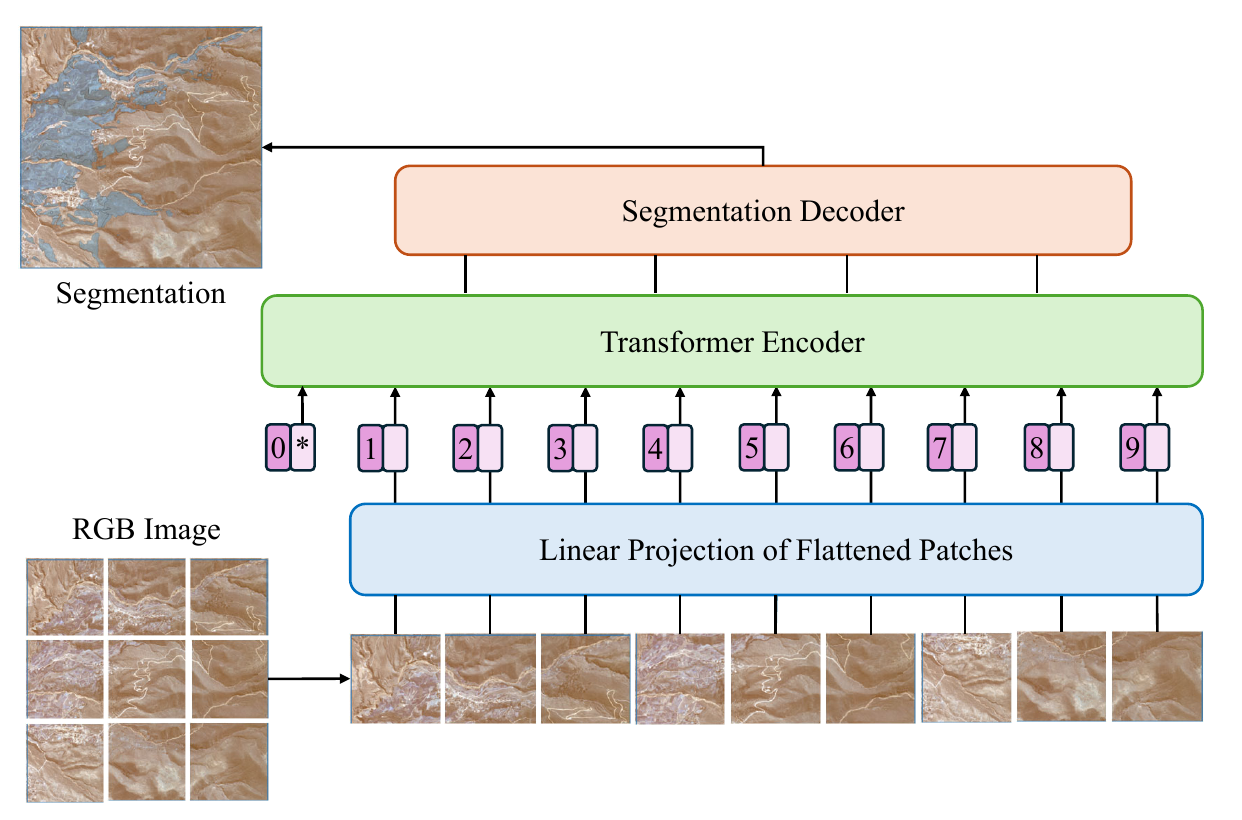}
  \caption{The Vision transformer architecture.\protect\footnotemark}
  \label{fig:vit}
\end{figure*}

\footnotetext{RGB Image and Segmentation © MAXAR 2024, provided through the NextView License Agreement.}

\subsection{Backbone}
    \subsubsection{Convolutional Neural Networks (CNNs)}
Convolutional Neural Networks \cite{CNNs} are a fundamental architecture in deep learning, designed to extract hierarchical spatial features from images through the use of convolutional layers. Each convolutional layer applies filters to the input data, detecting patterns like edges, textures, and shapes at different levels of abstraction. This makes CNNs well-suited for handling complex visual tasks in remote sensing, such as image classification, segmentation, and object detection.

Residual Neural Networks (ResNet) \cite{resnet}, {a type of Convolutional Neural Network (CNN) , address the degradation problem in deep neural networks by introducing residual connections, which allow gradients to bypass certain layers, facilitating the training of very deep networks.
 This capability is particularly beneficial in remote sensing, where deep models are often required to capture the intricate details and variations in satellite images.
ResNet, as an example, is characterized by their residual blocks, which include shortcut connections that bypass one or more layers. The residual block can be described by the following equation:
\[
\mathbf{y} = \mathcal{F}(\mathbf{x}, \{W_i\}) + \mathbf{x}
\]
where \(\mathbf{y}\) is the output, \(\mathcal{F}\) represents the residual mapping to be learned, \(\mathbf{x}\) is the input, and \(\{W_i\}\) are the layer weights \cite{he2015deep}. \\

ResNet has various architectures like ResNet-50, ResNet-101, and ResNet-152, with the number indicating the total layers. These networks have shown remarkable performance in various vision tasks due to their ability to train deeper networks without degradation. In remote sensing, ResNets are widely used for image classification, object detection and change detection tasks \cite{rs16020327} . For example, ResNet-based models can classify different land cover types \cite{ZHU202172, resnet_classification}, detect objects like buildings and vehicles \cite{rs16020327}, and monitor changes \cite{ZHU202172, rs15082092} in the landscape over time by comparing temporal sequences of satellite images.

    \subsubsection{Transformers and Vision Transformers (ViTs)}
Transformers, adapted for computer vision (CV) as Vision Transformers (ViT), model long-range dependencies through self-attention, making them effective for complex geospatial data. Figure \ref{fig:vit} illustrates the architecture of ViT. ViTs treat images as sequences of patches, capturing global and local patterns, which is useful for segmentation and change detection. The self-attention mechanism computes:
\[
\text{Attention}(Q, K, V) = \text{softmax}\left( \frac{QK^T}{\sqrt{d_k}} \right) V
\]
where \(Q\) (query), \(K\) (key), and \(V\) (value) are the input matrices, and \(d_k\) is the dimension of the key vectors \cite{vaswani2023attention}.

%unchanged
By incorporating these methodologies, foundation models for remote sensing can leverage vast amounts of data, handle complex structures, and achieve state-of-the-art performance across various applications. These methodologies enable models to effectively address the unique challenges of remote sensing, such as large image sizes, diverse data sources, and the need for high accuracy in environmental monitoring and analysis.

In the following sections, we will explore specific applications of these methodologies in different remote sensing tasks, analyze their performance, and discuss the datasets used to train and evaluate these models.

\section{Data and Tasks}

\subsection{Data}
Datasets play a crucial role in remote sensing, providing the foundation for training and evaluating models. High-quality datasets enable models to learn accurate representations of the Earth's surface, improving their performance on various remote sensing tasks. In figure \ref{fig:data-task-example}, we showcase some examples of the data used for training foundation models and their downstream tasks. In this section, we provide an overview of commonly used datasets in table \ref{tab:pre-train_datasets} for remote sensing, discussing their characteristics, applications, and relevance to foundation models. These datasets, with their varying resolutions, categories, and geographic coverage, provide a rich resource for advancing remote sensing research and applications. They facilitate the development of robust models capable of addressing diverse challenges in understanding and interpreting Earth's surface through remote sensing technologies.

Datasets used in remote sensing vary significantly in size, from hundreds of thousands of samples, as in RSD46-WHU \cite{Long2017, Xiao2017}, to over a million, as seen in MillionAID \cite{Long2022ASP_millionaid, Long2021DiRS_millionaid}Generally, larger datasets contribute to model generalization by encompassing diverse geographic areas, seasonal variations, and environmental conditions. Dataset resolutions also range from high (sub-meter), suitable for tasks requiring detailed spatial analysis, to moderate (10-60 meters), as with SEN12MS \cite{schmitt2019sen12mscurateddataset} and SSL4EO-S12 \cite{wang2023ssl4eos12largescalemultimodalmultitemporal}, which support broader pattern recognition applications.

These datasets leverage various sensor types, including RGB, multispectral, hyperspectral, and synthetic aperture radar (SAR). For instance, SEN12MS \cite{schmitt2019sen12mscurateddataset} integrates both SAR and multispectral imagery, enabling models to learn from distinct data modalities. This diversity in sensor types is critical for robust model development, as each sensor type captures unique surface characteristics, supporting tasks that benefit from cross-modal information.

Foundation models, in particular, benefit from such large-scale, multimodal datasets, which support self-supervised and supervised training approaches across tasks such as scene classification, segmentation, and object detection. For further insight, the appendix includes detailed descriptions of each dataset’s structure, unique characteristics, and application roles, enhancing the understanding of their impact on remote sensing advancements.

\subsection{Tasks}
Different applications in remote sensing address particular real-world challenges by leveraging the capabilities of foundation models. These tasks include environmental monitoring, archaeology, agriculture, urban planning and development, and disaster management. To highlight the versatility of foundation models in remote sensing, we present table \ref{tab:cv-domain} that categorizes models based on their applicability to various applications, as well as the different image analysis methods used. This figure serves as a quick reference for researchers to identify suitable models for their specific needs.

% \begin{figure*}
%     \centering
%     \includegraphics[width=\textwidth]{figures/fig6_domain-CV.pdf}
%     \caption{This diagram illustrates various domain-specific tasks in computer vision and their connections. Key areas include Environmental Monitoring, Agriculture and Forestry, Urban Planning and Development, Disaster Management, and Archaeology. Each domain comprises specific tasks such as change detection and object Detection. The relationships between these tasks and their application domains are depicted through connecting arrows, emphasizing the interconnected nature of computer vision applications across different fields.}
%     \label{fig:domain-cv-tasks}
% \end{figure*}

% \usepackage{multirow}
\begin{table*}[]
\centering
\resizebox{\textwidth}{!}{%
\begin{tabular}{|cc|
>{\columncolor[HTML]{F3F3F3}}c c
>{\columncolor[HTML]{F3F3F3}}c c|
>{\columncolor[HTML]{F3F3F3}}l |}
\hline
\multicolumn{2}{|c|}{\cellcolor[HTML]{C9DAF8}} &
  \multicolumn{4}{c|}{\cellcolor[HTML]{C9DAF8}
    \parbox[c][1.15cm]{\linewidth}{\centering\textbf{\Large Image Analysis by Levels}}} &
  \multicolumn{1}{c|}{\cellcolor[HTML]{C9DAF8}} \\ \cline{3-6}
  
\multicolumn{2}{|c|}{\multirow{-2}{*}{\cellcolor[HTML]{C9DAF8}\centering\textbf{\Large Tasks}}} &
  \multicolumn{1}{c|}{\cellcolor[HTML]{CFE2F3}
    \parbox[c][1cm]{0.2\linewidth}{\centering\textbf{\large Image-Level}}} &
  \multicolumn{1}{c|}{\cellcolor[HTML]{CFE2F3}
    \parbox[c][1cm]{0.2\linewidth}{\centering\textbf{\large Pixel-Level}}} &
  \multicolumn{1}{c|}{\cellcolor[HTML]{CFE2F3}
    \parbox[c][1cm]{0.2\linewidth}{\centering\textbf{\large Region-Level}}} &
  \multicolumn{1}{c|}{\cellcolor[HTML]{CFE2F3}
    \parbox[c][1cm]{0.2\linewidth}{\centering\textbf{\large Spatial-Temporal}}} &
    
  \multicolumn{1}{c|}{\multirow{-2}{*}{\cellcolor[HTML]{C9DAF8}\centering\textbf{\Large Related Work}}} \\ \hline

% Environmental Monitoring \cellcolor[HTML]{F4CCCC}
\multicolumn{1}{|c|}{\multirow{8}{*}{\centering \textbf{\large Environmental Monitoring}}} &

\cellcolor[HTML]{F4CCCC} \begin{tabular}[c]{@{}c@{}}Land Cover \\ Change Detection\end{tabular} &
  \multicolumn{1}{c|}{\cellcolor[HTML]{F3F3F3}} &
  \multicolumn{1}{c|}{} &
  \multicolumn{1}{c|}{\parbox[c][1.1cm]{0.25\linewidth}{\centering \cellcolor[HTML]{F3F3F3}}} &
  \multicolumn{1}{c|}{\scalebox{2.5}{\centering$\checkmark$}} &
  \begin{tabular}[c]{@{}l@{}}\cite{SeCo, MATTER, RSP, RingMo, CMID, CACo, GFM, RingMo-lite, SwiMDiff, SpectralGPT} \\ \cite{SkySense, UPetu, DINO-MC, MTP, SoftCon, LeMeViT, S2MAE, A2-MAE, HyperSIGMA} \end{tabular} \\ \cline{2-7}
\multicolumn{1}{|c|}{} &
\cellcolor[HTML]{F4CCCC} \begin{tabular}[c]{@{}c@{}}Deforestation Monitoring \end{tabular} &
  \multicolumn{1}{c|}{\cellcolor[HTML]{F3F3F3}} &
  \multicolumn{1}{c|}{\scalebox{2.5}{$\checkmark$}} &
  \multicolumn{1}{c|}{\parbox[c][1.2cm]{0.25\linewidth}{\centering \cellcolor[HTML]{F3F3F3}}} &
  \multicolumn{1}{c|}{} &
  \begin{tabular}[c]{@{}l@{}}
  \cite{GeoKR, MATTER, GASSL, RSP, Scheibenreif_2022_CVPR, RingMo, GeCo, RS-BYOL, RVSA, SatMAE,TOV} \\ 
  \cite{CMID, CACo, GFM, SatlasPretrain, RingMo-Sense, Scale-MAE, RingMo-lite, DeCUR, Feng2023} \\ 
  \cite{FG-MAE, CROMA, Cross-Scale-MAE, U-BARN, GeRSP, SMLFR, SpectralGPT, Presto, SkySense, UPetu} \\ 
  \cite{msGFM, OFA-Net, MTP, CtxMIM, SoftCon, LeMeViT, RS-DFM, A2-MAE, HyperSIGMA, DOFA} \end{tabular} \\ \cline{2-7}
\multicolumn{1}{|c|}{} &
\cellcolor[HTML]{F4CCCC}  \begin{tabular}[c]{@{}c@{}}Water Body Analysis \end{tabular} &
  \multicolumn{1}{c|}{\cellcolor[HTML]{F3F3F3}} &
  \multicolumn{1}{c|}{\scalebox{2.5}{$\checkmark$}} &
  \multicolumn{1}{c|}{\parbox[c][1.2cm]{0.25\linewidth}{\centering \cellcolor[HTML]{F3F3F3}}\scalebox{2.5}{$\checkmark$}} & 
   \scalebox{2.5}{$\checkmark$}&
  \begin{tabular}[c|]{@{}l@{}} \cite{RSP, RingMo, CMID, Feng2023, SkySense, MTP, LeMeViT, HyperSIGMA}\end{tabular} \\ \cline{2-7} 
\multicolumn{1}{|c|}{} &
\cellcolor[HTML]{F4CCCC}  \begin{tabular}[c]{@{}c@{}}Forest \\ Cover mapping\end{tabular} &
  \multicolumn{1}{c|}{\cellcolor[HTML]{F3F3F3}} &
  \multicolumn{1}{c|}{\scalebox{2.5}{$\checkmark$}} &
  \multicolumn{1}{c|}{\cellcolor[HTML]{F3F3F3}} &
   \scalebox{2.5}{$\checkmark$}&
  \begin{tabular}[c|]{@{}l@{}}\cite{RSP, RingMo, CACo, GFM, RingMo-lite, Feng2023, SpectralGPT, SoftCon, LeMeViT}\\ \cite{S2MAE, A2-MAE, HyperSIGMA} \end{tabular}\\ \cline{2-7}
\multicolumn{1}{|c|}{} &
\cellcolor[HTML]{F4CCCC}  \begin{tabular}[c]{@{}c@{}}Biomass Estimation\end{tabular} &
  \multicolumn{1}{c|}{\cellcolor[HTML]{F3F3F3}} &
  \multicolumn{1}{c|}{} &
  \multicolumn{1}{c|}{\parbox[c][0.8cm]{0.25\linewidth}{\centering \cellcolor[HTML]{F3F3F3}}} &
  \scalebox{2.5}{} &
  \cite{IaI-SimCLR} \\ \cline{2-7}
\multicolumn{1}{|c|}{} &
\cellcolor[HTML]{F4CCCC}  \begin{tabular}[c]{@{}c@{}}Weather/Climate \\ Prediction \end{tabular} &
  \multicolumn{1}{c|}{\cellcolor[HTML]{F3F3F3}\scalebox{2.5}{$\checkmark$}} &
  \multicolumn{1}{c|}{} &
  \multicolumn{1}{c|}{\parbox[c][1cm]{0.25\linewidth}{\centering \cellcolor[HTML]{F3F3F3}}} &
   &
  \cite{EarthPT, RingMo-Sense, ORBIT} \\ \cline{2-7}
\multicolumn{1}{|c|}{} &
\cellcolor[HTML]{F4CCCC}  \begin{tabular}[c]{@{}c@{}}Cloud Removal\end{tabular} &
  \multicolumn{1}{c|}{\cellcolor[HTML]{F3F3F3}} &
  \multicolumn{1}{c|}{} &
  \multicolumn{1}{c|}{\parbox[c][0.9cm]{0.25\linewidth}{\centering \cellcolor[HTML]{F3F3F3}}} &
   &
  \cite{msGFM} \\ \cline{2-7}
\multicolumn{1}{|c|}{} &
\cellcolor[HTML]{F4CCCC}  \begin{tabular}[c]{@{}c@{}}Moisture Content \\ Measurement\end{tabular} &
  \multicolumn{1}{c|}{\cellcolor[HTML]{F3F3F3}} &
  \multicolumn{1}{c|}{} &
  \multicolumn{1}{c|}{\parbox[c][0.9cm]{0.25\linewidth}{\centering \cellcolor[HTML]{F3F3F3}}} &
   &
  \cite{Presto} \\ \hline

% Agriculture \cellcolor[HTML]{FCE5CD}
\multicolumn{1}{|c|}{\multirow{7}{*}{\centering\textbf{\large Agriculture}}} &
    
\cellcolor[HTML]{FCE5CD}  \begin{tabular}[c]{@{}c@{}}Crop Type Mapping\end{tabular} &
  \multicolumn{1}{c|}{\cellcolor[HTML]{F3F3F3}\scalebox{2.5}{$\checkmark$}} &
  \multicolumn{1}{c|}{\scalebox{2.5}{$\checkmark$}} &
  \multicolumn{1}{c|}{\parbox[c][1.2cm]{0.25\linewidth}{\centering \cellcolor[HTML]{F3F3F3}\scalebox{2.5}{$\checkmark$}}} &
  \multicolumn{1}{c|}{\scalebox{2.5}{$\checkmark$}}&
  \cite{RSP, RingMo, CMID, Feng2023, SkySense, MTP, LeMeViT, HyperSIGMA, SSL4EO-L} \\ \cline{2-7}
\multicolumn{1}{|c|}{} &
\cellcolor[HTML]{FCE5CD}  \begin{tabular}[c]{@{}c@{}}Weed Detection\end{tabular} &
   \multicolumn{1}{c|}{\cellcolor[HTML]{F3F3F3}} &
   \multicolumn{1}{c|}{} &
   \multicolumn{1}{c|}{\parbox[c][1.2cm]{0.25\linewidth}{\centering \cellcolor[HTML]{F3F3F3}\scalebox{2.5}{$\checkmark$}}} & 
   &
  \begin{tabular}[c]{@{}l@{}}\cite{GeoKR, GASSL, RSP, RingMo, GeCo, CSPT, RVSA, TOV, CMID, Feng2023} \\ 
  \cite{FoMo-Bench, SkySense, MTP, CtxMIM, LeMeViT, RS-DFM, HyperSIGMA, SMLFR}\end{tabular} \\ \cline{2-7}
  
\multicolumn{1}{|c|}{} &
\cellcolor[HTML]{FCE5CD}  \begin{tabular}[c]{@{}c@{}}Disease  Monitoring\end{tabular} &
  \multicolumn{1}{c|}{\cellcolor[HTML]{F3F3F3}\scalebox{2.5}{$\checkmark$}} &
  \multicolumn{1}{c|}{\parbox[c][1.2cm]{0.25\linewidth}{\centering \scalebox{2.5}{$\checkmark$}}} & 
  \multicolumn{1}{c|}{\cellcolor[HTML]{F3F3F3}} &
  \scalebox{2.5}{$\checkmark$} &
  \begin{tabular}[c]{@{}l@{}}
  \cite{GeoKR, GASSL, RSP, RingMo, GeCo, RVSA, TOV, CMID, Feng2023} \\ 
  \cite{SkySense, MTP, CtxMIM, LeMeViT, RS-DFM, HyperSIGMA, SMLFR} \end{tabular} \\ \cline{2-7}
\multicolumn{1}{|c|}{} &
\cellcolor[HTML]{FCE5CD} Forecasting &
  \multicolumn{1}{c|}{\parbox[c][1.2cm]{0.25\linewidth}{\centering \cellcolor[HTML]{F3F3F3}}} &
  \multicolumn{1}{c|}{} &
  \multicolumn{1}{c|}{\cellcolor[HTML]{F3F3F3}} &
   &
  \cite{EarthPT, RingMo-Sense} \\ \cline{2-7}
\multicolumn{1}{|c|}{} &
\cellcolor[HTML]{FCE5CD}  \begin{tabular}[c]{@{}c@{}}Soil Parameter Estimation\end{tabular} &
  \multicolumn{1}{c|}{\cellcolor[HTML]{F3F3F3}} &
  \multicolumn{1}{c|}{} &
  \multicolumn{1}{c|}{\parbox[c][1.2cm]{0.25\linewidth}{\centering \cellcolor[HTML]{F3F3F3}}} &
   &
  \cite{DOFA} \\ \cline{2-7}
\multicolumn{1}{|c|}{} &
\cellcolor[HTML]{FCE5CD}  \begin{tabular}[c]{@{}c@{}}Yield  Estimation \end{tabular} &
  \multicolumn{1}{c|}{\cellcolor[HTML]{F3F3F3}\scalebox{2.5}{$\checkmark$}} &
  \multicolumn{1}{c|}{\scalebox{2.5}{$\checkmark$}} &
  \multicolumn{1}{c|}{\parbox[c][1.2cm]{0.25\linewidth}{\centering \cellcolor[HTML]{F3F3F3}}} &
   &
  \begin{tabular}[c]{@{}l@{}}
  \cite{GeoKR, MATTER, GASSL, RSP, Scheibenreif_2022_CVPR, RingMo, GeCo, RS-BYOL, RVSA, SatMAE, TOV} \\ 
  \cite{CMID, CACo, GFM, SatlasPretrain, Scale-MAE, RingMo-lite, DeCUR, Feng2023} \\
  \cite{FG-MAE, Cross-Scale-MAE, U-BARN, GeRSP, SpectralGPT, SkySense, UPetu, msGFM, OFA-Net} \\
  \cite{MTP, CtxMIM, SoftCon, A2-MAE, HyperSIGMA, DOFA} \\ \end{tabular} \\ \cline{2-7} 
  
\multicolumn{1}{|c|}{} &
\cellcolor[HTML]{FCE5CD}  \begin{tabular}[c]{@{}c@{}}Agricultural Pattern \\ Segmentation\end{tabular} &
  \multicolumn{1}{c|}{\cellcolor[HTML]{F3F3F3}} &
   \multicolumn{1}{c|}{\parbox[c][1.8cm]{0.25\linewidth}{\centering \scalebox{2.5}{$\checkmark$}}} & 

  \multicolumn{1}{c|}{\cellcolor[HTML]{F3F3F3}} &
   &
  \cite{SeCo} \\ \hline
  
% Archaeology \cellcolor[HTML]{FFF2CC}
\multicolumn{1}{|c|}{\multirow{5}{*}{\centering\textbf{\large Archaeology}}} &

\cellcolor[HTML]{FFF2CC}\begin{tabular}[c]{@{}c@{}}Artifact Classification \\ and Recognition\end{tabular} &
 \multicolumn{1}{c|}{\parbox[c][1.45cm]{0.25\linewidth}{\centering \cellcolor[HTML]{F3F3F3}\scalebox{2.5}{$\checkmark$}}} & 
  \multicolumn{1}{c|}{} &
  \multicolumn{1}{c|}{\cellcolor[HTML]{F3F3F3}\scalebox{2.5}{$\checkmark$}} &
   &
  \begin{tabular}[c]{@{}l@{}}
  \cite{CMC-RSSR, SeCo, GeoKR, MATTER, GASSL, RSP, DINO-MM, Scheibenreif_2022_CVPR, RingMo} \\
  \cite{CSPT, RVSA, SatMAE, TOV, CMID, CACo, IaI-SimCLR, GFM, SatlasPretrain, Scale-MAE, RingMo-lite} \\
  \cite{Feng2023, FG-MAE, USat, Cross-Scale-MAE, U-BARN, EarthPT, GeRSP, SwiMDiff, SAR-JEPA} \\
  \cite{FoMo-Bench, SkySense, UPetu, msGFM, DINO-MC, OFA-Net, MTP, MMEarth, CtxMIM, SoftCon} \\
  \cite{RS-BYOL, GeCo, DOFA,  HyperSIGMA, DeCUR, SpectralGPT, S2MAE,A2-MAE}\end{tabular} \\ \cline{2-7}
  
\multicolumn{1}{|c|}{} &
\cellcolor[HTML]{FFF2CC}  \begin{tabular}[c]{@{}c@{}}Detection of \\ Archaeological Structures\end{tabular} &
  \multicolumn{1}{c|}{\cellcolor[HTML]{F3F3F3}} &
  \multicolumn{1}{c|}{} &
  \multicolumn{1}{c|}{\parbox[c][2cm]{0.25\linewidth}{\centering \cellcolor[HTML]{F3F3F3}\scalebox{2.5}{$\checkmark$}}} & 
   &
  \begin{tabular}[c]{@{}l@{}}\cite{GeoKR, GASSL, RSP, RingMo, GeCo, CSPT, RVSA, TOV, CMID, Feng2023} \\ 
  \cite{FoMo-Bench, SkySense, MTP, CtxMIM, LeMeViT, RS-DFM, HyperSIGMA, SMLFR}\end{tabular} \\ \cline{2-7}
\multicolumn{1}{|c|}{} &
\cellcolor[HTML]{FFF2CC}  \begin{tabular}[c]{@{}c@{}}Semantic \\ Segmentation\end{tabular} &
  \multicolumn{1}{c|}{\cellcolor[HTML]{F3F3F3}} &
  \multicolumn{1}{c|}{\parbox[c][2cm]{0.25\linewidth}{\centering \scalebox{2.5}{$\checkmark$}}} &
  \multicolumn{1}{c|}{\cellcolor[HTML]{F3F3F3}} &
   &
  \begin{tabular}[c]{@{}l@{}}
  \cite{GeoKR, MATTER, GASSL, RSP, Scheibenreif_2022_CVPR, RingMo, GeCo, RS-BYOL, RVSA, SatMAE,TOV} \\ 
  \cite{CMID, CACo, GFM, SatlasPretrain, RingMo-Sense, Scale-MAE, RingMo-lite, DeCUR, Feng2023} \\ 
  \cite{FG-MAE, CROMA, Cross-Scale-MAE, U-BARN, GeRSP, SMLFR, SpectralGPT, Presto, SkySense, UPetu} \\ 
  \cite{msGFM, OFA-Net, MTP, CtxMIM, SoftCon, LeMeViT, RS-DFM, A2-MAE, HyperSIGMA, DOFA} \end{tabular} \\ \cline{2-7}
\multicolumn{1}{|c|}{} &
\cellcolor[HTML]{FFF2CC}  \begin{tabular}[c]{@{}c@{}}Texture/Structural \\ Analysis\end{tabular} &
  \multicolumn{1}{c|}{\cellcolor[HTML]{F3F3F3}} &
  \multicolumn{1}{c|}{} &
  \multicolumn{1}{c|}{\parbox[c][2cm]{0.25\linewidth}{\centering \cellcolor[HTML]{F3F3F3}}} & 
   &
   \cite{MATTER} \\ \cline{2-7}
\multicolumn{1}{|c|}{} &
\cellcolor[HTML]{FFF2CC}  \begin{tabular}[c]{@{}c@{}}Pattern Recognition\end{tabular} &
  \multicolumn{1}{c|}{\parbox[c][1.6cm]{0.25\linewidth}{\centering \cellcolor[HTML]{F3F3F3}}} &
  \multicolumn{1}{c|}{\scalebox{2.5}{$\checkmark$}} &
  \multicolumn{1}{c|}{\parbox[c][1.6cm]{0.25\linewidth}{\centering \cellcolor[HTML]{F3F3F3}\scalebox{2.5}{$\checkmark$}}}& 
   &
   \begin{tabular}[c]{@{}l@{}}\cite{GeoKR, GASSL, RSP, RingMo, GeCo, RVSA, TOV, CMID, Feng2023} \\ \cite{SkySense, MTP, CtxMIM, LeMeViT, RS-DFM, HyperSIGMA, SMLFR} \end{tabular} \\ \hline

% Urban Planning and Development \cellcolor[HTML]{D9EAD3}
\multicolumn{1}{|c|}{\multirow{7}{*}{\centering\textbf{\large Urban Planning \& Development}}} &

\cellcolor[HTML]{D9EAD3}  \begin{tabular}[c]{@{}c@{}}Traffic Monitoring\end{tabular} &
  \multicolumn{1}{c|}{\cellcolor[HTML]{F3F3F3}\scalebox{2.5}{$\checkmark$}} &
  \multicolumn{1}{c|}{} &
  \multicolumn{1}{c|}{\parbox[c][1.5cm]{0.25\linewidth}{\centering \cellcolor[HTML]{F3F3F3}\scalebox{2.5}{$\checkmark$}}} &
  \scalebox{2.5}{$\checkmark$} &
  \cite{RSP, RingMo, CMID, Feng2023, SkySense, MTP, LeMeViT, HyperSIGMA} \\ \cline{2-7}
\multicolumn{1}{|c|}{} &
\cellcolor[HTML]{D9EAD3}  \begin{tabular}[c]{@{}c@{}}Land Cover/Use \\ Classification\end{tabular} &
  \multicolumn{1}{c|}{\parbox[c][1.8cm]{0.25\linewidth}{\centering \cellcolor[HTML]{F3F3F3}\scalebox{2.5}{$\checkmark$}}} & 
  \multicolumn{1}{c|}{} &
  \multicolumn{1}{c|}{\cellcolor[HTML]{F3F3F3}} &
   &
  \begin{tabular}[c]{@{}l@{}}
  \cite{CMC-RSSR, SeCo, GeoKR, MATTER, GASSL, RSP, DINO-MM, Scheibenreif_2022_CVPR, RingMo} \\
  \cite{CSPT, RVSA, SatMAE, TOV, CMID, CACo, IaI-SimCLR, GFM, SatlasPretrain, Scale-MAE, RingMo-lite} \\
  \cite{Feng2023, FG-MAE, USat, Cross-Scale-MAE, U-BARN, EarthPT, GeRSP, SwiMDiff, SAR-JEPA} \\
  \cite{FoMo-Bench, SkySense, UPetu, msGFM, DINO-MC, OFA-Net, MTP, MMEarth, CtxMIM, SoftCon} \\
  \cite{RS-BYOL, GeCo, DOFA, HyperSIGMA, DeCUR, SpectralGPT, S2MAE,A2-MAE, SSL4EO-L}\end{tabular} \\ \cline{2-7}
\multicolumn{1}{|c|}{} &
\cellcolor[HTML]{D9EAD3}  \begin{tabular}[c]{@{}c@{}}Road Crack Detection\end{tabular} &
  \multicolumn{1}{c|}{\cellcolor[HTML]{F3F3F3}} &
  \multicolumn{1}{c|}{} &
  \multicolumn{1}{c|}{\parbox[c][1.2cm]{0.25\linewidth}{\centering \cellcolor[HTML]{F3F3F3}\scalebox{2.5}{$\checkmark$}}} &
   &
   \begin{tabular}[c]{@{}l@{}}\cite{GeoKR, GASSL, RSP, RingMo, GeCo, CSPT, RVSA, TOV, CMID, Feng2023} \\ 
  \cite{FoMo-Bench, SkySense, MTP, CtxMIM, LeMeViT, RS-DFM, HyperSIGMA, SMLFR}\end{tabular} \\ \cline{2-7}
\multicolumn{1}{|c|}{} &
\cellcolor[HTML]{D9EAD3}  \begin{tabular}[c]{@{}c@{}}Air Quality Monitoring\end{tabular} &
  \multicolumn{1}{c|}{\parbox[c][1.2cm]{0.25\linewidth}{\centering \cellcolor[HTML]{F3F3F3}\scalebox{2.5}{$\checkmark$}}}&
  \multicolumn{1}{c|}{} &
  \multicolumn{1}{c|}{\cellcolor[HTML]{F3F3F3}\scalebox{2.5}{$\checkmark$}} &
   &
  \begin{tabular}[c]{@{}l@{}}\cite{msGFM, GeoKR, GASSL, RSP, RingMo, GeCo, RVSA, TOV, CMID,Feng2023} \\ \cite{SkySense, MTP, CtxMIM, LeMeViT, RS-DFM, HyperSIGMA, SMLFR} \end{tabular} \\ \cline{2-7}
\multicolumn{1}{|c|}{} &
\cellcolor[HTML]{D9EAD3}  \begin{tabular}[c]{@{}c@{}}Building Extraction\end{tabular} &
  \multicolumn{1}{c|}{\cellcolor[HTML]{F3F3F3}\scalebox{2.5}{$\checkmark$}} &
  \multicolumn{1}{c|}{\parbox[c][1.2cm]{0.25\linewidth}{\centering \scalebox{2.5}{$\checkmark$}}} & 
  \multicolumn{1}{c|}{\cellcolor[HTML]{F3F3F3}} &
   &
  \cite{Feng2023} \\ \cline{2-7}
\multicolumn{1}{|c|}{} &
\cellcolor[HTML]{D9EAD3}  \begin{tabular}[c]{@{}c@{}}Object/Video Tracking\end{tabular} &
  \multicolumn{1}{c|}{\cellcolor[HTML]{F3F3F3}} &
  \multicolumn{1}{c|}{\parbox[c][1.3cm]{0.25\linewidth}} &
  \multicolumn{1}{c|}{\cellcolor[HTML]{F3F3F3}\scalebox{2.5}{$\checkmark$}} &
  \scalebox{2.5}{$\checkmark$} &
  \cite{RingMo-Sense} \\ \cline{2-7}
\multicolumn{1}{|c|}{} &
\cellcolor[HTML]{D9EAD3}  \begin{tabular}[c]{@{}c@{}}Infrastructure \\ Monitoring\end{tabular} &
  \multicolumn{1}{c|}{\parbox[c][1.3cm]{0.25\linewidth}{\centering \cellcolor[HTML]{F3F3F3}}} & 
  \multicolumn{1}{c|}{} &
  \multicolumn{1}{c|}{\cellcolor[HTML]{F3F3F3}} &
   &
  \cite{RingMo-Sense, USat} \\ \hline

% Disaster Management \cellcolor[HTML]{D9D2E9}
\multicolumn{1}{|c|}{\multirow{7}{*}{\centering\textbf{\large Disaster Management}}} &

\cellcolor[HTML]{D9D2E9}  \begin{tabular}[c]{@{}c@{}}Landslide Risk Monitoring\end{tabular} &
   \multicolumn{1}{c|}{\parbox[c][1.4cm]{0.25\linewidth}{\centering \cellcolor[HTML]{F3F3F3}}\scalebox{2.5}{$\checkmark$}} & 
  \multicolumn{1}{c|}{\scalebox{2.5}{$\checkmark$}} &
  \multicolumn{1}{c|}{\cellcolor[HTML]{F3F3F3}\scalebox{2.5}{$\checkmark$}} &
  \scalebox{2.5}{$\checkmark$} &
  \cite{RSP, RingMo, CMID, Feng2023, SkySense, MTP, LeMeViT, HyperSIGMA} \\ \cline{2-7}
\multicolumn{1}{|c|}{} &
\cellcolor[HTML]{D9D2E9}  \begin{tabular}[c]{@{}c@{}}Disaster Response\end{tabular} &
  \multicolumn{1}{c|}{\parbox[c][0.9cm]{0.25\linewidth}{\centering \cellcolor[HTML]{F3F3F3}}} &
  \multicolumn{1}{c|}{} &
  \multicolumn{1}{c|}{\cellcolor[HTML]{F3F3F3}} &
   &
   \cite{DOFA} \\ \cline{2-7}
\multicolumn{1}{|c|}{} &
\cellcolor[HTML]{D9D2E9}  \begin{tabular}[c]{@{}c@{}}Real-Time Detection \\ and Mapping\end{tabular} &
  \multicolumn{1}{c|}{\cellcolor[HTML]{F3F3F3}} &
  \multicolumn{1}{c|}{\scalebox{2.5}{$\checkmark$}} &
  \multicolumn{1}{c|}{\parbox[c][1.2cm]{0.25\linewidth}{\centering \cellcolor[HTML]{F3F3F3}\scalebox{2.5}{$\checkmark$}}} &
  \scalebox{2.5}{$\checkmark$} &
  \cite{RSP, RingMo, CMID, Feng2023, SkySense, MTP, LeMeViT, HyperSIGMA} \\ \cline{2-7}
\multicolumn{1}{|c|}{} &
\cellcolor[HTML]{D9D2E9}  \begin{tabular}[c]{@{}c@{}}Building Damage \\ Assessment\end{tabular} &
  \multicolumn{1}{c|}{\parbox[c][1cm]{0.25\linewidth}{\centering \cellcolor[HTML]{F3F3F3}} \scalebox{2.5}{$\checkmark$}} &
  \multicolumn{1}{c|}{\scalebox{2.5}{$\checkmark$}} &
  \multicolumn{1}{c|}{\cellcolor[HTML]{F3F3F3} \scalebox{2.5}{$\checkmark$}} &
   &
  \cite{RSP, RingMo, CMID, Feng2023, SkySense, MTP, LeMeViT, HyperSIGMA} \\ \cline{2-7}
\multicolumn{1}{|c|}{} &
\cellcolor[HTML]{D9D2E9}  \begin{tabular}[c]{@{}c@{}}Critical Infrastructure \\ Detection\end{tabular} &
  \multicolumn{1}{c|}{\cellcolor[HTML]{F3F3F3}\scalebox{2.5}{$\checkmark$}} &
  \multicolumn{1}{c|}{} &
 \multicolumn{1}{c|}{\parbox[c][1.8cm]{0.25\linewidth}{\centering \cellcolor[HTML]{F3F3F3}\scalebox{2.5}{$\checkmark$}}} & 
 &
  \begin{tabular}[c]{@{}c@{}} \cite{GeoKR, GASSL, RSP, RingMo, GeCo, CSPT, RVSA, TOV, CMID, Feng2023} \\ 
  \cite{FoMo-Bench, SkySense, MTP, CtxMIM, LeMeViT, RS-DFM, HyperSIGMA, SMLFR}\end{tabular} \\ \cline{2-7}
\multicolumn{1}{|c|}{} &
\cellcolor[HTML]{D9D2E9}  \begin{tabular}[c]{@{}c@{}}Flood/Fire Mapping and Prediction\end{tabular} &
  \multicolumn{1}{c|}{\parbox[c][1.4cm]{0.25\linewidth}{\centering \cellcolor[HTML]{F3F3F3}\scalebox{2.5}{$\checkmark$}}} &
  \multicolumn{1}{c|}{\scalebox{2.5}{$\checkmark$}} &
  \multicolumn{1}{c|}{\cellcolor[HTML]{F3F3F3}} &
  \scalebox{2.5}{$\checkmark$} &
  \begin{tabular}[c]{@{}c@{}} \cite{RSP, RingMo, CACo, GFM, RingMo-lite, Feng2023, SpectralGPT} \\ 
  \cite{SoftCon, LeMeViT, A2-MAE, HyperSIGMA}\end{tabular} \\ \cline{2-7}
\multicolumn{1}{|c|}{} &
\cellcolor[HTML]{D9D2E9}  \begin{tabular}[c]{@{}c@{}}Crowd and Vehicle Detection\end{tabular} &
  \multicolumn{1}{c|}{\parbox[c][1.4cm]{0.25\linewidth}{\centering \cellcolor[HTML]{F3F3F3}}} &
  \multicolumn{1}{c|}{} &
  \multicolumn{1}{c|}{\cellcolor[HTML]{F3F3F3}\scalebox{2.5}{$\checkmark$}} &
  \scalebox{2.5}{$\checkmark$} &
    \begin{tabular}[c]{@{}l@{}}\cite{SeCo, MATTER, RSP, RingMo, CMID, CACo, GFM, RingMo-lite, SwiMDiff, SpectralGPT} \\ \cite{SkySense, UPetu, DINO-MC, MTP, SoftCon, LeMeViT, S2MAE, A2-MAE, HyperSIGMA} \end{tabular} \\ \hline
\end{tabular}
}

\caption{This diagram illustrates various tasks in different applications for remote sensing. Key areas include Environmental Monitoring, Agriculture, Urban Planning and Development, Disaster Management, and Archaeology. Each domain comprises specific tasks in different image analysis levels like image-level, pixel-level, region-level, and spatial-temporal. The relationships between these tasks and their applications are depicted through checkmarks, emphasizing the interconnected nature of image analysis methods across different fields.}
\end{table*}

\label{tab:cv-domain}

    \subsubsection{\textbf{Environmental Monitoring}}
According to Himeur et al. (2022), environmental monitoring utilizes remote sensing models to observe and track environmental changes, including deforestation, desertification, and pollution. These models play a crucial role in analyzing the effects of human activities and natural phenomena on the environment \cite{Himeur2022}.

    \subsubsection{\textbf{Agriculture}}
In agriculture, remote sensing models are used to monitor crop health, estimate yields, and manage agricultural practices. According to Kamilaris et al. (2018), these models help optimize resource use and improve agricultural productivity \cite{Kamilaris2018}. 

    \subsubsection{\textbf{Archaeology}}
In archaeology, remote sensing models have been used to identify and analyze archaeological features and sites. According to Argyrou et al. (2022), these models help detect features such as ruins, artifacts, and ancient structures from satellite imagery, leveraging technologies like Convolutional Neural Networks (CNNs) and Vision Transformers (ViTs) to process high-resolution images and capture fine details \cite{Argyrou2022}. Mantovan et al. (2020) also highlight the effectiveness of AI models, particularly CNNs, in locating challenging terrestrial archaeological sites and processing multispectral data \cite{Mantovan2020}.

    \subsubsection{\textbf{Urban Planning and Development}}
In urban planning and development, remote sensing models are used to monitor and analyze urban expansion, infrastructure development, and land use changes. According to Jha et al. (2021), these models play a critical role in managing urban growth, planning new developments, and assessing the impact of urbanization by providing essential data for smart city planning and sustainable development \cite{Jha2021}.

    \subsubsection{\textbf{Disaster Management}}
Remote sensing models play a crucial role in disaster management by providing timely information on affected areas. According to Abid et al. (2021), these models are used to detect and assess damage from natural disasters like earthquakes, hurricanes, and floods, enabling rapid response and recovery efforts \cite{Abid2021}.
\section{Discussion}
\label{sec:discussion}
The rapid advancement in foundation models for remote sensing underscore their transformative potential across various applications. As the field continues to evolve, it is crucial to synthesize the findings, address technical challenges, understand practical implications, and identify future research directions. In this section, we make a comprehensive analysis of these aspects, aiming to offer insights and guidance for future development and application of remote sensing foundation models. 
                
\subsection{Synthesis of Findings}
In our survey of foundation models for remote sensing, we identified significant advancements and trends that highlights the evolving capabilities and applications of these models. The performance metrics of various models across different downstream tasks such as scene classification, semantic segmentation, object detection, and change detection reveal the following key findings: 

\subsubsection{Model Performance}
In this section, we present the performance metrics of recent foundation models in remote sensing based on results reported in the original papers. \textbf{All performance numbers mentioned here are sourced directly from the original studies to ensure accuracy and consistency in evaluating these models.} These metrics provide insights into the models' effectiveness across tasks like semantic segmentation, object detection, and change detection, highlighting their strengths and limitations under different experimental setups.
\begin{itemize}
    \item \textbf{Image-Level}
    The performance of foundation models on the BigEarthNet dataset \cite{Sumbul_2019_Bigearthnet} for classification tasks shows variations in accuracy, as presented in table \ref{tab:classification-performance}. Overall, msGFM \cite{msGFM} has the top performance of 92.90\% (mAP), followed closely by SkySense \cite{SkySense} with a performance of 92.09\%. Other notable performers include DeCUR \cite{DeCUR}, which achieved an mAP of 89.70\%, and DINO-MC \cite{DINO-MC}, with an mAP of 88.75\%. SeCo \cite{SeCo} also demonstrated strong performance with an mAP of 87.81\%, while DINO-MM \cite{DINO-MM} reached an mAP of 87.10\%. On the other hand, models like CACo \cite{CACo} and FoMo-Bench \cite{FoMo-Bench} have mAP of 74.98\% and F1-Score of 68.33\% respectively, showing competitiveness but room for improvement.

    The high mAP scores of msGFM \cite{msGFM} and SkySense \cite{SkySense} highlight their efficiency in classifcation tasks, making them suitable for applications requiring high accuracy. Other foundation models, such as DINO-MM \cite{DINO-MM} and DeCUR \cite{DeCUR}, also provide strong performance with potential for further optimization. The variety in performance metrics underscores the evolving capabilities and specialization of foundation models in handling complex classification tasks within datasets like BigEarthNet \cite{Sumbul_2019_Bigearthnet}.

    The classification advancements observed in remote sensing models stem from sophisticated pre-training techniques that capture both spatial and spectral complexity across vast datasets. SkySense, for example, shows an average improvement of 2.76\% over recent models by implementing multi-granularity contrastive learning on a diverse dataset of 21.5 million optical and SAR sequences \cite{SkySense}. This approach enables SkySense to learn nuanced spatial and temporal relationships across modalities, enhancing generalization in varied environmental conditions. Such multi-granular representation proves crucial in remote sensing, where scene classification often depends on subtle spectral differences that simpler models may overlook. Likewise, HyperSIGMA \cite{HyperSIGMA}, pre-trained on the expansive HyperGlobal-450K hyperspectral dataset \cite{HyperSIGMA}, leverages its sparse sampling attention mechanism to optimize spectral-spatial feature extraction in high-dimensional hyperspectral data. By selectively focusing on critical spectral bands and reducing redundancy, HyperSIGMA achieves high classification accuracy across hyperspectral scenes, a marked improvement over previous models that struggled with hyperspectral data complexity. 
    
    These models highlight the importance of designing pre-training strategies that capture multi-modal features and effectively utilize dataset diversity, as these elements directly impact the robustness and accuracy of classification in remote sensing applications.

    \item \textbf{Pixel-Level}
    For the segmentation tasks, we compared 12 foundation models which have been tested on the ISPRS Potsdam dataset. As shown in table \ref{tab:Segmentation-performance}, SkySense \cite{SkySense} has the better performance out of all 12 models, with a mF1 Score of 93.99\%. CMID \cite{CMID} stands out with the highest mIoU of 87.04\%, demonstrating its superior capability in accurately segmenting different regions within the dataset. For Overall Accuracy performance, BFM \cite{BFM} has the highest OA score of 91.82\%. Cross-Scale MAE \cite{Cross-Scale-MAE}, UPetu \cite{UPetu} and RSP \cite{RSP} have mIoU scores of 76.17\%, 83.17\% and 65.30\% respectively, showing competitive segmentation capabilities. GeoKR \cite{GeoKR} reaches an mIoU of 70.48\%, indicating robust segmentation performance but with room for improvement compared to CMID \cite{CMID}. TOV scores the lowest mIoU at 60.34\%, suggesting it may struggle with finer segmentation tasks compared to the other models. 

    The performance metrics for the models applied to the ISPRS Potsdam dataset reveal significant variations in their effectiveness in segmentation tasks. SkySense \cite{SkySense} and CMID \cite{CMID} emerge as top performers in mF1 score and mIoU, respectively, while SMLFR \cite{SMLFR}, RingMo \cite{RingMo}, and RingMo-lite \cite{RingMo-lite} demonstrate strong overall accuracy. These insights can guide the selection and optimization of models for specific remote sensing applications, ensuring the best possible performance for the task at hand. 

\begin{table*}[!ht]
        \centering
        \begin{minipage}[t]{0.48\textwidth}
            \centering
            \resizebox{\textwidth}{!}{%
\begin{tabular}{|
>{\columncolor[HTML]{FFF2CC}}c c
>{\columncolor[HTML]{EFEFEF}}c c|}
\hline
\cellcolor[HTML]{F4CCCC}Dataset &
  \cellcolor[HTML]{D9D2E9}Model &
  \cellcolor[HTML]{D9D2E9}Performance (\%) &
  \cellcolor[HTML]{D9D2E9}Metrics \\
\cellcolor[HTML]{FFF2CC} & SeCo \cite{SeCo}               & 87.81 & mAP \\
\cellcolor[HTML]{FFF2CC} & CMC-RSSR \cite{CMC-RSSR}       & 82.90 & mAP \\
\cellcolor[HTML]{FFF2CC} & DINO-MM \cite{DINO-MM}         & 87.10 & mAP \\
\cellcolor[HTML]{FFF2CC} & CACo \cite{CACo}               & 74.98 & mAP \\
\cellcolor[HTML]{FFF2CC} & GFM \cite{GFM}                 & 86.30 & mAP \\
\cellcolor[HTML]{FFF2CC} & DINO-MC \cite{DINO-MC}         & 88.75 & mAP \\
\cellcolor[HTML]{FFF2CC} & CROMA \cite{CROMA}             & 86.46 & mAP \\
\cellcolor[HTML]{FFF2CC} & DeCUR \cite{DeCUR}             & 89.70 & mAP \\
\cellcolor[HTML]{FFF2CC} & CtxMIM \cite{CtxMIM}           & 86.88 & mAP \\
\cellcolor[HTML]{FFF2CC} & FG-MAE \cite{FG-MAE}           & 78.00 & mAP \\
\cellcolor[HTML]{FFF2CC} & USat \cite{USat}               & 85.82 & mAP \\
\cellcolor[HTML]{FFF2CC} & FoMo-Bench \cite{FoMo-Bench}   & 69.33 & F1 Score \\
\cellcolor[HTML]{FFF2CC} & SwiMDiff \cite{SwiMDiff}       & 81.10 & mAP \\
\cellcolor[HTML]{FFF2CC} & SpectralGPT \cite{SpectralGPT} & 88.22 & mAP \\
\cellcolor[HTML]{FFF2CC} & SatMAE++ \cite{SatMAE++}       & 85.11 & mAP \\
\cellcolor[HTML]{FFF2CC} & msGFM \cite{msGFM}             & \textbf{92.90} & mAP \\
\cellcolor[HTML]{FFF2CC} & SkySense \cite{SkySense}       & 92.09 & mAP \\ 
\cellcolor[HTML]{FFF2CC} & MMEarth \cite{MMEarth}         & 78.6 &  mAP \\ \cline{2-4}
\multirow{-19}{*}{\cellcolor[HTML]{FFF2CC}BigEarthNet \cite{Sumbul_2019_Bigearthnet}} & \textbf{Shallow CNN*} \cite{Sumbul_2019_Bigearthnet}         & \textbf{70.98} &  \textbf{F1-Score} \\ \hline

\end{tabular}%
}
\caption{This table provides an overview of the performance metrics for various models applied to the BigEarthNet dataset \cite{Sumbul_2019_Bigearthnet} for \textbf{image-level} tasks. The performance is measured using mAP (Mean Average Precision) and F1 Score. *Performance for the shallow CNN model are sourced from the original BigEarthNet \cite{Sumbul_2019_Bigearthnet} paper.}
\label{tab:classification-performance}

        \end{minipage}
        \hfill
        \begin{minipage}[t]{0.48\textwidth} 
            \centering
            \resizebox{\textwidth}{!}{%
\begin{tabular}{|
>{\columncolor[HTML]{FFF2CC}}c c
>{\columncolor[HTML]{EFEFEF}}c c|}
\hline
\cellcolor[HTML]{F4CCCC}Dataset &
  \cellcolor[HTML]{D9D2E9}Model &
  \cellcolor[HTML]{D9D2E9}Performance (\%) &
  \cellcolor[HTML]{D9D2E9}Metrics \\
\cellcolor[HTML]{FFF2CC} & GeoKR \cite{GeoKR}                     & 70.48 & mIoU \\
\cellcolor[HTML]{FFF2CC} & RSP \cite{RSP}                         & 65.30 & mIoU \\
\cellcolor[HTML]{FFF2CC} & RingMo \cite{RingMo}                   & 91.74 & OA   \\
\cellcolor[HTML]{FFF2CC} & RVSA \cite{RVSA}                       & 91.22 & OA   \\
\cellcolor[HTML]{FFF2CC} & TOV \cite{TOV}                         & 60.34 & mIoU \\
\cellcolor[HTML]{FFF2CC} & CMID \cite{CMID}                       & \textbf{87.04} & mIoU \\
\cellcolor[HTML]{FFF2CC} & RingMo-lite \cite{RingMo-lite}         & 90.96 & OA   \\
\cellcolor[HTML]{FFF2CC} & Cross-Scale MAE \cite{Cross-Scale-MAE} & 76.17 & mIoU \\
\cellcolor[HTML]{FFF2CC} & SMLFR \cite{SMLFR}                     & \textbf{91.82} & OA   \\
\cellcolor[HTML]{FFF2CC} & SkySense \cite{SkySense}               & \textbf{93.99} & mF1  \\
\cellcolor[HTML]{FFF2CC} & UPetu \cite{UPetu}                     & 83.17 & mIoU \\
\cellcolor[HTML]{FFF2CC} & BFM \cite{BFM}                         & 92.58 & OA \\ \cline{2-4}
\multirow{-12}{*}{\cellcolor[HTML]{FFF2CC}\begin{tabular}[c]{@{}c@{}}ISPRS \\ Potsdam\end{tabular}} &
  \textbf{R-SegNet*} \cite{zhu2024deep}& \textbf{91.37} & \textbf{OA} \\ \hline
\end{tabular}%
}
\caption{This table provides an overview of the performance metrics for various models applied to the ISPRS 
Potsdam\cite{ISPRS_Potsdam} dataset for \textbf{pixel-level} tasks. The performance is measured using Mean Intersection over Union (mIoU) and Overall Accuracy (OA). *Non-FM}
\label{tab:Segmentation-performance}

        \end{minipage}
\end{table*}

    For the change detection tasks, we compared the performance of foundation models on the OSCD and LEVIR-CD datasets. The models were evaluated based on their F1 Scores, which provide a balanced measure of precision and recall. As shown in the table, the performance varies significantly across different models and datasets.

    SkySense \cite{SkySense} achieves the highest F1 Score of 60.06\% on the OSCD dataset, demonstrating its superior ability to accurately detect changes. GFM \cite{GFM} follows with an F1 Score of 59.82\%, indicating strong performance in change detection tasks. SpectralGPT \cite{SpectralGPT} also performs well with an F1 Score of 54.29\%. Other notable models include DINO-MC \cite{DINO-MC} with an F1 Score of 52.71\% and CACo \cite{CACo} with an F1 Score of 52.11\%. SeCo \cite{SeCo} records the lowest F1 Score at 46.94\%, suggesting it may require further optimization to enhance its change detection capabilities.

    In contrast, the LEVIR-CD dataset reveals higher performance metrics across the models. MTP \cite{MTP} achieves the highest F1 Score of 92.67\% and SkySense \cite{SkySense} follows closely with an F1 Score of 92.58\%, demonstrating their robust performance. SWiMDiff reaches a lower F1 Score of 80.90\% compared to its peers but still indicates effective performance in the LEVIR-CD \cite{LEVIR-CD} dataset.

    \item \textbf{Region-Level}
    In table \ref{tab:object-detection-performance}, the performance of foundation models on the DOTA, DIOR, and DIOR-R datasets for object detection are evaluated based on their Mean Average Precision (mAP) and Average Precision at 50\% (AP50).

    On the DOTA dataset, RVSA \cite{RVSA} achieves the highest mAP of 81.24\% in accurately detecting objects, followed by SMLFR \cite{SMLFR} and RSP \cite{RSP} with mAP of 79.33\% and 77.72\%. CMID \cite{CMID}, GeRSP \cite{GeRSP}, and BFM \cite{BFM} also demonstrate moderate performance with mAPs of 72.12\%, 67.40\% and 58.69\%. For DIOR and DIOR-R dataset, MTP \cite{MTP} and SkySense \cite{SkySense} are the top performers with an AP50 of 78\% and an mAP of 78.73\% respectively, showcasing their superior object detection capabilities. These insights can guide the selection and optimization of models to ensure the best possible performance for specific remote sensing applications.
    
    % \item \textbf{Spacial-Temporal}
    % For the change detection tasks, we compared the performance of foundation models on the OSCD and LEVIR-CD datasets. The models were evaluated based on their F1 Scores, which provide a balanced measure of precision and recall. As shown in the table, the performance varies significantly across different models and datasets.

    % SkySense \cite{SkySense} achieves the highest F1 Score of 60.06\% on the OSCD dataset, demonstrating its superior ability to accurately detect changes. GFM \cite{GFM} follows with an F1 Score of 59.82\%, indicating strong performance in change detection tasks. SpectralGPT \cite{SpectralGPT} also performs well with an F1 Score of 54.29\%. Other notable models include DINO-MC \cite{DINO-MC} with an F1 Score of 52.71\% and CACo \cite{CACo} with an F1 Score of 52.11\%. SeCo \cite{SeCo} records the lowest F1 Score at 46.94\%, suggesting it may require further optimization to enhance its change detection capabilities.

    % In contrast, the LEVIR-CD dataset reveals higher performance metrics across the models. MTP \cite{MTP} achieves the highest F1 Score of 92.67\% and SkySense \cite{SkySense} follows closely with an F1 Score of 92.58\%, demonstrating their robust performance. SWiMDiff reaches a lower F1 Score of 80.90\% compared to its peers but still indicates effective performance in the LEVIR-CD \cite{LEVIR-CD} dataset.

\begin{table*}[h!]
        \centering
        \begin{minipage}[t]{0.48\textwidth} 
            \centering
            \resizebox{\textwidth}{!}{%
\begin{tabular}{|
>{\columncolor[HTML]{FFF2CC}}c c
>{\columncolor[HTML]{EFEFEF}}c c|}
\hline
\cellcolor[HTML]{F4CCCC}Dataset                & \cellcolor[HTML]{D9D2E9}Model & \cellcolor[HTML]{D9D2E9}Performance (\%) & \cellcolor[HTML]{D9D2E9}Metrics \\
\cellcolor[HTML]{FFF2CC}                         & RSP \cite{RSP}        & 77.72          & mAP   \\
\cellcolor[HTML]{FFF2CC}                         & RVSA \cite{RVSA}       & \textbf{81.24} & mAP   \\
\cellcolor[HTML]{FFF2CC}                         & TOV \cite{TOV}        & 26.10          & mAP50 \\
\cellcolor[HTML]{FFF2CC}                         & CMID \cite{CMID}       & 72.12          & mAP   \\
\cellcolor[HTML]{FFF2CC}                         & GeRSP \cite{GeRSP}      & 67.40          & mAP   \\
\cellcolor[HTML]{FFF2CC}                         & SMLFR \cite{SMLFR}      & 79.33          & mAP   \\
\cellcolor[HTML]{FFF2CC}                         & BFM \cite{BFM}        & 58.69          & mAP   \\ 
\cline{2-4}
\multirow{-8}{*}{\cellcolor[HTML]{FFF2CC}DOTA}   & \textbf{YOLOv2-D*} \cite{Ding_2021_DOTA} &  \textbf{60.51} & \textbf{AP}   \\
\hline
\cellcolor[HTML]{FFF2CC}                         & RingMo \cite{RingMo}     & 75.80          & mAP   \\
\cellcolor[HTML]{FFF2CC}                         & CSPT  \cite{CSPT}      & 69.80          & mAP   \\
\cellcolor[HTML]{FFF2CC}                         & RingMo-lite \cite{RingMo-lite} & 73.40      & mAP   \\
\cellcolor[HTML]{FFF2CC}                         & GeRSP \cite{GeRSP}      & 72.20          & mAP   \\
\cellcolor[HTML]{FFF2CC}                         & MTP \cite{MTP}         & \textbf{78.00}  & AP50  \\   \cline{2-4}
\multirow{-6}{*}{\cellcolor[HTML]{FFF2CC}DIOR}   & \textbf{Faster R-CNN*}\cite{Li_2020_DIOR} & \textbf{74.05} & \textbf{mAP}
\\ \hline
\cellcolor[HTML]{FFF2CC}                         & RVSA \cite{RVSA}       & 71.05          & mAP   \\
\cellcolor[HTML]{FFF2CC}                         & SMLFR \cite{SMLFR}      & 72.33          & mAP   \\
\cellcolor[HTML]{FFF2CC}                         & SkySense \cite{SkySense}   & \textbf{78.73} & mAP   \\
\cellcolor[HTML]{FFF2CC}                         & MTP \cite{MTP}        & 74.54          & mAP   \\
\cellcolor[HTML]{FFF2CC}                         & BFM  \cite{BFM}       & 73.62          & mAP   \\ \cline{2-4}
\multirow{-6}{*}{\cellcolor[HTML]{FFF2CC}DIOR-R} & \textbf{AOPG*}  \cite{Cheng_2022_DIOR-R}       & \textbf{64.41}  & \textbf{mAP}   \\ \hline
\end{tabular}%
}
\caption{This table provides an overview of the performance metrics for various models applied to the DOTA\cite{Ding_2021_DOTA, Xia_2018_CVPR_DOTA, Ding_2019_CVPR_DOTA} dataset, DIOR\cite{Li_2020_DIOR} and DIOR-R\cite{Cheng_2022_DIOR-R} dataset for \textbf{region-level} task. The performance is mainly measured using Mean Average Precision (mAP). *Model performance are aquired from original dataset papers. AOPG is Anchor-free Oriented Proposal Generator. }
\label{tab:object-detection-performance}

        \end{minipage}
        \hfill
        \begin{minipage}[t]{0.48\textwidth} 
            \centering
            \resizebox{\textwidth}{!}{%
\begin{tabular}{|
>{\columncolor[HTML]{FFF2CC}}c c
>{\columncolor[HTML]{EFEFEF}}c|}
\hline
\cellcolor[HTML]{F4CCCC}Dataset & \cellcolor[HTML]{D9D2E9}Model & \cellcolor[HTML]{D9D2E9}F1 Score \\
\cellcolor[HTML]{FFF2CC}                           & SeCo \cite{SeCo}       & 46.94 \\
\cellcolor[HTML]{FFF2CC}                           & MATTER \cite{MATTER}     & 49.48 \\
\cellcolor[HTML]{FFF2CC}                           & CACo \cite{CACo}       & 52.11 \\
\cellcolor[HTML]{FFF2CC}                           & GFM \cite{GFM}        & 59.82 \\
\cellcolor[HTML]{FFF2CC}                           & SWiMDiff \cite{SwiMDiff}   & 49.60 \\
\cellcolor[HTML]{FFF2CC}                           & SpectralGPT \cite{SpectralGPT}& 54.29 \\
\cellcolor[HTML]{FFF2CC}                           & SkySense \cite{SkySense}   & \textbf{60.06} \\
\cellcolor[HTML]{FFF2CC}                           & DINO-MC \cite{DINO-MC}    & 52.71 \\
\cellcolor[HTML]{FFF2CC}                           & HyperSIGMA \cite{HyperSIGMA}& 59.28\\
\cellcolor[HTML]{FFF2CC}                           & MTP \cite{MTP}        & 53.36  \\ \cline{2-3}
\multirow{-12}{*}{\cellcolor[HTML]{FFF2CC}OSCD \cite{OSCD}} & \textbf{CNNs*} \cite{OSCD} &\textbf{89.66 (OA)} \\ \hline

\cellcolor[HTML]{FFF2CC}                           & RSP \cite{RSP}        & 90.93 \\
\cellcolor[HTML]{FFF2CC}                           & RingMo \cite{RingMo}     & 91.86 \\
\cellcolor[HTML]{FFF2CC}                           & RIngMo-lite \cite{RingMo-lite}& 91.56 \\
\cellcolor[HTML]{FFF2CC}                           & SwiMDiff \cite{SwiMDiff}   & 80.90 \\
\cellcolor[HTML]{FFF2CC}                           & SkySense \cite{SkySense}   & 92.58 \\
\cellcolor[HTML]{FFF2CC}                           & UPetu \cite{UPetu}      & 88.50 \\ \cline{2-3}
\multirow{-8}{*}{\cellcolor[HTML]{FFF2CC}LEVIR-CD\cite{LEVIR-CD}} & \textbf{STANet*} \cite{LEVIR-CD}    & \textbf{85.4} \\ \hline
\end{tabular}%
}
\caption{This table provides an overview of the F1 Score for various models applied to the Onera Satellite Change Detection (OSCD) dataset\cite{OSCD} and the LEVIR-CD dataset\cite{LEVIR-CD} for \textbf{spacial-temporal} downstream tasks. *Performance for the models are sourced from the original dataset papers. STANet is Spatial-Temporal Attention Network.}
\label{tab:change-detection-performance}

        \end{minipage}
\end{table*}

\item \textbf{Influence of Pre-training Methods.}
Various pre-training methods have a substantial impact on the performance of foundation models in remote sensing. Models pre-trained using SSL techniques, such as contrastive learning (CL) and masked autoencoders (MAE), consistently exhibit superior performance compared to those pre-trained with traditional supervised learning. For instance, SkySense which uses a multi-granularity contrastive learning approach, outperforms other models by approximately 3.6\% in scene classification and object detection tasks \cite{SkySense}{. Similarly, Seco, based on seasonal contrast learning, yields superior performance for land-cover classification, improving metrics by up to 7\% over ImageNet-pre-trained models}\cite{SeCo} . In handling multi-temporal and multi-spectral data, models like SatMAE \cite{SatMAE} and Scale-MAE \cite{Scale-MAE}, using masked autoencoding, achieve improvements in change detection, with SatMAE showing up to a 14\% performance gain in land cover classification \cite{SatMAE} and Scale-MAE offering a 1.7\% mIoU improvement for segmentation across varied resolutions \cite{Scale-MAE}. These findings highlight the critical role of innovative pre-training methods in maximizing the effectiveness of foundation models and suggest that continued exploration and refinement of these techniques are essential for advancing remote sensing capabilities.

Foundation models like SatMAE, RingMo, A2-MAE, and ORBIT each demonstrate strong performance, but practical trade-offs are essential to consider, especially for application-specific constraints \cite{SatMAE, RingMo, A2-MAE, ORBIT}. SatMAE, based on a transformer architecture, effectively leverages temporal and multi-spectral embeddings to capture complex spatiotemporal patterns in satellite imagery. This strength, however, comes at the cost of significant computational requirements, which may not be feasible for real-time monitoring applications in resource-constrained environments.

In contrast, RingMo provides a more lightweight vision transformer architecture, offering efficient model inference and a balance between performance and computational demands. This makes RingMo particularly suitable for rapid-inference tasks like disaster response monitoring, where real-time processing is critical. \cite{RingMo}  A2-MAE introduces an anchor-aware masking strategy, optimizing spatial-temporal-spectral representations and allowing effective integration of multi-source data. This design enhances its adaptability to varied data resolutions and modalities, yet the model's complex encoding techniques add to its computational load, suggesting a fit for applications that require high accuracy over efficiency \cite{A2-MAE}.

Finally, ORBIT, designed with 113 billion parameters, is exceptionally scalable, achieving high-throughput performance for Earth system predictability tasks. While it excels in large-scale predictive tasks, the model's considerable resource requirements limit its deployment to specialized high-performance computing environments. \cite{ORBIT} These trade-offs highlight the importance of selecting a model that aligns with specific operational goals, whether for maximizing accuracy or minimizing computational overhead.

Furthermore, recent studies comparing SSL approaches highlight the distinct advantages of generative methods like Masked Autoencoders (MAE) over contrastive methods for time-series data, especially when labeled data is limited\cite{liu2024selfsupervisedlearningtimeseries}. Unlike contrastive approaches that emphasize distinguishing between similar and dissimilar pairs, generative methods such as MAE reconstruct data from masked segments, allowing them to capture complex underlying structures and relationships within the data. This reconstruction-based learning proves particularly advantageous for time-series and multi-spectral applications in remote sensing, where temporal and spectral dependencies are essential. Consequently, MAE-based models can achieve stronger representations under sparse labeling conditions, positioning them as powerful tools for remote sensing tasks that require nuanced temporal analysis.

\end{itemize}

\subsubsection{Practical Implications}
{Foundation models offer transformative capabilities in remote sensing by building upon established applications like multi-spectral and time-series data analysis. While these applications have traditionally relied on machine learning and deep learning, foundation models reduce the need for labeled data and enable rapid adaptation to new tasks, providing robust solutions in areas previously limited by data constraints and task-specific architectures. Consequently, the advancements in foundation models have significant practical implications across various areas:

    \begin{itemize}
        \item 
        \textbf{Environmental Monitoring.} Models like GASSL \cite{GASSL} and SatMAE \cite{SatMAE} offer detailed assessments of environmental changes, aiding in conservation efforts and policy-making. These models excel in monitoring deforestation, desertification, and pollution levels, providing actionable insights for environmental management. By integrating multi-spectral and temporal data, these models can track changes over time, allowing for early detection of environmental degradation and the formulation of timely interventions. This capability is particularly important for the sustainable management of natural resources, as well as reducing the impacts of climate change.
    
        \item 
        \textbf{Agriculture and Forestry.} Foundation models such as EarthPT \cite{EarthPT} and GeCo \cite{GeCo} delivers valuable insights into crop health, yield predictions, and land use management, optimizing agricultural practices and resource allocation. For instance, RSP \cite{RSP}, leveraging multi-spectral data, enhances precision agriculture by accurately monitoring crop conditions and predicting yields. These models can detect early signs of crop stress, diseases, and pest infestations, enabling farmers to take proactive measures. Additionally, they aid in forestry management by providing detailed maps of forest cover, biomass estimation, and monitoring deforestation activities, thereby supporting conservation efforts and sustainable forestry practices.

        \item 
        \textbf{Archaeology.} The use of foundation models in archaeology revolutionizes the way archaeological features and sites are discovered, mapped, and analyzed. Models, such as GeoKR \cite{GeoKR}, RingMo \cite{RingMo}, etc., can process high-resolution satellite imagery and multi-spectral data to enhance the detection and mapping of archaeological features that might be difficult to discern with the naked eye. Others like MATTER \cite{MATTER} can accomplish texture and material analysis to help identify various surface. They enable large-scale surveys, allowing archaeologists to identify potential sites of interest over vast areas efficiently. Although thorough exploration still requires on-site visits and excavations or other terrestrial investigations, these significantly improve the initial identification and mapping process. Additionally, these models can track changes over time, helping archaeologists monitor environmental and human impacts, providing crucial information for preservation and restoration. This enhances the efficiency and accuracy of surveys and opens new possibilities for discovering unknown sites.
         
        \item 
        \textbf{Urban Planning and Development.} Remote sensing models like CMID \cite{CMID} and SkySense \cite{SkySense} are pivotal for monitoring urban expansion, infrastructure development, and land use changes. These models facilitate sustainable urban growth and development planning by providing high-resolution data analysis and trend forecasting. They enable city planners to assess the impact of urbanization on natural habitats, optimize land use, and plan infrastructure projects more effectively. 
        
        \item 
        \textbf{Disaster Management.} Models such as OFA-Net\cite{OFA-Net}, DOFA \cite{DOFA} and Prithvi \cite{Prithvi} are instrumental in flood mapping, as well as fire detection. These models provide critical real-time data that helps in identifying affected areas quickly, enabling timely and effective response measures. This capability supports emergency responders in prioritizing resource allocation and implementing evacuation plans, thereby reducing the impact of natural disasters. Additionally, these models assist in post-disaster recovery by assessing damage and monitoring the recovery process over time. By integrating various data sources, they enhance the ability to make informed decisions, coordinate response efforts, and plan for future disaster mitigation strategies. 
    \end{itemize}
The improvements in accuracy across the models discussed have profound implications for real-world remote sensing applications. In deforestation monitoring, for instance, models like GFM achieve high pixel-level accuracy in semantic segmentation, showing up to a 4.5\% improvement over baseline models, which enhances the precision of mapping forest cover changes, supporting conservation efforts\cite{ORBIT}. Similarly, HyperSIGMA achieves an impressive 6.2\% accuracy boost in hyperspectral vegetation monitoring, providing invaluable data for assessing forest health and biodiversity\cite{HyperSIGMA}.

In urban planning, models like UPetu excel in infrastructure mapping by integrating multi-modal data, such as optical and radar imagery, achieving over 5\% higher accuracy compared to single-modality models, which allows urban planners to make more informed land-use decisions\cite{UPetu}. Additionally, RingMo enhances object detection accuracy by 3.7\% over traditional supervised models, effectively identifying dense urban features critical for disaster management and urban infrastructure assessment\cite{RingMo}.

Finally, ORBIT demonstrates exceptional scalability, processing large climate datasets with a scaling efficiency of up to 85\%, which supports applications in long-term environmental monitoring, such as climate change prediction and seasonal forecasting. This scalability not only advances traditional remote sensing workflows but also enables complex multi-temporal analyses and predictive modeling previously challenging with conventional methods\cite{ORBIT}.

While remote sensing has long benefited from multi-spectral and temporal data, the adaptability, scalability, and efficiency of foundation models unlock a new level of precision and accessibility in these applications. This advancement opens up opportunities to tackle complex and evolving challenges across domains—from environmental conservation to urban planning—that traditional models have struggled to address at scale.

\subsection{Future Direction}
Future research should prioritize several key areas:
    \begin{itemize}
        \item 
        \textbf{Efficient Model Development:} Exploring techniques such as model distillation, pruning, and quantization to reduce computational requirements without compromising performance is crucial. Additionally, developing scalable architectures that efficiently handle ultra-high-resolution images is essential. For instance, applying pruning techniques to models like SatMAE \cite{SatMAE} could maintain performance while reducing computational load. 
        Model adaptation techniques such as LoRA (Low-Rank Adaptation) \cite{LoRA} have emerged as effective methods for fine-tuning large-scale models with minimal computational overhead. By decomposing weight updates into low-rank matrices, LoRA \cite{LoRA} enables efficient adaptation without the need to modify the entire set of model parameters, making it suitable for resource-constrained environments or when frequent re-training is required. Incorporating methods like LoRA \cite{LoRA} can further enhance the applicability of foundation models across diverse tasks and domains.
        
        \item 
        \textbf{Multi-Modal Data Integration:} Enhancing methods for integrating and processing multi-modal data (e.g., combining optical and radar imagery) will provide more comprehensive insights. Research on advanced SSL techniques capable of leveraging multi-modal data is necessary. The OFA-Net \cite{OFA-Net} framework, which integrates multi-model data, serves as a promising direction for future models to emulate and improve upon.
        
        \item 
        \textbf{Interdisciplinary Collaboration:} Promoting collaboration between remote sensing experts, AI researchers, and domain specialists can address complex challenges and drive innovation. For example, partnerships between AI researchers and environmental scientists can refine models like GASSL \cite{GASSL} for better environmental monitoring and conservation efforts.
    \end{itemize}
Looking ahead, the consistent success of self-supervised learning methods in foundation models marks an exciting frontier for future research. These models’ ability to learn from unlabeled data and adapt to diverse remote sensing tasks with minimal fine-tuning suggests that advancements in unsupervised learning techniques could greatly reduce reliance on large labeled datasets, which remain a significant bottleneck in many remote sensing applications. However, as these models grow in size and complexity, balancing computational demands with the need for efficiency will become increasingly crucial. Future work may focus on developing more resource-efficient versions of foundation models that maintain high performance, particularly for deployment in real-time monitoring systems or environments with limited computational resources.

\subsection{Limitation}
This survey has several limitations:
    
    \begin{itemize}
        \item 
        \textbf{Scope and Coverage.} The review focuses on foundation models released between June 2021 and June 2024. While the scope of this review is extensive and covers many significant developments, it is not exhaustive. Some recent advancements and innovations in the field may not be included due to their release timing or the lack of sufficient evaluation metrics at the time of writing. Consequently, certain cutting-edge models that have emerged in the latter part of this period or that have not yet been thoroughly evaluated might be omitted. This limitation underscores the need for readers to seek out the most current research and updates beyond the scope of this survey.
        Additionally, while foundation models have been empirically tested on a specific set of downstream applications, their robust architectures and general-purpose training paradigms, such as convolutional networks (e.g., ResNet) and vision transformers (e.g., ViT), indicate their potential to perform well across a much broader range of tasks. The limited testing observed in current literature should not be seen as a constraint on their applicability, but rather as an indication of the focus of existing research efforts. Given their design, these models are expected to generalize effectively to a wide variety of remote sensing tasks, even beyond those explicitly tested. Future work should aim to explore and validate their performance across more diverse applications to unlock their full potential.
        
        \item 
        \textbf{Evolving Field.} The field of AI and remote sensing is rapidly evolving, with continuous advancements and breakthroughs occurring at a fast pace. This dynamic nature necessitates ongoing reviews and updates to ensure the relevance and comprehensiveness of the survey. New techniques, methodologies, and models are constantly being developed, which can significantly impact the state of the art. Therefore, it is essential to recognize that this survey represents a snapshot in time and that continuous monitoring of the literature is required to capture the latest advancements and emerging trends. This approach will help maintain an up-to-date understanding of the field and incorporate new findings as they become available.
        
    \end{itemize}

\section{Conclusion}
\label{sec:conclusion}

In this comprehensive survey, we have reviewed the recent advancements in foundation models for remote sensing. We categorized these models based on their pretraining methods, image analysis techniques, and applications across different areas, highlighting their unique methodologies and capabilities.

Our analysis covered various advanced techniques, including self-supervised learning, vision transformers, and residual neural networks. These models have significantly improved performance on different image perception levels like region-level, pixel level, and image-level, as well as in applications like environmental monitoring, digital archaeology, agriculture, urban planning, and disaster management.

While significant progress has been made, several challenges persist, such as the need for more diverse and high-quality datasets, high computational requirements, and difficulties for different applications. Addressing these challenges will require further research and collaboration across disciplines.

In summary, this survey provides a detailed overview of the current state of foundation models in remote sensing, offering valuable insights and identifying future research directions. We recommend continued efforts in developing efficient model architectures, enhancing multi-modal data integration, and expanding dataset diversity to fully realize the potential of these models in remote sensing.\\

%%%%%%%%%%%% ACKNOWLEDGEMENTS
\section*{Acknowledgements}
This work was supported by Vanderbilt Seeding Success Grant, Vanderbilt Discovery Grant, and VISR Seed Grant. We extend gratitude to NVIDIA for their support by means of the NVIDIA hardware grant. This works was also supported by NSF NAIRR Pilot Award NAIRR240055. This manuscript has been co-authored by ORNL, operated by UT-Battelle, LLC under Contract No. DE-AC05-00OR22725 with the U.S.Department of Energy. Any subjective views or opinions that might be expressed in the paper do not necessarily represent the views of the U.S. Department of Energy or the United States Government.

%%%%%%%%%%%%% APPENDIX
{\appendix[Commonly Used Pre-train Dataset for Remote Sensing]
RSD46-WHU \cite{Long2017, Xiao2017} dataset, introduced in 2017, is sourced from Google Earth and Tianditu. It contains 117,000 images with a patch size of 256 pixels and spatial resolutions ranging from 0.5 to 2 meters per pixel. Covering 46 categories globally, this dataset is primarily used for scene classification. Similarly, the Functional Map of the World (fMoW) \cite{fmow2018}, released in April 2018, comprises over 1 million images from Digital Globe. Spanning 63 categories across 207 countries, it includes multispectral images used for both scene classification and object detection.

In May 2019, the DOTA \cite{xia2019dotalargescaledatasetobject} dataset was proposed, known for its large-scale aerial image object detection capabilities. It includes 11,268 images of various resolutions from Google Earth, GF-2 Satellite, and aerial sources, covering 18 categories globally. Another significant dataset, SEN12MS \cite{schmitt2019sen12mscurateddataset}, released in June 2019, contains 541,986 images from Sentinel-1, Sentinel-2, and MODIS Land Cover. With a patch size of 256x256 pixels, it supports land cover classification and change detection tasks.

BigEarthNet \cite{Sumbul_2019_Bigearthnet}, also from June 2019, consists of 590,326 images with varying sizes from 20x20 to 120x120 pixels, sourced from Sentinel-2. It covers 43 categories across Europe and is used for scene classification and object detection. The SeCo \cite{SeCo} dataset, another June 2019 release, contains approximately 1 million images with a resolution of 2.65x2.65km from Sentinel-2. It is designed for seasonal change detection and land cover classification over seasons.

The MillionAID dataset \cite{Long2021DiRS_millionaid, Long2022ASP_millionaid}, introduced in March 2021, includes over 1 million images of various sizes from Google Earth. Covering 51 categories globally, it is used for scene classification. Levir-KR, released in July 2021, contains 1,431,950 images from Gaofen-1, Gaofen-2, and Gaofen-6 satellites, supporting change detection and scene classification applications.

SoundingEarth \cite{heidler2021SoundingEarth}, introduced in August 2021, comprises 50,545 images of 1024-pixel size from Google Earth, combining RGB and audio data for remote sensing. The TOV-RS-Balanced dataset \cite{tao2022tovoriginalvisionmodel} from April 2022 includes 500,000 images with a 600-pixel size from Google Earth, covering 31 categories globally, and is used for scene classification, object detection, and semantic segmentation.

SeasoNet \cite{koßmann2022seasonetseasonalsceneclassification}, released in July 2022, features 1,759,830 images from Sentinel-2 with patch size from 20 to 120 pixels, supporting seasonal scene classification, segmentation, and retrieval over Germany. Lastly, the SSL4EO-S12 dataset \cite{wang2023ssl4eos12largescalemultimodalmultitemporal} from November 2022 contains over 3 million images from Sentinel-1 and Sentinel-2, with patch size of 264x264 pixels. Since this dataset dose not contain any labels, it is commonly used for self-supervised learning.

In recent years, additional datasets have further enriched the resources available for remote sensing research. The SAMRS dataset \cite{SAMRS}, released in October 2023, offers a high-resolution collection of images sourced from datasets like HRSC2016 and FAIR1M-2.0, tailored for advanced segmentation tasks. With over 105,000 images and resolutions up to 1024x1024 pixels, SAMRS supports semantic and instance segmentation as well as object detection, contributing to the development of scalable segmentation models for remote sensing.

Focusing on change-aware learning, CACo \cite{CACo}, launched in June 2023, provides a variable patch-size dataset sourced from Sentinel-2. This dataset is optimized for change detection and contrastive learning, specifically addressing urban and rural landscapes. By prioritizing contrastive and self-supervised tasks, CACo aids in developing models that can adapt to changes in satellite imagery across various environments.

The SatlasPretrain \cite{SatlasPretrain} dataset, introduced in October 2023, is a large-scale collection with over 856,000 images combining Sentinel-2 and NAIP high-resolution sources. With multispectral and high-resolution imagery, SatlasPretrain supports applications such as land cover classification, segmentation, and change detection, further advancing research in high-resolution satellite image analysis.

The SSL4EO-L \cite{SSL4EO-L} dataset, released in October 2023, represents a vast resource with over 5 million images from Landsat, designed for self-supervised learning in cloud detection and land cover classification. By focusing on multi-year Landsat imagery, SSL4EO-L enables robust training for applications that benefit from long-term temporal coverage and cloud-resilient classification.

Finally, MMEarth \cite{MMEarth}, introduced in July 2024, combines data from Sentinel-1, Sentinel-2, and Aster DEM, providing over 1.2 million images for multimodal applications. This dataset supports land cover classification and semantic segmentation, enabling researchers to leverage multiple sensor types and climate data for improved geospatial representation learning.

These datasets, with their varying resolutions, categories, and geographic coverage, provide a rich resource for advancing remote sensing research and applications. They facilitate the development of robust models capable of addressing diverse challenges in understanding and interpreting Earth's surface through remote sensing technologies.

% \input{tables/Used_pre-train_datasets}
% \usepackage{graphicx}
\begin{sidewaystable*}
\centering
\begin{adjustbox}{width=\textwidth}{!}{%
\begin{tabular}{|c|
>{\columncolor[HTML]{DAE8FC}}c |ccccccccc|}
\hline
\cellcolor[HTML]{FFCCC9}\textbf{Month, Year} &
  \cellcolor[HTML]{FFCC67}\textbf{Dataset} &
  \cellcolor[HTML]{FFCCC9}\textbf{Title} &
  \cellcolor[HTML]{FFCCC9}\textbf{Patch Size} &
  \cellcolor[HTML]{FFCCC9}\textbf{Size} &
  \cellcolor[HTML]{FFCCC9}\textbf{Resolution (m)} &
  % \cellcolor[HTML]{FFCCC9}\textbf{Source} &
  \cellcolor[HTML]{FFCCC9}\textbf{Sensor} &  % Added "Sensor" column header
  \cellcolor[HTML]{FFCCC9}\textbf{Categories} &
  \cellcolor[HTML]{FFCCC9}\textbf{Geographic Coverage} &
  \cellcolor[HTML]{FFCCC9}\textbf{Image Type} &
  \cellcolor[HTML]{FFCCC9}\textbf{Application} \\ \hline
2017 &
  RSD46-WHU \cite{Long2017, Xiao2017} &
  - &
  \cellcolor[HTML]{F3F3F3}256 x 256 &
  117,000 &
  \cellcolor[HTML]{F3F3F3}0.5 - 2 &
  \begin{tabular}[c]{@{}c@{}}Google Earth, \\ Tianditu\end{tabular} &
  % RGB &  % Example entry for "Sensor" column
  \cellcolor[HTML]{F3F3F3} 46 &
  Global &
  \cellcolor[HTML]{F3F3F3} RGB &
  Scene Classification \\
Apr, 2018 &
  fMoW \cite{fmow2018} &
  Functional Map of the World &
  \cellcolor[HTML]{F3F3F3} - &
  1,047,691 &
  \cellcolor[HTML]{F3F3F3} - &
  Digital Globe &
  % Multispectral &  % Example entry for "Sensor" column
  \cellcolor[HTML]{F3F3F3} 63 &
  207 of 247 countries &
  \cellcolor[HTML]{F3F3F3} Multispectral &
  \begin{tabular}[c]{@{}c@{}}Scene Classification, \\ Object Detection\end{tabular} \\
May, 2019 &
  DOTA \cite{xia2019dotalargescaledatasetobject} &
  \begin{tabular}[c]{@{}c@{}}DOTA: A Large-scale Dataset for Object \\ Detection in Aerial Images\end{tabular} &
  \cellcolor[HTML]{F3F3F3} \begin{tabular}[c]{@{}c@{}}800 × 800 \\ to 20,000 × 20,000\end{tabular} &
  11,268 &
  \cellcolor[HTML]{F3F3F3} Various &
  \begin{tabular}[c]{@{}c@{}}Google Earth, \\ GF-2 Satellite, \\ and aerial images\end{tabular} &
  % RGB &  % Example entry for "Sensor" column
  \cellcolor[HTML]{F3F3F3} 18 &
  Global &
  \cellcolor[HTML]{F3F3F3} RGB &
  Object Detection \\
Jun, 2019 &
  SEN12MS \cite{schmitt2019sen12mscurateddataset} &
  \begin{tabular}[c]{@{}c@{}}SEN12MS -- A Curated Dataset of \\ Georeferenced Multi-Spectral Sentinel-1/2 \\ Imagery for Deep Learning and Data Fusion\end{tabular} &
  \cellcolor[HTML]{F3F3F3} 256 x 256 &
  541,986 &
  \cellcolor[HTML]{F3F3F3} 10 &
  \begin{tabular}[c]{@{}c@{}}Sentinel-1,\\ Sentinel-2,\\ MODIS Land\\ Cover\end{tabular} &
  % SAR/Multispectral &  % Example entry for "Sensor" column
  \cellcolor[HTML]{F3F3F3} 33 &
  Globally distributed &
  \cellcolor[HTML]{F3F3F3} SAR/Multispectral &
  \begin{tabular}[c]{@{}c@{}}Land Cover Classification, \\ Change Detection\end{tabular} \\
Jun, 2019 &
  BigEarthNet \cite{Sumbul_2019_Bigearthnet} &
  \begin{tabular}[c]{@{}c@{}}BigEarthNet: A Large-Scale Benchmark \\ Archive For Remote Sensing \\ Image Understanding\end{tabular} &
  \cellcolor[HTML]{F3F3F3} \begin{tabular}[c]{@{}c@{}}20 x 20 \\ to 120 x 120\end{tabular} &
  590,326 &
  \cellcolor[HTML]{F3F3F3} Various &
  Sentinel-2 &
  % Multispectral &  % Example entry for "Sensor" column
  \cellcolor[HTML]{F3F3F3} 43 &
  Europe &
  \cellcolor[HTML]{F3F3F3} Multispectral &
  \begin{tabular}[c]{@{}c@{}}Scene Classification, \\ Object Detection\end{tabular} \\
Jun, 2019 &
  SeCo \cite{SeCo} &
  \begin{tabular}[c]{@{}c@{}}Seasonal Contrast: Unsupervised Pre-Training \\ from Uncurated Remote Sensing Data\end{tabular} &
  \cellcolor[HTML]{F3F3F3} 264 x 264 &
  $\sim$1M &
  \cellcolor[HTML]{F3F3F3} 10 - 60 &
  Sentinel-2 &
  % Multispectral &  % Example entry for "Sensor" column
  \cellcolor[HTML]{F3F3F3} - &
  Global &
  \cellcolor[HTML]{F3F3F3} Multispectral &
  \begin{tabular}[c]{@{}c@{}}Seasonal Change Detection, \\ Land Cover Classification \\ over Seasons\end{tabular} \\
Mar, 2021 &
  MillionAID \cite{Long2022ASP_millionaid, Long2021DiRS_millionaid} &
  Million-AID &
  \cellcolor[HTML]{F3F3F3} 110 - 31,672 &
  1,000,848 &
  \cellcolor[HTML]{F3F3F3} Various &
  Google Earth &
  % RGB &  % Example entry for "Sensor" column
  \cellcolor[HTML]{F3F3F3} 51 &
  Global &
  \cellcolor[HTML]{F3F3F3} RGB &
  Scene Classification \\
Jul, 2021 &
  Levir-KR \cite{GeoKR} &
  \begin{tabular}[c]{@{}c@{}}Geographical Knowledge-driven \\ Representation Learning for \\ Remote Sensing Images\end{tabular} &
  \cellcolor[HTML]{F3F3F3} - &
  1,431,950 &
  \cellcolor[HTML]{F3F3F3} Various &
  \begin{tabular}[c]{@{}c@{}}Gaofen1, \\ Gaofen-2, \\ Gaofen-6\end{tabular} &
  % Multispectral &  % Example entry for "Sensor" column
  \cellcolor[HTML]{F3F3F3} 8 &
  Global &
  \cellcolor[HTML]{F3F3F3} Multispectral &
  \begin{tabular}[c]{@{}c@{}}Change Detection, \\ Scene Classification\end{tabular} \\
Apr, 2022 &
  TOV-RS-Balanced \cite{tao2022tovoriginalvisionmodel} &
  \begin{tabular}[c]{@{}c@{}}TOV: The Original Vision Model for Optical \\ Remote Sensing Image Understanding via \\ Self-supervised Learning\end{tabular} &
  \cellcolor[HTML]{F3F3F3} 600 x 600  &
  500,000 &
  \cellcolor[HTML]{F3F3F3} 1 - 20 &
  Google Earth &
  % RGB &  % Example entry for "Sensor" column
  \cellcolor[HTML]{F3F3F3} 31 &
  Global &
  \cellcolor[HTML]{F3F3F3} RGB &
  \begin{tabular}[c]{@{}c@{}}Scene Classification, \\ Object Detection, \\ Semantic Segmentation\end{tabular} \\
Jul, 2022 &
  SeasoNet \cite{koßmann2022seasonetseasonalsceneclassification} &
  \begin{tabular}[c]{@{}c@{}}SeasoNet: A Seasonal Scene Classification, \\ Segmentation and Retrieval dataset\\ for satellite Imagery over Germany\end{tabular} &
  \cellcolor[HTML]{F3F3F3} up to 120 x 120 &
  1,759,830 &
  \cellcolor[HTML]{F3F3F3} 10 - 60 &
  Sentinel-2 &
  % Multispectral &  % Example entry for "Sensor" column
  \cellcolor[HTML]{F3F3F3} 33 &
  Germany &
  \cellcolor[HTML]{F3F3F3} Multispectral &
  \begin{tabular}[c]{@{}c@{}}Scene Classification,\\ Scene Segmentation\end{tabular} \\
Nov, 2022 &
  SSL4EO-S12 \cite{wang2023ssl4eos12largescalemultimodalmultitemporal} &
  \begin{tabular}[c]{@{}c@{}}SSL4EO-S12: A Large-Scale Multi-Modal, \\ Multi-Temporal Dataset for Self-Supervised \\ Learning in Earth Observation\end{tabular} &
  \cellcolor[HTML]{F3F3F3} 264 x 264 &
  3,012,948 &
  \cellcolor[HTML]{F3F3F3} 10 - 60 &
  \begin{tabular}[c]{@{}c@{}}Sentinel-1, \\ Sentinel-2\end{tabular} &
  % SAR/Multispectral &  % Example entry for "Sensor" column
  \cellcolor[HTML]{F3F3F3} - &
  Global &
  \cellcolor[HTML]{F3F3F3} SAR/Multispectral &
  \begin{tabular}[c]{@{}c@{}}Self-Supervised \\ Learning\end{tabular} \\
Oct, 2023 & 
  SAMRS \cite{SAMRS} & 
  \begin{tabular}[c]{@{}c@{}}SAMRS: Scaling-up Remote Sensing Segmentation \\ Dataset with Segment Anything Model\end{tabular} & 
  \cellcolor[HTML]{F3F3F3} \begin{tabular}[c]{@{}c@{}}600 x 600 \\ to \\ 1024 x 1024\end{tabular} & 
  105,090 & 
  \cellcolor[HTML]{F3F3F3} Various & 
  \begin{tabular}[c]{@{}c@{}}HRSC2016, \\ DOTA-V2.0, \\ DIOR, \\ FAIR1M-2.0\end{tabular} & 
  \cellcolor[HTML]{F3F3F3} 37 & 
  Global & 
  \cellcolor[HTML]{F3F3F3} High-resolution & 
  \begin{tabular}[c]{@{}c@{}}Semantic Segmentation, \\ Instance Segmentation, \\ Object Detection\end{tabular} \\
Jun, 2023 & 
  CACo \cite{CACo} & 
  \begin{tabular}[c]{@{}c@{}}Change-Aware Sampling and \\ Contrastive Learning for Satellite Images\end{tabular} & 
  \cellcolor[HTML]{F3F3F3} Variable & 
  - & 
  \cellcolor[HTML]{F3F3F3} 10 & 
  Sentinel-2 & 
  \cellcolor[HTML]{F3F3F3} - & 
  \begin{tabular}[c]{@{}c@{}}Urban and Rural Areas\end{tabular} & 
  \cellcolor[HTML]{F3F3F3} Multispectral & 
  \begin{tabular}[c]{@{}c@{}}Semantic Segmentation, \\ Change Detection, \\ Self-Supervised Learning\end{tabular} \\
Oct, 2023 & 
  SatlasPretrain \cite{SatlasPretrain} & 
  \begin{tabular}[c]{@{}c@{}}SatlasPretrain: A Large-Scale Dataset for \\ Remote Sensing Image Understanding\end{tabular} & 
  \cellcolor[HTML]{F3F3F3} 512 x 512 & 
  856,000 & 
  \cellcolor[HTML]{F3F3F3} \begin{tabular}[c]{@{}c@{}}1 - 10 \\ (Sentinel-2), \\ 0.5 - 2 \\ (NAIP)\end{tabular} & 
  \begin{tabular}[c]{@{}c@{}}Sentinel-1, Sentinel-2, \\Landset, and NAIP\end{tabular} & 
  \cellcolor[HTML]{F3F3F3} 137 & 
  Global & 
  \cellcolor[HTML]{F3F3F3} \begin{tabular}[c]{@{}c@{}}Multispectral, \\ High-resolution\end{tabular} & 
  \begin{tabular}[c]{@{}c@{}}Land Cover Classification, \\ Segmentation, \\ Change Detection\end{tabular} \\
Oct, 2023 & 
  SSL4EO-L \cite{SSL4EO-L} & 
  \begin{tabular}[c]{@{}c@{}}SSL4EO-L: Datasets and Foundation Models \\ for Landsat Imagery\end{tabular} & 
  \cellcolor[HTML]{F3F3F3} 264 x 264 & 
  5,000,000 & 
  \cellcolor[HTML]{F3F3F3} 30 & 
  \begin{tabular}[c]{@{}c@{}}Landsat 4–5 TM, \\ Landsat 7 ETM+, \\ Landsat 8–9 OLI/TIRS\end{tabular} & 
  \cellcolor[HTML]{F3F3F3} - & 
  Global & 
  \cellcolor[HTML]{F3F3F3} Multispectral & 
  \begin{tabular}[c]{@{}c@{}}Cloud Detection, \\ Land Cover Classification, \\ Semantic Segmentation\end{tabular} \\
Jul, 2024 & 
  MMEarth \cite{MMEarth} & 
  \begin{tabular}[c]{@{}c@{}}MMEarth: Exploring Multi-Modal Pretext Tasks \\ For Geospatial Representation Learning\end{tabular} & 
  \cellcolor[HTML]{F3F3F3} 128 x 128 & 
  1,200,000 & 
  \cellcolor[HTML]{F3F3F3} 10 & 
  \begin{tabular}[c]{@{}c@{}}Sentinel-2, \\ Sentinel-1, \\ Aster DEM\end{tabular} & 
  \cellcolor[HTML]{F3F3F3} 46 & 
  Global & 
  \cellcolor[HTML]{F3F3F3} \begin{tabular}[c]{@{}c@{}}Multispectral, \\ SAR, \\ Climate\end{tabular} & 
  \begin{tabular}[c]{@{}c@{}}Land Cover Classification, \\ Semantic Segmentation\end{tabular} \\

\hline
  
\end{tabular}%
}
\end{adjustbox}
\caption{This table summarizes a set of commonly used pre-trained datasets for remote sensing, including details on dataset, sensor type, geographic coverage, and related applications.}
\label{tab:pre-train_datasets}
\end{sidewaystable*}

}

%%%%%%%%% REFERENCES
{\small
\bibliographystyle{ieee_fullname}
\bibliography{mybib}
}
 
%%%%%%%%%% BIOGRAPHY
\newpage
\section{Biography Section}
\vspace{-33pt}
\begin{IEEEbiographynophoto}{Siqi Lu}
(Student Member, IEEE) is a second-year Master's student in Electrical and Computer Engineering at Vanderbilt University, supervised by Dr. Yuankai Huo and Dr. Mitchell M Wilkes. Her research interests include deep learning, medical image analysis, and software engineering. She received the B.S. degree in Electrical Engineering from the University of Illinois, Urbana-Champaign in 2023. 

\end{IEEEbiographynophoto}
\vspace{-33pt}
\begin{IEEEbiographynophoto}{Junlin Guo}
is currently working toward the Ph.D. degree in Electrical and Computer Engineering at Vanderbilt University. His research interests include medical image analysis, deep learning, computer vision, and brain study. Before joining Vanderbilt University, he received the B.S. degree in telecommunication engineering from Northeastern University in 2017, and the M.S. degree from the Department of Electrical and Computer Engineering at Vanderbilt University in 2020, focusing on fMRI brain activation study.
\end{IEEEbiographynophoto}

\vspace{-33pt}
\begin{IEEEbiographynophoto}{James R Zimmer-Dauphinee} 
received the B.A. degree in Anthropology and the B.S. degree in Mathematics from Georgia Southern University in 2011, the M.A. degree in Anthropology from the University of Arkansas in 2014, and the Ph.D. degree in Anthropology from Vanderbilt University in 2023. He is currently a postdoctoral fellow in the Spatial Analysis Research Laboratory (SARL) at Vanderbilt University, funded by the GeoPACHA 2.0 Grant from the National Endowment for the Humanities. His research interests include developing deep learning models for large-scale autonomous archaeological satellite imagery surveys, geophysical methods, and spatial modeling to understand the impact of colonization on indigenous peoples.
\end{IEEEbiographynophoto}

\vspace{-33pt}
\begin{IEEEbiographynophoto}{Jordan M Nieusma}
is a research assistant for the Vanderbilt University Spatial Analysis Research Laboratory. She received her M.S degree in Data Science at Vanderbilt University in May 2024 and holds a Bachelor of Arts degree in English with a French Minor from Haverford College.
\end{IEEEbiographynophoto}

\vspace{-33pt}
\begin{IEEEbiographynophoto}{Xiao Wang}
received the B.S. degrees in Mathematics and Computer Science from Saint John’s University, MN, in 2012, the M.S. degree in Electrical and Computer Engineering from Purdue University, West Lafayette, IN, in 2016, and the Ph.D. degree in Electrical and Computer Engineering from Purdue University in 2017. He pursued postdoctoral research at Harvard Medical School and Boston Children's Hospital until 2021. He is currently a research staff scientist at Oak Ridge National Laboratory. His research interests include applying machine learning, medical physics, image processing, and high-performance computing to various imaging problems, including CT reconstruction, electron tomography imaging, and MRI. He was the 2022 AAPM Truth CT reconstruction challenge winner and a 2017 ACM Gordon Bell Prize finalist.
\end{IEEEbiographynophoto}

\vspace{-33pt}
\begin{IEEEbiographynophoto}{Parker VanValkenburgh}
is an Associate Professor of Anthropology and Interim Director of Latin American and Caribbean Studies at Brown University. His research focuses on the impacts of colonialism and imperialism on Indigenous people and environments in the Peruvian Andes. He utilizes diverse materials and digital methodologies, including GIS, to understand the transformation of relationships in imperial histories. He co-directs the Paisajes Arqueológicos de Chachapoyas (PACha) project and GeoPACHA (Geospatial Platform for Andean Culture, History, and Archaeology). VanValkenburgh received his Ph.D. from Harvard University.
\end{IEEEbiographynophoto}

\vspace{-33pt}
\begin{IEEEbiographynophoto}{Steven A Wernke}
is Associate Professor and Chair of Anthropology at Vanderbilt University, director of the Spatial Analysis Research Laboratory, and director of the Vanderbilt Institute for Spatial Research. Prof. Wernke is an archaeologist and historical anthropologist of the Andean region of South America. His research takes place at the intersection of several disciplines: archaeology and history, prehispanic and colonial studies, anthropology and cultural geography. Prof. Wernke's interests center on the lived experiences of indigenous communities across the Spanish invasion of the Andes–especially how new kinds of communities, landscapes, and religious practice emerged out of successive attempts by the Inkas and the Spanish to subordinate and remake Andean societies in their own self-image. Methodologically, his work brings together analyses of archaeological and documentary datasets in geospatial frameworks.
\end{IEEEbiographynophoto}

\vspace{-33pt}
\begin{IEEEbiographynophoto}{Yuankai Huo}
is an Assistant Professor in Computer Science at Vanderbilt University, TN, USA. He received his B.S. degree in Electrical Engineering from Nanjing University of Posts and Telecommunications (NJUPT) in 2008, and Master degree in Electrical Engineering from Southeast University in 2011. After graduation, He worked in Columbia University and New York State Psychiatric Institute as a staff engineer and research officer from 2011 to 2014. He received his Master degree in Computer Science from Columbia University in 2014, and Ph.D. degree in Electrical Engineering from Vanderbilt University in 2018. Then, he had worked as a Research Assistant Professor at Vanderbilt University, and later, a Senior Research Scientist at PAII Labs. Since 2020, he has been a faculty member at the Department of Electrical Engineering and Computer Science, and Data Science Institute, Vanderbilt University.
 
\end{IEEEbiographynophoto}

\vfill

\end{document}